\documentclass[10pt,twocolumn,letterpaper]{article}

\usepackage[pagenumbers]{cvpr} % To force page numbers, e.g. for an arXiv version

\pdfoutput=1

\usepackage{graphicx}
\usepackage{amsmath}
\usepackage{amssymb}
\usepackage{booktabs}
\usepackage{pifont}
\usepackage{courier}
\usepackage{array}
\usepackage{multirow}
\usepackage{mathabx}
\usepackage{dbfloatfix}
\usepackage{anyfontsize}

\usepackage[pagebackref,breaklinks,colorlinks,linkcolor=blue]{hyperref}

\usepackage[capitalize]{cleveref}
\crefname{section}{Sec.}{Secs.}
\Crefname{section}{Section}{Sections}
\Crefname{table}{Table}{Tables}
\crefname{table}{Tab.}{Tabs.}

\begin{document}
\def\model{MaskGIT\xspace}
\def\bestfid{6.18}
\def\bestis{182.1}
\def\bestfidhighres{7.32}

\title{\model: Masked Generative Image Transformer}

\author{Huiwen Chang\quad Han Zhang\quad Lu Jiang\quad Ce Liu$^\ast$ \quad William T. Freeman\\
\vspace{1.3ex}
{Google Research}
}

\renewcommand*{\thefootnote}{\fnsymbol{footnote}}
\newcommand{\todo}[1]{\textcolor{red}{Todo:#1}}
\newcommand{\huiwen}[1]{\textcolor{red}{Huiwen: #1}}
\newcommand{\han}[1]{\textcolor{green}{Han: #1}}
\newcommand{\lu}[1]{\textcolor{blue}{Lu: #1}}

\interfootnotelinepenalty=1000000

\twocolumn[{%
\vspace{-5mm}
\maketitle

\begin{center}
    \newcommand{\teaserwidth}{\textwidth}
    \vspace{-8mm}
    \includegraphics[width=\teaserwidth]{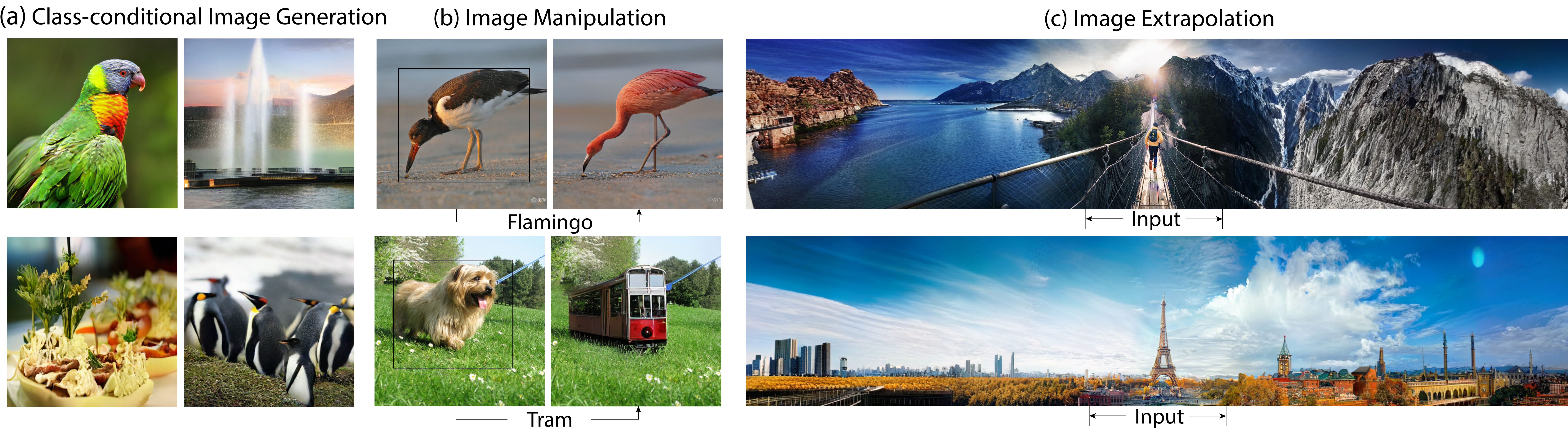}
    \vspace{-6mm}
    \captionof{figure}{\textbf{Example generation by MaskGIT on image synthesis and manipulation tasks}. We show that MaskGIT is a flexible model that can generate high-quality samples on (a) class-conditional synthesis, (b) class-conditional image manipulation, \eg replacing selected objects in the bounding box with ones from the given classes, and (c) image extrapolation. Examples shown here have resolutions 512$\times$512, 512$\times$512, and 512$\times$2560 in the three columns, respectively. Zoom in to see the details.} 
    \vspace{1mm}
	\label{fig:teaser}
\end{center}%
}]
\footnotetext{$^\ast$ Currently affiliated with Microsoft Azure AI.}

\begin{abstract}

Generative transformers have experienced rapid popularity growth in the computer vision community in synthesizing high-fidelity and high-resolution images. The best generative transformer models so far, however, still treat an image naively as a sequence of tokens, and decode an image sequentially following the raster scan ordering (i.e. line-by-line). We find this strategy neither optimal nor efficient. This paper proposes a novel image synthesis paradigm using a bidirectional transformer decoder, which we term MaskGIT. During training, MaskGIT learns to predict randomly masked tokens by attending to tokens in all directions. At inference time, the model begins with generating all tokens of an image simultaneously, and then refines the image iteratively conditioned on the previous generation. Our experiments demonstrate that MaskGIT significantly outperforms the state-of-the-art transformer model on the ImageNet dataset, and accelerates autoregressive decoding by up to 64x. Besides, we illustrate that MaskGIT can be easily extended to various image editing tasks, such as inpainting, extrapolation, and image manipulation.

\end{abstract}

\vspace{-6mm}

\section{Introduction}
\label{sec:intro}

\begin{figure*}[!ht]
    \centering
    \vspace{-3mm}
	\includegraphics[width=.9\textwidth]{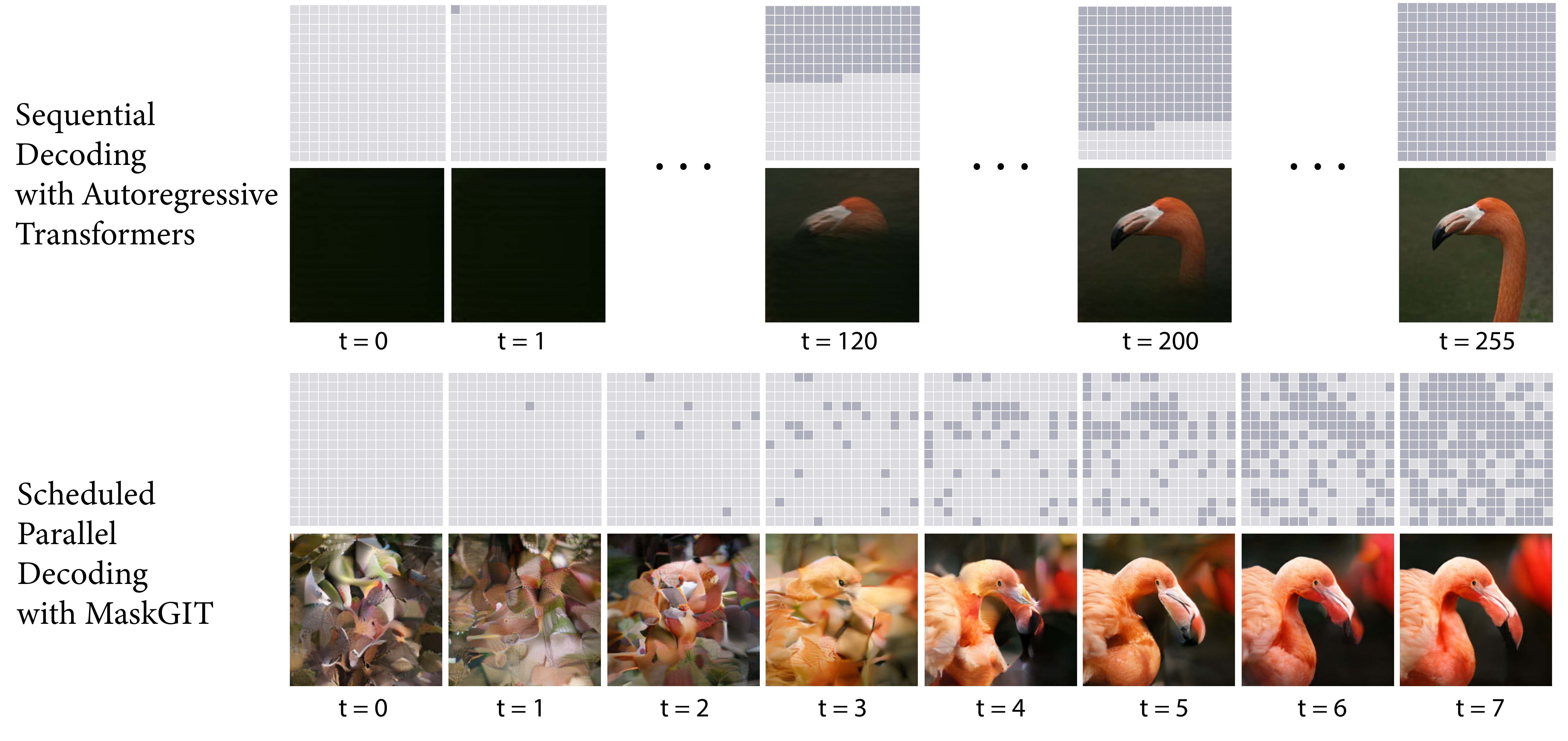}
    \vspace{-3mm}
    \caption{\textbf{Comparison between sequential decoding and \model's scheduled parallel decoding.} Rows 1 and 3 are the input latent masks at each iteration, and rows 2 and 4 are samples generated by each model at that iteration. Our decoding starts with all unknown codes (marked in lighter gray), and gradually fills up the latent representation with more and more scattered predictions in parallel (marked in darker gray), where the number of predicted tokens increases sharply over iterations. \model finishes its decoding in 8 iterations compared to the 256 rounds the sequential method takes.}
    \label{fig:decoding}
\vspace{-3mm}
\end{figure*}

Deep image synthesis as a field has seen a lot of progress in recent years. Currently holding state-of-the-art results are Generative Adversarial Networks (GANs), which are capable of synthesizing high-fidelity images at blazing speeds. They suffer from, however, well known issues include training instability and mode collapse, which lead to a lack of sample diversity. Addressing these issues still remains open research problems.

Inspired by the success of Transformer~\cite{Vaswani17attention} and GPT~\cite{gpt3} in NLP, generative transformer models have received  growing interests in image synthesis~\cite{chen2020imagegpt,Esser21vqgan,Razavi19vqvae2}. Generally, these approaches aim at modeling an image like a sequence and leveraging the existing autoregressive models to generate image. Images are generated in two stages; the first stage is to quantize an image to a sequence of discrete tokens (or visual words). In the second stage, an autoregressive model (e.g., transformer) is learned to generate image tokens sequentially based on the previously generated result (\ie autoregressive decoding). Unlike the subtle min-max optimization used in GANs, these models are learned by maximum likelihood estimation. Because of the design differences, existing works have demonstrated their advantages over GANs in offering stabilized training and improved distribution coverage or diversity.

Existing works on generative transformers mostly focus on the first stage, \ie how to quantize images such that information loss is minimized, and share the same second stage borrowed from NLP. Consequently, even the state-of-the-art generative transformers~\cite{Esser21vqgan,Ramesh21dalle} still treat an image naively as a sequence, where an image is flattened into a 1D sequence of tokens following a raster scan ordering, \ie from left to right line-by-line (\cf Figure~\ref{fig:decoding}). We find this representation neither optimal nor efficient for images. Unlike text, images are not sequential. Imagine how an artwork is created. A painter starts with a sketch and then progressively refines it by filling or tweaking the details, which is in clear contrast to the line-by-line printing used in previous work~\cite{chen2020imagegpt,Esser21vqgan}. Additionally, treating image as a flat sequence means that the autoregressive sequence length grows quadratically, easily forming an extremely long sequence--longer than any natural language sentence. This poses challenges for not only modeling long-term correlation but also renders the decoding intractable.  For example, it takes a considerable 30 seconds to generate a single image on a GPU autoregressively with 32x32 tokens.

This paper introduces a new bidirectional transformer for image synthesis called Masked Generative Image Transformer (\model). During training, \model is trained on a similar proxy task to the mask prediction in BERT~\cite{Devlin19bert}. At inference time, \model adopts a novel non-autoregressive decoding method to synthesize an image in constant number of steps. Specifically, at each iteration, the model predicts all tokens simultaneously in parallel but only keeps the most confident ones. The remaining tokens are masked out and will be re-predicted in the next iteration. The mask ratio is decreased until all tokens are generated with a few iterations of refinement. As illustrated in Figure~\ref{fig:decoding}, \model's decoding is an order-of-magnitude faster than the autoregresive decoding as it only takes 8 steps, instead of 256 steps, to generate an image and the predictions within each step are parallelizable. Moreover, instead of conditioning only on previous tokens in the order of raster scan, bidirectional self-attention allows the model to generate new tokens from generated tokens in all directions. We find that the mask scheduling (\ie fraction of the image masked each iteration) significantly affects generation quality. We propose to use the cosine schedule and substantiate its efficacy in the ablation study.

On the ImageNet benchmark, we empirically demonstrate that \model is both significantly faster (by up to 64x) and capable of generating higher quality samples than the state-of-the-art autoregressive transformer, \ie VQGAN, on class-conditional generation with 256$\times$256 and 512$\times$512 resolution. Even compared with the leading GAN model, \ie BigGAN, and diffusion model, \ie ADM~\cite{dhariwal2021diffusion}, \model offers comparable sample quality while yielding more favourable diversity. Notably, our model establishes new state-of-the-arts on classification accuracy score (CAS)~\cite{Ravuri19CAS} and on FID\cite{FID} for synthesizing 512$\times$512 images. To our knowledge, this paper provides the first evidence demonstrating the efficacy of the masked modeling for image generation on the common ImageNet benchmark.  

Furthermore, MaskGIT's multidirectional nature makes it readily extendable to image manipulation tasks that are otherwise difficult for autoregressive models. Fig.~\ref{fig:teaser} shows a new application of class-conditional image editing in which \model re-generates content inside the bounding box based on the given class while keeping the context (outside of the box) unchanged. This task, which is either infeasible for autoregressive model or difficult for GAN models, is trivial for our model.
Quantitatively, we demonstrate this flexibility by applying \model to image inpainting, and image extrapolation in arbitrary directions.
%shown in Fig.~\ref{fig:teaser}
Even though our model is not designed for such tasks, it obtains comparable performance to the dedicated models on each task.

\section{Related Work}
\label{sec:related}
\subsection{Image Synthesis}\label{sec:related_synthesis}
Deep generative models~\cite{KingmaW14, Vahdat2020nvae, goodfellow2014generative, SAGAN, Song19generative,dhariwal2021diffusion,Oord16pixelcnn, Parmar18imagetransformer} have achieved lots of successes in image synthesis tasks. GAN based methods demonstrate amazing capability in yielding high-fidelity samples~\cite{goodfellow2014generative, biggan, Karras2019stylegan2, SAGAN, tseng2021regularizing}.
In contrast, likelihood-based methods, such as Variational Autoencoders (VAEs) \cite{KingmaW14, Vahdat2020nvae}, Diffusion Models \cite{Song19generative,dhariwal2021diffusion,ho2021cascaded} and Autoregressive Models \cite{Oord16pixelcnn, Parmar18imagetransformer}, offer distribution coverage and hence can generate more diverse samples~\cite{Song19generative, Vahdat2020nvae, Oord16pixelcnn}. 

However, maximizing likelihood directly in pixel space can be challenging. So instead,
VQVAE \cite{Oord17vqvae, Razavi19vqvae2} proposes to generate images in latent space in two stages. In the first stage, which is known as \textbf{tokenization}, it tries to compress images into discrete latent space, and primarily consists of three components:\vspace{-2.5mm}
\begin{itemize}
    \item an encoder $E$ that learns to tokenize images $x \in \mathbb{R}^{H\times W \times 3}$ into latent embedding $E(x)$, \vspace{-2.5mm}
    \item a codebook $\mathbf{e}_k \in \mathbb{R}^D, k\in 1, 2, \cdots, K$ which serves for a nearest neighbor look up used to quantize the embedding into visual tokens, and  \vspace{-2.5mm}
    \item a decoder $G$ which predicts the reconstructed image $\hat{x}$ from the visual tokens $\mathbf{e}$. 
\end{itemize}\vspace{-1mm}
%  The training objective of the first stage is 
% \begin{equation} \label{eq:1}
% \begin{split}
%  \mathcal{L}_{VQ} = {} & ||sg[E(x)] -\mathbf{e}||_2^2 + \beta ||sg[\mathbf{e}] - E(x)||_2^2 \\
%   &+ \mathcal{L}_{rec}(x,\hat{x}).
% \end{split}
% \end{equation}
% Here the terms represent the codebook loss penalizing token $\mathbf{e}$ far from embedding $E(x)$, the commitment loss constraining $E$ to stick to one token in the codebook, and the reconstruction loss between input and reconstruction. $sg$ is the stop-gradient operator and $\beta$ is a commitment loss hyper-parameter. VQVAE first introduces this training objective and uses $L_2$ for the reconstruction loss. Later VQGAN proposes the additions of the perceptual loss and the adversarial loss to achieve higher compression efficiency.

\noindent In the second stage, it first predicts the latent priors of the visual tokens using deep autoregressive models, and then uses the decoder from the first stage to map the token sequences into image pixels. Several approaches have followed this paradigm due to the efficacy of the two-stage approach. DALL-E \cite{Ramesh21dalle} uses Transformers \cite{Vaswani17attention} to improve token prediction in the second stage. VQGAN~\cite{Esser21vqgan} adds adversarial loss and perceptual loss~\cite{johnson2016perceptual, zhang2018unreasonable} in the first stage to improve the image fidelity. A contemporary work to ours, VIM~\cite{vim2021}, proposes to use a VIT backbone ~\cite{dosovitskiy2021vit} to further improve the tokenization stage. Since these approaches still employ an auto-regressive model, the decoding time in the second stage scales with the token sequence length. 

\subsection{Masked Modeling with Bi-directional Transformers}
The transformer architecture~\cite{Vaswani17attention}, was first proposed in NLP, and has recently extended its reach to computer vision~\cite{dosovitskiy2021vit, caron2021dino}.
Transformer consists of multiple self-attention layers, allowing interactions between all pairs of elements in the sequence to be captured.
In particular, BERT~\cite{Devlin19bert} introduces the masked language modeling (MLM) task for language representation learning. The bi-directional self-attention used in BERT~\cite{Devlin19bert} allows the masked tokens in MLM to be predicted utilizing context from both directions.
In vision, the masked modeling in BERT~\cite{Devlin19bert} has been extended to image representation learning~\cite{he2021mae, Bao2022Beit} with images quantized to discrete tokens. However, few works have successfully applied the same masked modeling to image generation~\cite{zhang2021ufcbert} because of the difficulty in performing autoregressive decoding using bi-directional attentions.
To our knowledge, this paper provides the first evidence demonstrating the efficacy of masked modeling for image generation on the common ImageNet benchmark.
Our work is inspired by bi-directional machine translation~\cite{ghazvininejad2019maskpredict, gu2020fully, gu2017non} in NLP, and our novelty lies in the proposed new masking strategy and decoding algorithm which, as substantiated by our experiments, are essential for image generation.

\begin{figure}[!t]
    \centering
	\includegraphics[width=\linewidth]{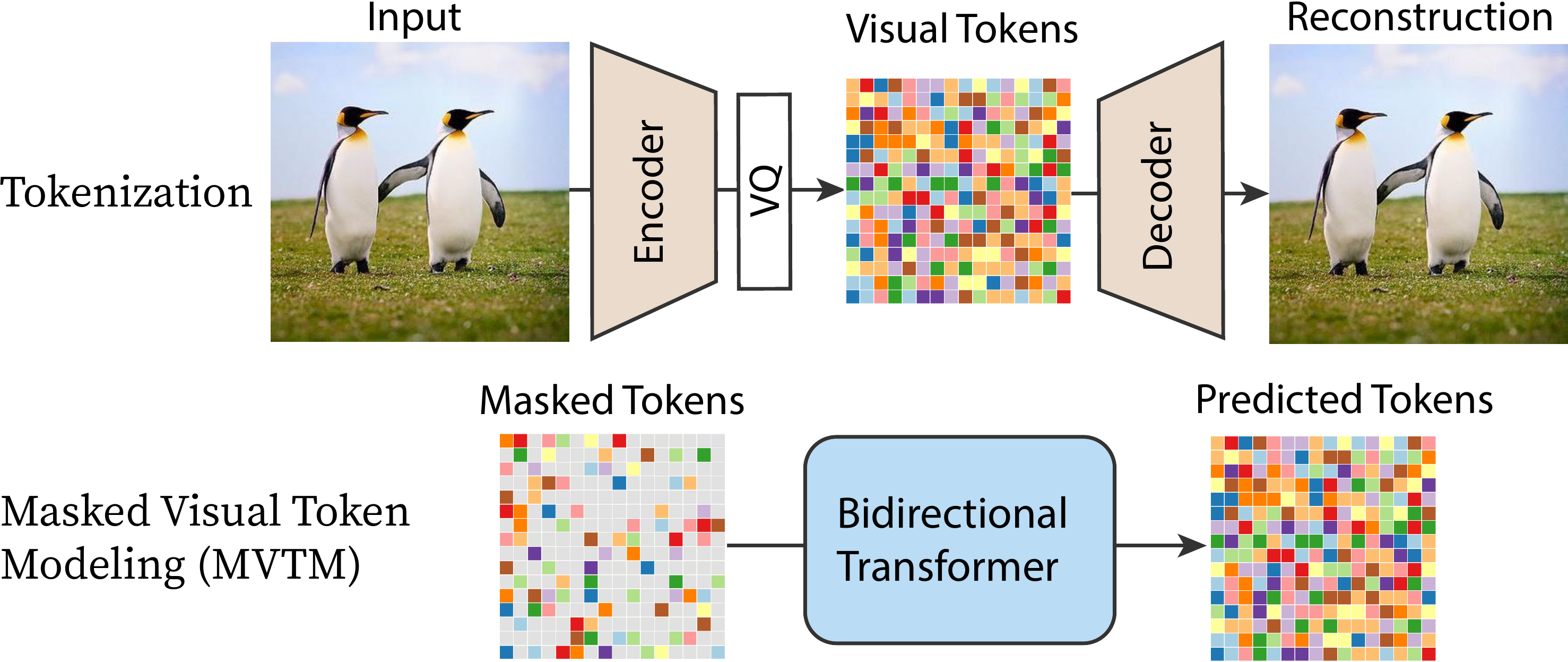}
    % \vspace{-25mm}
    \caption{\textbf{Pipeline Overview.} \model follows a two-stage design, with 1) a tokenizer that tokenizes images into visual tokens, and 2) a bidirectional tranformer model that performs MVTM, i.e. learns to predict visual tokens masked at random.}
    \vspace{-3mm}
    \label{fig:pipeline}
\end{figure}
\section{Method}
\label{sec:method}
Our goal is to design a new image synthesis paradigm utilizing parallel decoding and bi-directional generation.

We follow the two-stage recipe discussed in \ref{sec:related_synthesis}, as illustrated in Figure~\ref{fig:pipeline}. Since our goal is to improve the second stage, we employ the same setup for the first stage as in the VQGAN model~\cite{Esser21vqgan}, and leave potential improvements to the tokenization step to future work.

For the second stage, we propose to learn a bidirectional transformer by \textit{Masked Visual Token Modeling} (MVTM). We introduce MVTM training in \ref{ssec:model} and the sampling procedure in \ref{ssec:decoding}. We then discuss the key technique of masking design in \ref{ssec:masking}.

\subsection{MVTM in Training}
\label{ssec:model}
\newcommand{\unknownmask}[0]{{\mathbf{M}}}
\newcommand{\mathy}[0]{{\mathbf{Y}}}
\newcommand{\knownmask}[0]{{\overline{\mathbf{M}}}}
Let $\mathy=[y_i]_{i=1}^N$ denote the latent tokens obtained by inputting the image to the VQ-encoder, where $N$ is the length of the reshaped token matrix, and $\unknownmask=[m_i]_{i=1}^N$ the corresponding binary mask. During training, we sample a subset of tokens and replace them with a special \texttt{[MASK]} token. The token $y_i$ is replaced with \texttt{[MASK]} if $m_i=1$, otherwise, when $m_i=0$, $y_i$ will be left intact. 

The sampling procedure is parameterized by a mask scheduling function $\gamma (r) \in (0,1]$, and executes as follows: we first sample a ratio from $0$ to $1$, then uniformly select $\lceil \gamma(r) \cdot N \rceil$ tokens in $\mathy$ to place masks, where $N$ is the length. The mask scheduling significantly affects the quality of image generation and will be discussed in \ref{ssec:masking}.

Denote $Y_{\knownmask}$ the result after applying mask $\unknownmask$ to $\mathy$.
The training objective is to minimize the negative log-likelihood of the masked tokens:
\begin{equation}
\label{eq:loss}
\mathcal{L}_{\text{mask}} = - \mathop{\mathbb{E}} \limits_{\mathy  \in \mathcal{D}} \Big[ \sum_{\forall i \in [1,N], m_i=1} \log p(y_i| Y_{\knownmask}) \Big],
\end{equation}

\noindent Concretely, we feed the masked $Y_{\knownmask}$ into a multi-layer bidirectional transformer to predict the probabilities $P(y_i | Y_{\knownmask} )$ for each masked token, where the negative log-likelihood is computed as the cross-entropy between the ground-truth one-hot token and predicted token.
Notice the key difference to autoregressive modeling: the conditional dependency in MVTM has two directions, which allows image generation to utilize richer contexts by attending to all tokens in the image.

\subsection{Iterative Decoding}
\label{ssec:decoding}
In autoregressive decoding, tokens are generated sequentially based on previously generated output. This process is not parallelizable and thus very slow for image because the image token length, \eg 256 or 1024, is typically much larger than that of language. We introduce a novel decoding method where all tokens in the image are generated simultaneously in parallel. This is feasible due to the bi-directional self-attention of MTVM. 

In theory, our model is able to infer all tokens and generate the entire image in a single pass. We find this challenging due to inconsistency with the training task. Below, the proposed iterative decoding is introduced.
To generate an image at inference time, we start from a blank canvas with all the tokens masked out, \ie $Y_\unknownmask^{(0)}$. For iteration $t$, our algorithm runs as follows: 
 
 \vspace{-2mm}
 \begin{enumerate}
     \item \textbf{Predict.}  Given the masked tokens $Y_\unknownmask^{(t)}$ at the current iteration, our model predicts the probabilities, denoted as $p^{(t)} \in \mathbb{R}^{N \times K}$, for all the masked locations in parallel.  \\ \vspace{-5mm}
     \item \textbf{Sample.} At each masked location $i$, we sample a token $y_i^{(t)}$ based on its prediction probabilities $p_i^{(t)} \in \mathbb{R}^K$ over all possible tokens in the codebook. After a token $y_i^{(t)}$ is sampled, its corresponding prediction score is used as a ``confidence" score indicating the model's belief of this prediction. 
     For the unmasked position in $Y_\unknownmask^{(t)}$, we simply set its confidence score to $1.0$. \\ \vspace{-5mm}
    \item \textbf{Mask Schedule.} We compute the number of tokens to mask according to the mask scheduling function $\gamma$ by $n = \lceil \gamma(\frac{t}{T}) N \rceil$, where $N$ is the input length and $T$ is the total number of iterations.
    \item \textbf{Mask.} We obtain $Y_\unknownmask^{(t+1)}$ by masking $n$ tokens in $Y_\unknownmask^{(t)}$. The mask $\unknownmask^{(t+1)}$ for iteration $t+1$ is calculated from:\vspace{-2mm}
        \begin{equation*}
         m_{i}^{(t+1)} = 
        \begin{cases}
            1,  & \text{if $c_i < {\text{sorted}}_j(c_j)[n]$.}\\
            0,  & \text{otherwise.}\vspace{-2mm}
        \end{cases},
        \end{equation*}
    where $c_i$ is the confidence score for the $i$-th token.

 \end{enumerate}

The decoding algorithm synthesizes an image in $T$ steps. At each iteration, the model predicts all tokens simultaneously but only keeps the most confident ones. The remaining tokens are masked out and re-predicted in the next iteration. The mask ratio is made decreasing until all tokens are generated within $T$ iterations. In practice, the masking tokens are randomly sampled with temperature annealing to encourage more diversity, and we will discuss its effect in \ref{ssec:ablation}. Figure~\ref{fig:decoding} illustrates an example of our decoding process. It generates an image in $T=8$ iterations, where the unmasked tokens at each iteration are highlighted in the grid, \eg when $t=1$ we only keep 1 token and mask out the rest.

\subsection{Masking Design}
\label{ssec:masking}

We find that the quality of image generation is significantly affected by the masking design. We model the masking procedure by a mask scheduling function $\gamma (\cdot)$ that computes the mask ratio for the given latent tokens. As discussed, the function $\gamma$ is used in both training and inference. During inference time, it takes the input of $0/T, 1/T, \cdots, (T-1)/T$ indicating the progress in decoding. In training, we randomly sample a ratio $r$ in $[0,1)$ to simulate the various decoding scenarios.

BERT uses a fixed mask ratio of 15\%~\cite{Devlin19bert}, \ie, it always masks 15\% of the tokens, which is unsuitable for our task since our decoder needs to generate images from scratch. New masking scheduling is thus needed. Before discussing specific schemes, we first examine the property of the mask scheduling function. First, $\gamma(r)$ needs to be a continuous function bounded between $0$ and $1$ for $r \in [0,1]$. Second, $\gamma(r)$ should be (monotonically) decreasing with respect to $r$, and it holds that $\gamma(0)\rightarrow1$ and $\gamma(1)\rightarrow0$. 
The second property ensures the convergence of our decoding algorithm.

This paper considers common functions and makes simple transformations so that they satisfy the properties. Figure~\ref{fig:scheduling} visualizes these functions which are divided into three groups:
\begin{itemize}
    \vspace{-2mm}
    \item \textbf{Linear function} is a straightforward solution, which masks an equal amount of tokens each time.
    \vspace{-2mm}
    \item \textbf{Concave function} captures the intuition that image generation follows a less-to-more information flow. In the beginning, most tokens are masked so the model only needs to make a few correct predictions for which the model feel confident. Towards the end, the mask ratio sharply drops, forcing the model to make a lot more correct predictions. The effective information is increasing in this process. The concave family includes cosine, square, cubic, and exponential.
    \vspace{-2mm}
    \item \textbf{Convex function}, conversely, implements a more-to-less process. The model needs to finalize a vast majority of tokens within the first couple of iterations. This family includes square root and logarithmic.
    
    \vspace{-2mm}
\end{itemize}

We empirically compare the above mask scheduling functions in \ref{ssec:ablation} and find the $\text{cosine}$ function works the best in all of our experiments.

\section{Experiments}
\label{sec:experiments}

In this section, we empirically evaluate \model on image generation in terms of quality, efficiency and flexibility.
In \ref{ssec:class_conditional_synthesis}, we evaluate \model on the standard class-conditional image generation tasks on ImageNet~\cite{deng2009imagenet} 256$\times$256 and 512$\times$512.
In~\ref{ssec:applications}, we show \model's versatility by demonstrating its performance on three image editing tasks, image inpainting, outpainting, and editing. In~\ref{ssec:ablation}, we verify the necessity of our design of mask scheduling. We will release the code and model for reproducible research.

\subsection{Experimental Setup}
\label{ssec:architecture}

For each dataset, we only train a single autoencoder, decoder, and codebook with 1024 tokens on cropped 256x256 images for all the experiments. The image is always compressed by a fixed factor of 16, \ie from $H\times W$ to a grid of tokens in the size of $h \times w$, where $h$=$H / 16$ and $w$=$W/16$. We find that this autoencoder, together with the codebook, can be reused to synthesize 512$\times$512 images.

All models in this work have the same configuration: 24 layers, 8 attention heads, 768 embedding dimensions and 3072 hidden dimensions. Our models use learnable positional embedding\cite{Vaswani17attention}, LayerNorm\cite{ba2016layer}, and truncated normal initialization (stddev=$0.02$). We employ the following training hyperparameters: label smoothing=$0.1$, dropout rate=$0.1$, Adam optimizer~\cite{kingma2014adam} with $\beta_{1}$=$0.9$ and $\beta_{2}$=$0.96$. We use RandomResizeAndCrop for data augmentation. All models are trained on 4x4 TPU devices with a batch size of 256. ImageNet models are trained for 300 epochs while the Places2 model is trained for 200 epochs. 

\subsection{Class-conditional Image Synthesis}
\label{ssec:class_conditional_synthesis}

We evaluate the performance of our model on class-conditional image synthesis on ImageNet 256$\times$256 and 512$\times$512. Our main results are summarized in Table~\ref{tab:maintable}.

\noindent\textbf{Quality.}
On ImageNet 256$\times$256, without any special sampling strategies such as beam-search, top-k or nucleus sampling heuristics~\cite{holtzman2019nucleus} or classifier guidance~\cite{Razavi19vqvae2}, we significantly outperform VQGAN~\cite{Esser21vqgan} in both Fr\'{e}chet Inception Distance (FID)~\cite{FID} ($\bestfid$ vs $15.78$) and Inception Score (IS) ($\bestis$ vs $78.3$). We also report the results with classifier-based rejection sampling in the appendix \ref{sec:supp_diversity_comparison}. 

We also train a VQGAN baseline with the same tokenizer and hyperparameters as \model's in order to further highlight the difference between bi-directional and uni-directional transformers, and find that on both resolutions, \model still outperforms our implemented baseline by a significant margin. 

Furthermore, \model improves BigGAN's FIDs on both resolutions, achieving a new state-of-the-art on 512$\times$512 with an FID of $7.32$.

\begin{figure}[!t]
	\centering
	\includegraphics[width=1.\linewidth]{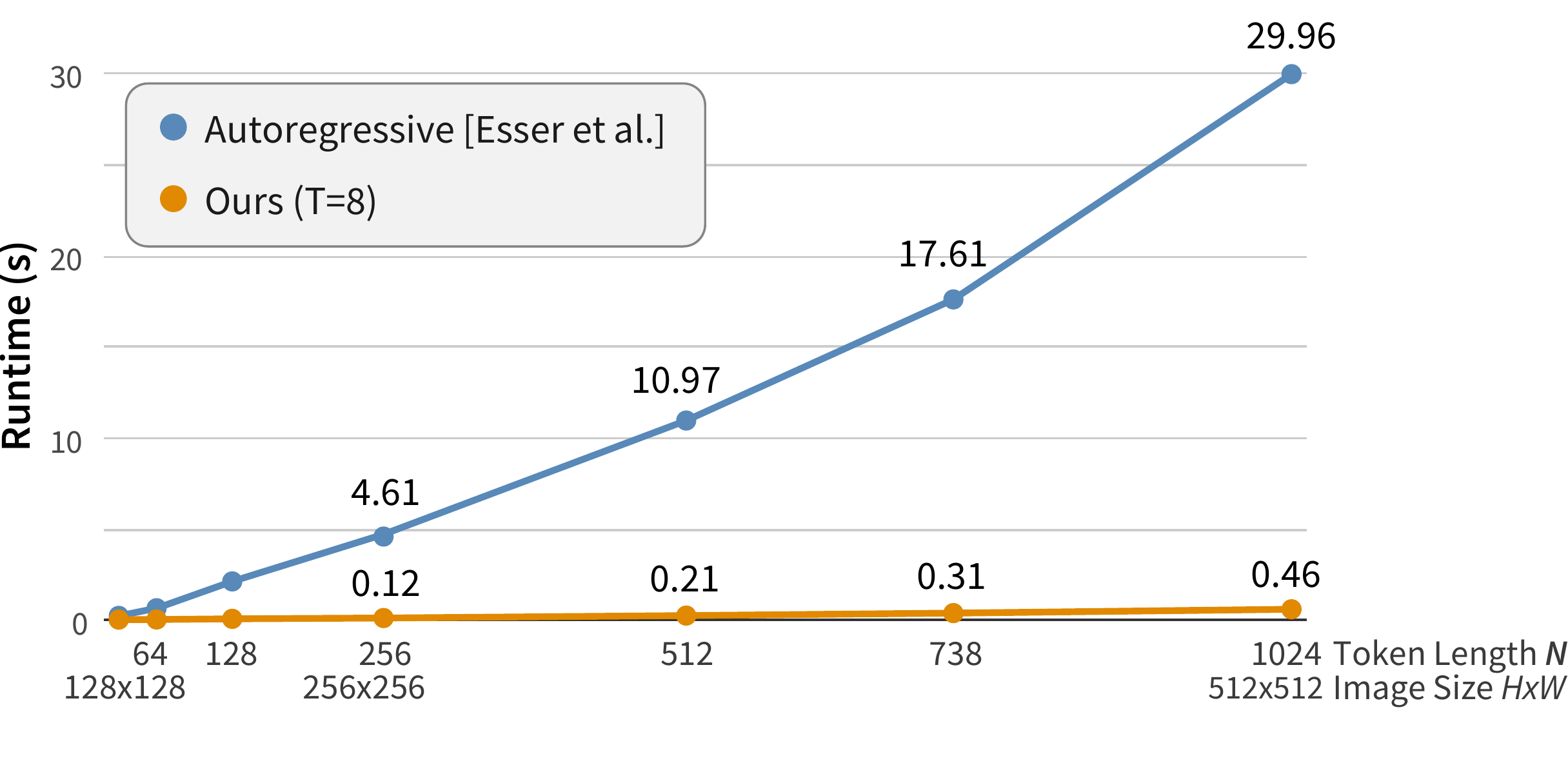}
	\vspace{-5mm}
	\caption{\textbf{Transformer wall-clock runtime comparison between VQGAN\cite{Esser21vqgan} and ours.} All results are run on a single GPU. }
	\label{fig:speed}
 	\vspace{-4mm}
\end{figure}
{
\newcolumntype{N}{@{}m{0pt}@{}}

\begin{table*}[h]
\small
    \centering
    {\small
    \begin{tabular}{lc ccc ccc ccc cc}
    \toprule
     {Model} & 
    
     & {FID}  $\downarrow$ & {IS}  $\uparrow$ & 
     &  {Prec}  $\uparrow$ & {Rec}   $\uparrow$ &
     
     &{\# params} & {\# steps} &
     
     & \multicolumn{2}{c}{{CAS $\times 100$} $\uparrow$}  
     \\
     \cline{12-13}
    &&  && & &&& & && {Top-1 (76.6)} & {Top-5 (93.1)} \\[-7pt]
    \textbf{ImageNet 256$\times$256} && &&& &&& &&& &
    \\[-2pt]
    \midrule
    
    DCTransformer~\cite{nash2021generating} $^\square$ &
    & 36.51 & n/a &
    & 0.36 & \textbf{0.67} &
    & 738M & $>$1024 &   & &
     \\ 
    
    BigGAN-deep~\cite{biggan}  & %, 1.0 trunc. &
    & 6.95 & \bfseries{198.2} &
    & \textbf{0.87} & 0.28 &
    & 160M & 1 & & 43.99 &67.89
    \\
    
    % Improved DDPM~\cite{nichol2021improved} &
    % & 31.5 &   &
    % & ?0.70 & ?0.62 &
    % &  & &
    % & 270M & 250 \\
    Improved DDPM~\cite{nichol2021improved}$^\square$  &
    & 12.26 & n/a  &
    & 0.70 & 0.62 &
     & 280M & 250  &  &  & \\
    
    ADM~\cite{dhariwal2021diffusion}$^\square$   & 
    & 10.94 & 101.0  &
    & 0.69 & 0.63 &
    & 554M & 250 &  &  & \\
    
    % ADM-G, 1.0 guid. & 
    % & 4.59 & 186.7  &
    % & ?0.69 & ?0.63 &
    % &  &  &
    % & 554M & 250  %554M + 54M classfier
    % \\

    VQVAE-2~\cite{Razavi19vqvae2}$^\square$ &
    & 31.11 & $\sim$45 &
    & 0.36 & 0.57 &
    & 13.5B$^\dagger$ & 5120 & & 54.83 & 77.59 \\
   % \midrule
   % \bfseries{\model Baseline} && 18.98 & 192.52 && ? & ? && ?? & ?? \\
    \midrule
    VQGAN~\cite{Esser21vqgan}$^\square$ &
    
    & 15.78 & 78.3 &
    & n/a & n/a &
    & 1.4B & 256 &  &  &
    \\
 
    VQGAN$^\ast$ &
    & 18.65 & 80.4 &
    & 0.78 & 0.26 &
    & 227M & 256 & 
    & 53.10 & 76.18
     \\   
    % VQ-GAN, 0.05 RS &
    % & 1.4B +  & 5120 &
    % & 5.20 & 380.3 &
    % % & ? & ? &
    % &  & \\
%     \bfseries{\model-Linear} &
    
%     &  &  &
%     &  &  &
%   & 227M  & 16 &
%      & &
   
    \bfseries{\model (Ours)} &
    
    & \bfseries{\bestfid} & \bestis &
    & 0.80 & 0.51 &
   & 227M  & \textit{8} &
   & \bfseries{63.14} & \bfseries{84.45}
    \\
%    & & \textbf{5.14} & 184.0 & &  0.78 & 0.52 & & 227M & \textit{64} &
%    & &  \\
    % 227M + 54M
    % \bfseries{Ours, 0.05 RS} &
    % & +  &  &
    % & 5.90 & 152.9 &
    % % & ? & ? &
    % &  &  \\

    % Validation &
    % &- & - &
    % & 1.62 & 234.0  &
    % % & ? & ? &
    % &  73.09 &  91.47 \\
    \bottomrule
    \\[-4pt]
    \textbf{ImageNet 512$\times$512} && &&& &&& &&& &\\[-2pt]
    \midrule
    
     BigGAN-deep~\cite{biggan} & & 8.43 & \bfseries{232.5} & & \textbf{0.88} & 0.29 &
     &160M&1& 
     &44.02&68.22\\
     
     ADM~\cite{dhariwal2021diffusion}$^\square$ & & 23.24 & 58.06  &&0.73& \textbf{0.60} & 
     &559M&250&
     && \\
    %  ADM, 1.0 guid. & 7.72 & & 0.87 & 0.42  \\
     \midrule
     VQGAN$^\ast$ & & 26.52 & 66.8 &
     & 0.73 & 0.31 &
      & 227M &1024 &
     &51.29&74.24 \\
     \bfseries{\model (Ours)} & & \bfseries{\bestfidhighres} & 156.0 & &  0.78 & 0.50 & & 227M &\textit{12}&
     &\textbf{63.43} & \textbf{84.79}  \\
    %  & & \bfseries{5.94} & 169.4 & & 0.77 & 0.48 & & 227M & \textit{256} &
    %  & &  \\
     % \bfseries{Ours}, 0.5 RS & ? & ? & &  \\
    \bottomrule
    
    \end{tabular}
    }
    \vspace{-5pt}
    \caption{Quantitative comparison with state-of-the-art generative models on ImageNet 256$\times$256 and 512$\times$512. 
    \footnotesize{``\# steps” refers to the number of neural network runs needed to generate a sample.  $^\ast$ denotes the model we train with the same architecture and setup with ours; $^\square$ denotes values taken from prior publications;  $^\dagger$ estimated based on the pytorch implementation~\cite{pytorch2020vqvae2}.} 
    }
    % $^\blackdiamond$ indicates classifiers trained with data augmentation Randaugment~\cite{cubuk2019randaugment};
    \label{tab:maintable}
    % On average 10-15ms per step

\end{table*}

% \begin{table}[h]
% \small
%     \centering
%     {\small
%     \begin{tabular}{lcccc}
%     \toprule    
%     \bfseries{Model} & \bfseries{FID}  $\downarrow$ & \bfseries{IS}  $\uparrow$ & \bfseries{Prec} $\uparrow$  & \bfseries{Rec} $\uparrow$ \\ \midrule
    %  BigGAN-deep~\cite{biggan} & 8.43 & \bfseries{232.5} & 0.88 & 0.29 \\
    %  ADM~\cite{dhariwal2021diffusion} & 23.24 & &0.73& 0.60  \\
    % %  ADM, 1.0 guid. & 7.72 & & 0.87 & 0.42  \\
    % % \midrule
    %  VQGAN$^\ast$ & 25.36 & 102.8 &&  \\
    %  \bfseries{Ours} & \bfseries{\bestfidhighres} & 156 & &  \\
    %  % \bfseries{Ours}, 0.5 RS & ? & ? & &  \\
    % \bottomrule
%     \end{tabular}
%     }
%     % \vspace*{0.35cm}
%     \caption{Quantitative comparison with state-of-the-art generative models on ImageNet $512 \times 512$. $^\ast$ denotes models we train with the same architecture and setup with ours. 
%     \todo{rerun with the same codebook} }
%     \label{tab:imagenet512}
% \vspace{-.2cm}
% \end{table}

% \begin{table}[h]
% \small
%     \centering
%     {\small
%     \begin{tabular}{lc ccc ccc cc}
%     \toprule
%  Reconstruction &
%     &  & &
%     & 2.64 & 191.4 &
%     % & ? & ? &
%     & ? & ? \\
%         \bottomrule
%     \end{tabular}
%     }
%     \caption{Reconstruction}
%     \label{tab:reconstruction}
% \vspace{-.2cm}
% \end{table}
\newcommand{\tmpwidth}{0.31\linewidth}

\begin{figure*}[!ht]
    \centering
    \begin{tabular}{c c c}
     \includegraphics[ width=\tmpwidth]{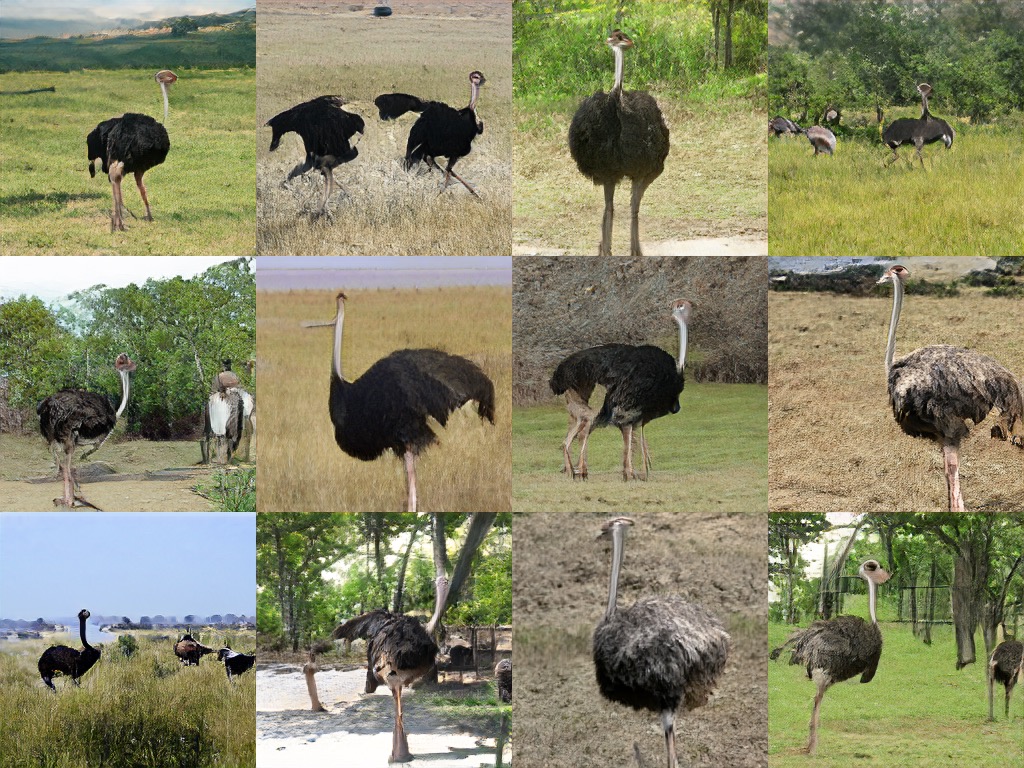} & \hspace{-3mm}
    \includegraphics[ width=\tmpwidth]{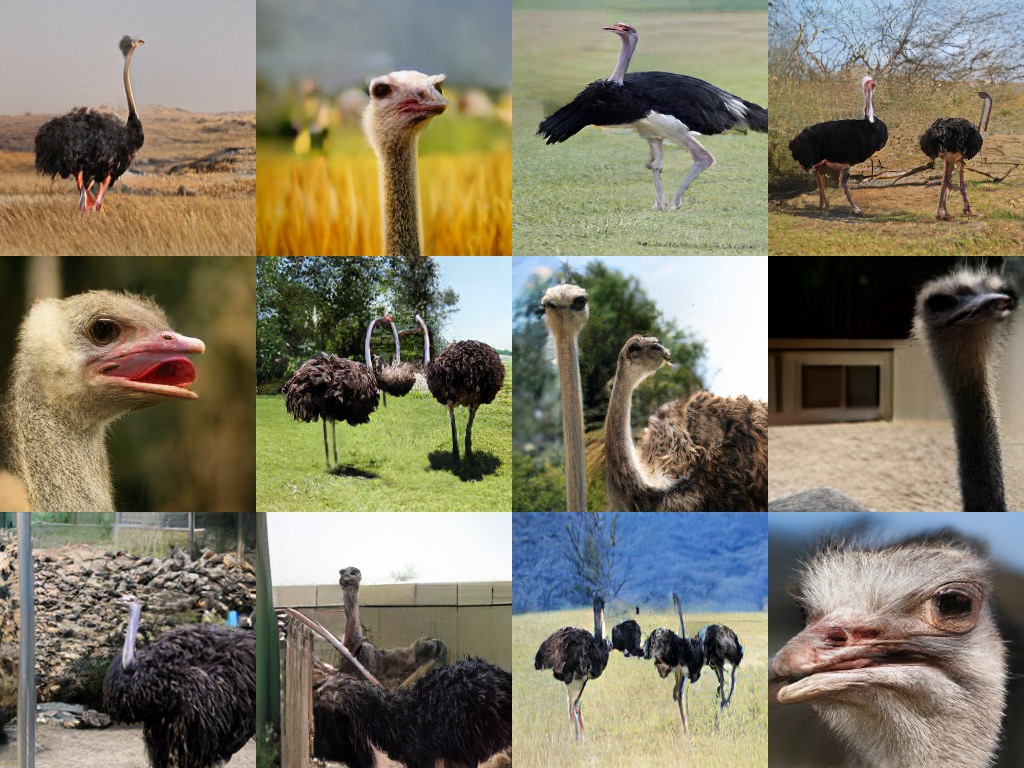} & \hspace{-3mm}
    \includegraphics[ width=\tmpwidth]{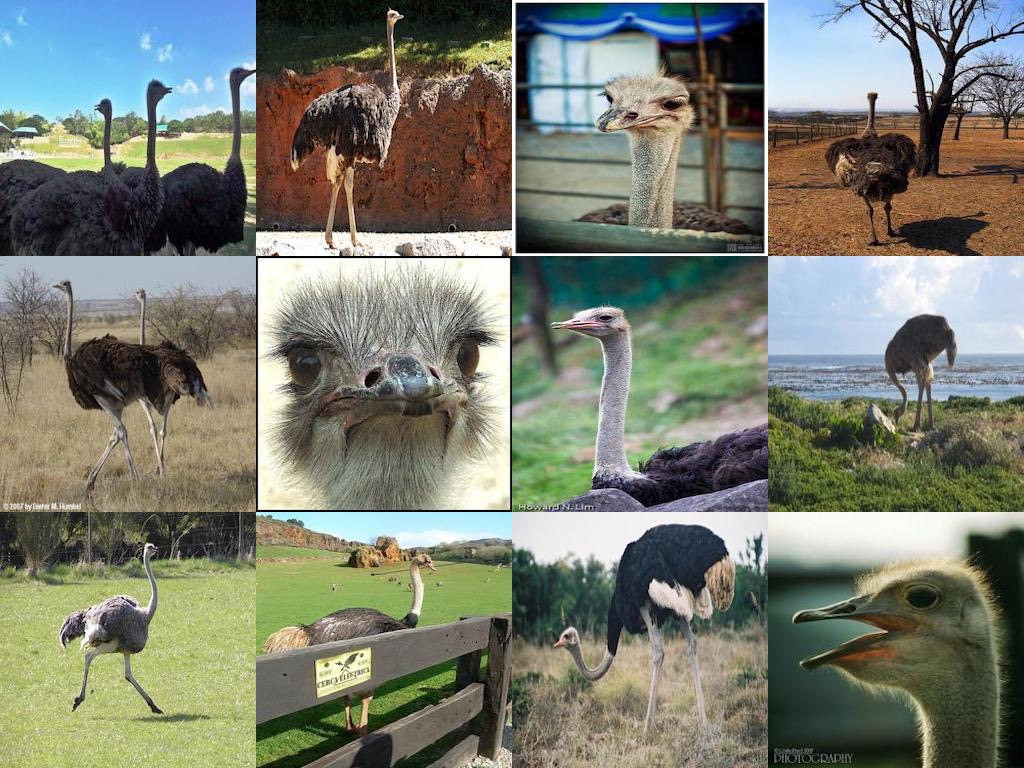} \\

    \includegraphics[ width=\tmpwidth]{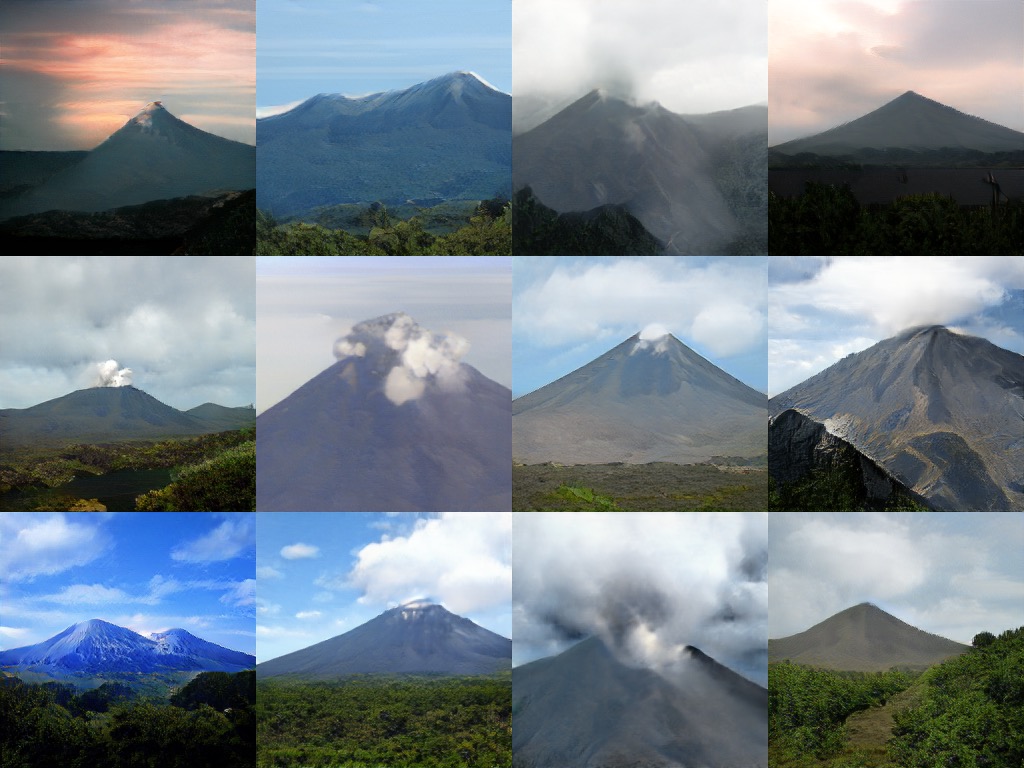} & \hspace{-3mm}
    \includegraphics[ width=\tmpwidth]{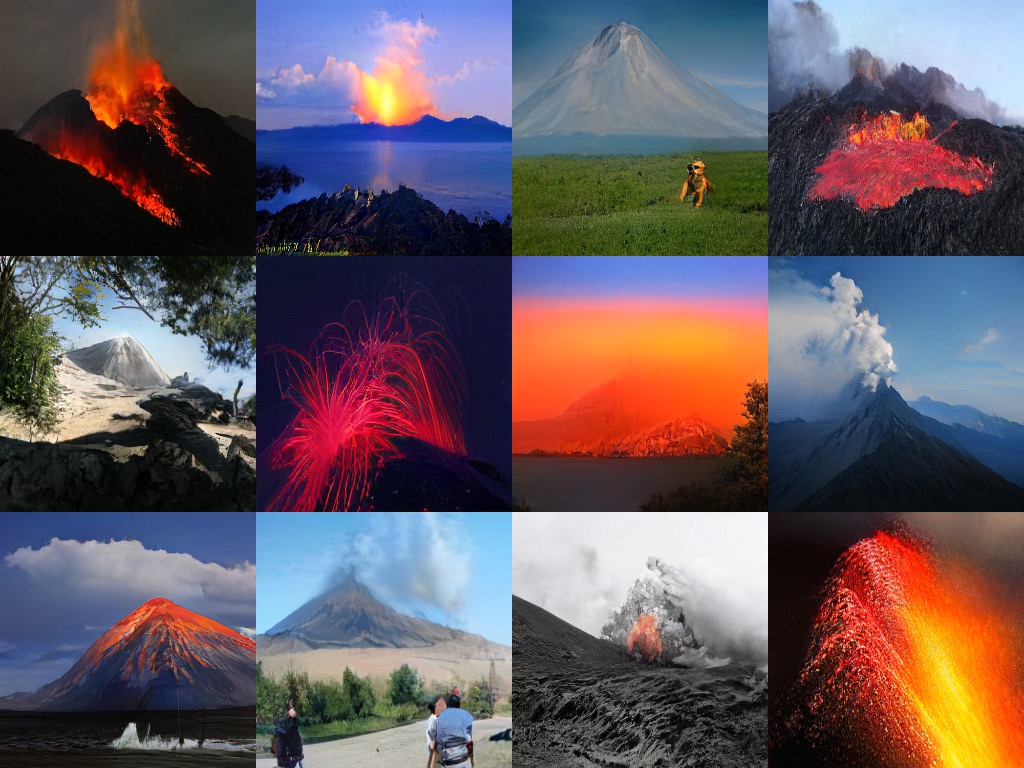} & \hspace{-3mm}
    \includegraphics[ width=\tmpwidth]{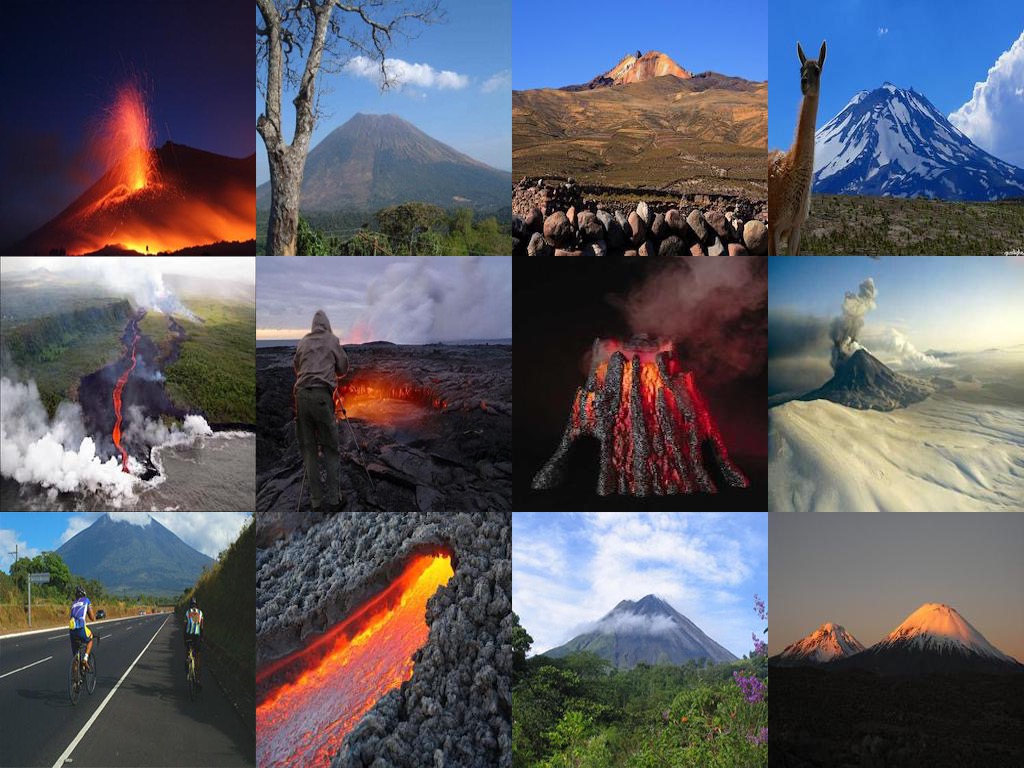} \\

    \includegraphics[ width=\tmpwidth]{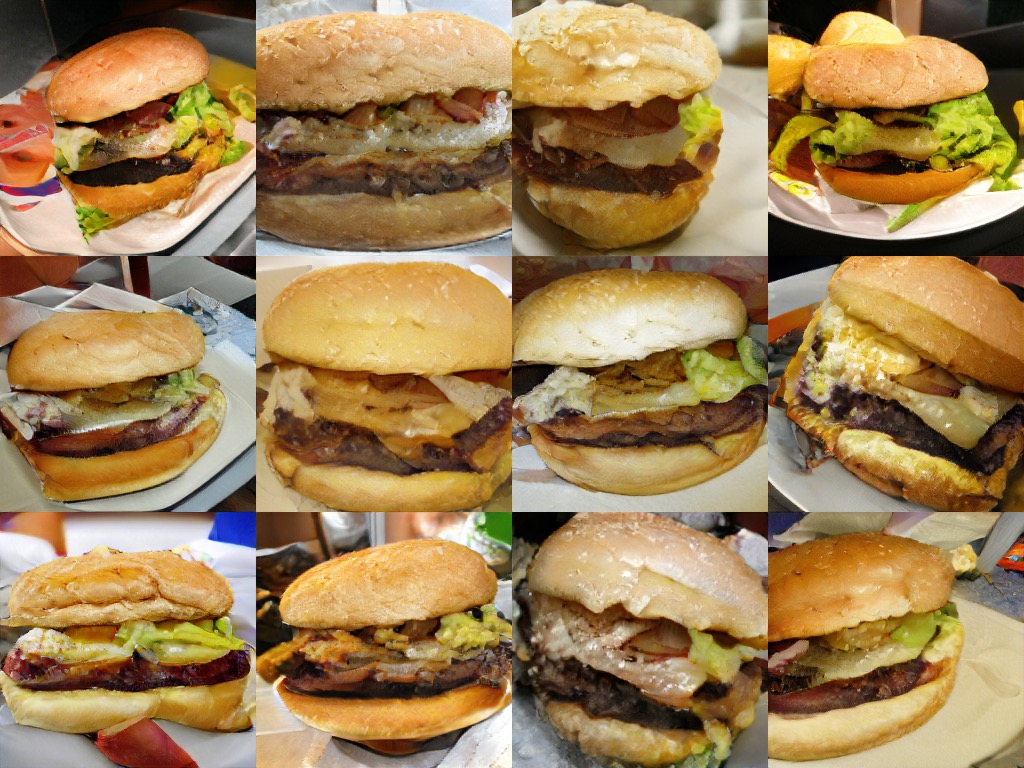} & \hspace{-3mm}
    \includegraphics[ width=\tmpwidth]{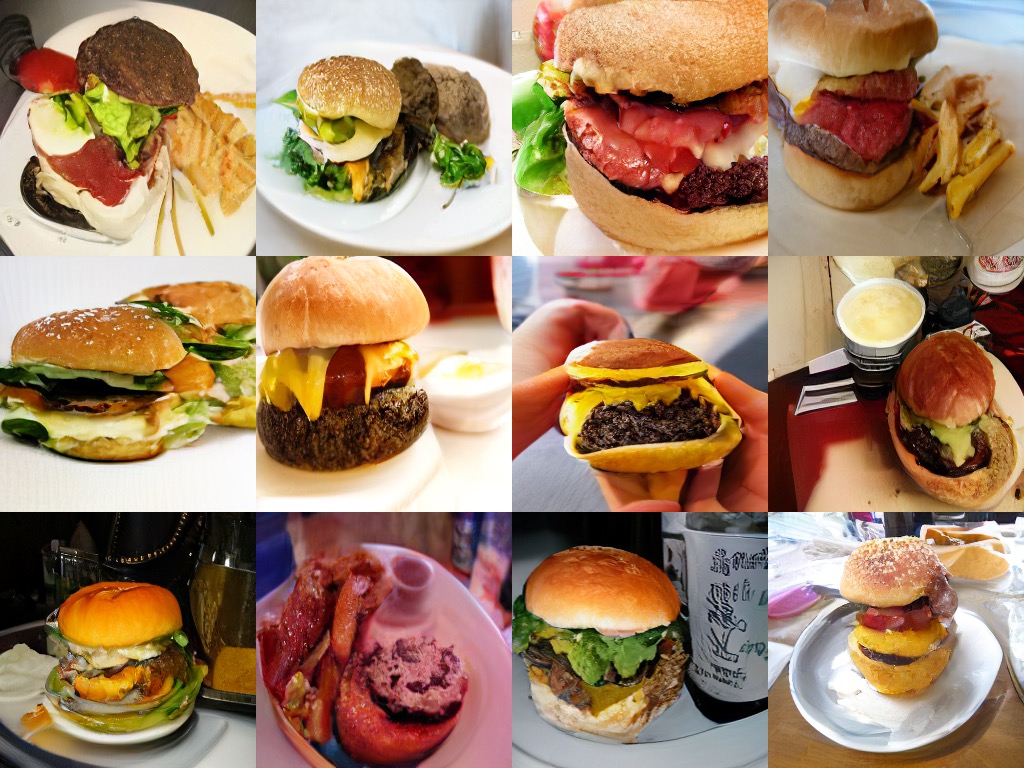} & \hspace{-3mm}
    \includegraphics[ width=\tmpwidth]{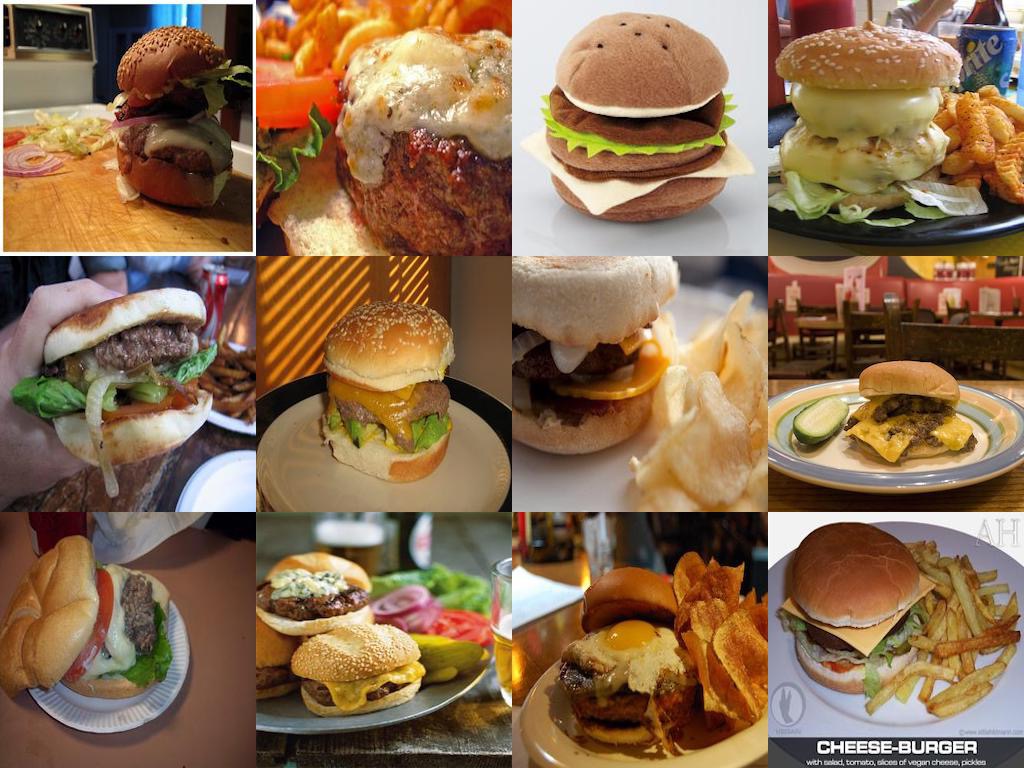} \\

    BigGAN-deep (FID=$6.95$) & \model (FID=\bestfid) & Training Set \\
    \end{tabular}
    \vspace{-2mm}
    \caption{Sample Diversity Comparison between our proposed method \model and BigGAN-deep~\cite{biggan} on ImageNet 256$\times$256. 
    %BigGAN was sampled with $1.0$ truncation to yield the maximum diversity. In contrast to BigGAN's samples, our proposed method generate more diverse samples such as varied lighting, poses, scales and context, which are visually closer to the training set.
    The class ids of the samples from top to bottom  are {\footnotesize{ \texttt{009}}, \footnotesize{\texttt{980}} and \footnotesize{\texttt{993}}} respectively. Please refer to appendix for more comparisons.}
    \vspace{-6mm}
    \label{fig:diversity}
\end{figure*}
}

\newcommand{\tmpwidth}{0.23\linewidth}
\setlength{\tabcolsep}{1pt}
\begin{figure}[!t]
    \centering
    \begin{tabular}{c c c c }

	\includegraphics[width=\tmpwidth]{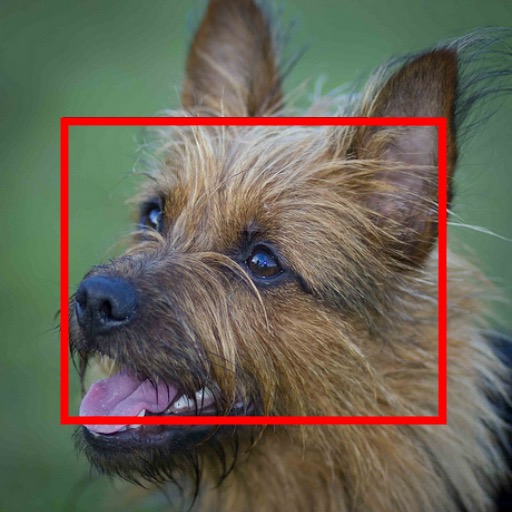} &
	\includegraphics[width=\tmpwidth]{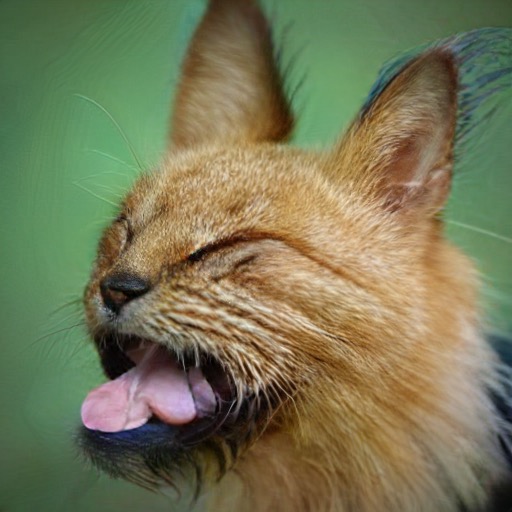}&
	\includegraphics[width=\tmpwidth]{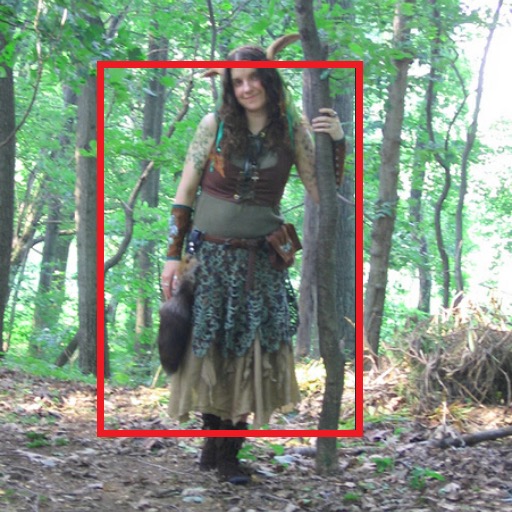} &
	\includegraphics[width=\tmpwidth]{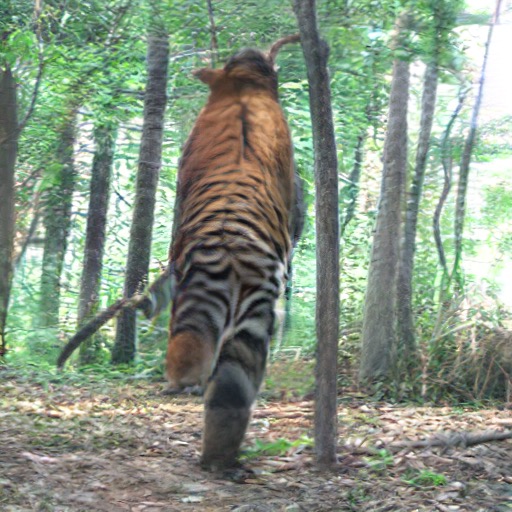} 
	\\
	\includegraphics[width=\tmpwidth]{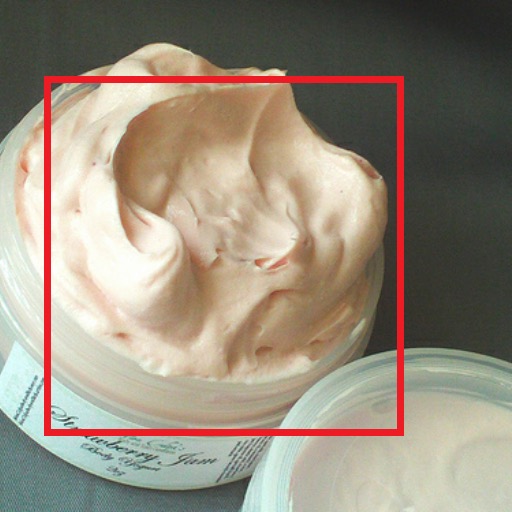} &
	\includegraphics[width=\tmpwidth]{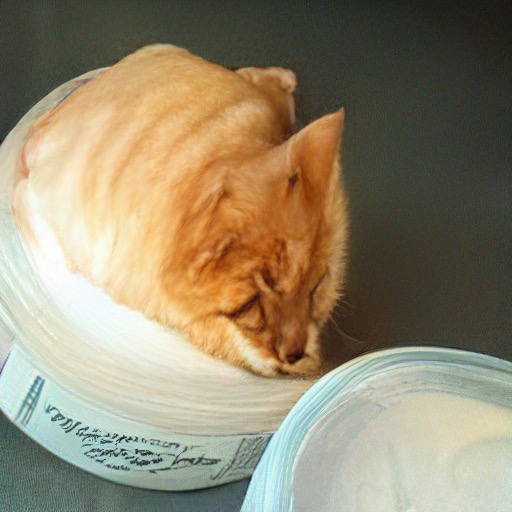} &
	\includegraphics[width=\tmpwidth]{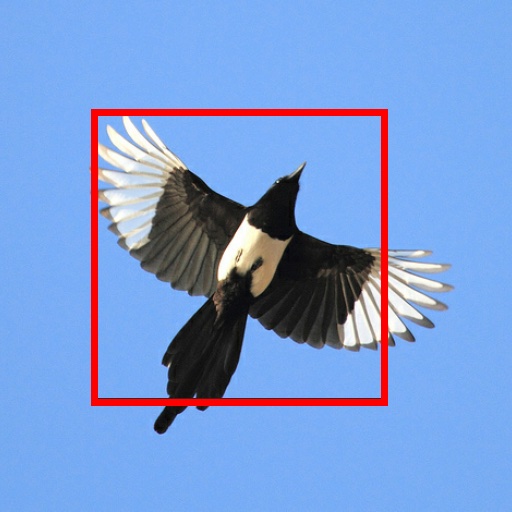} &
	\includegraphics[width=\tmpwidth]{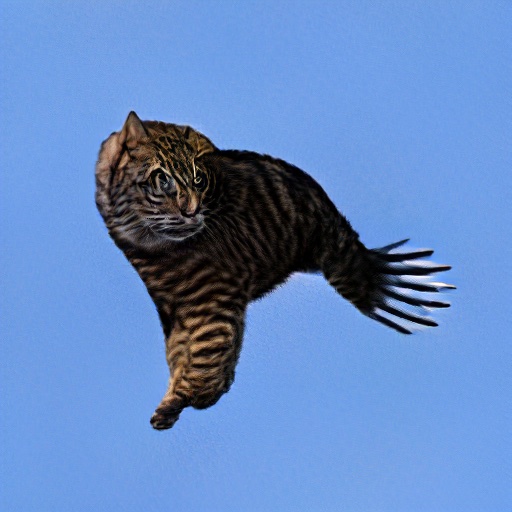} 
	
	\\
	
	\includegraphics[width=\tmpwidth]{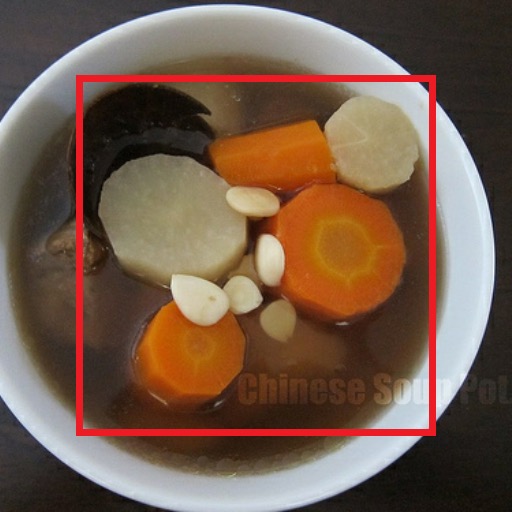} &
	\includegraphics[width=\tmpwidth]{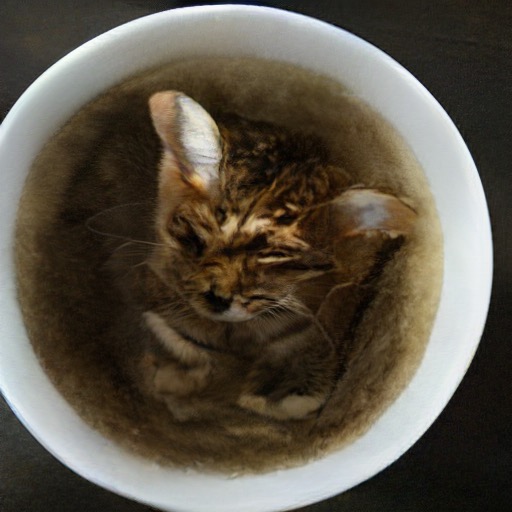} &
	\includegraphics[width=\tmpwidth]{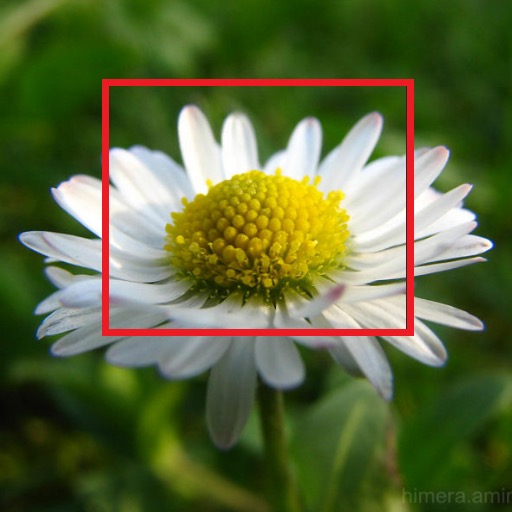} &
    \includegraphics[width=\tmpwidth]{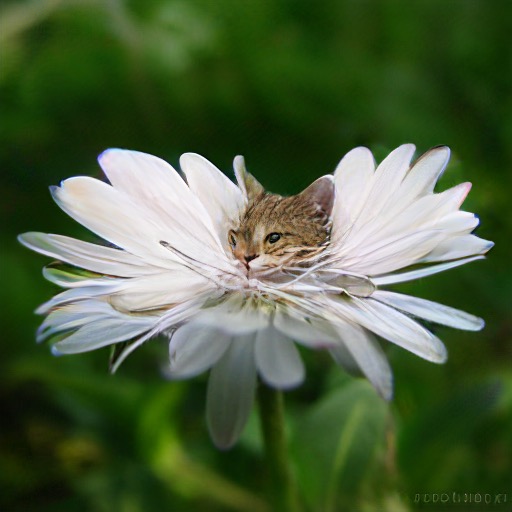} 
    \\

    \end{tabular}
    \vspace{-3mm}
    \caption{\textbf{Class-conditional image editing.} Given input images on the left of each pair, and a target class "tiger cat", \model replaces the bounding boxed regions with tiger cats, suggesting the composition ability of our model. % The fact that the synthesized tiger cat blends in with the rest of the image seamlessly highlights the composition ability of our model. 
    }
    \vspace{-2mm}
    \label{fig:editing}
\end{figure}

\begin{figure}[!t]
    \centering
    \begin{tabular}{cccc}
  
    \includegraphics[width=\tmpwidth]{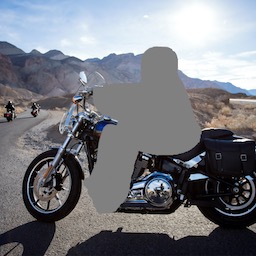}&

    \includegraphics[width=\tmpwidth]{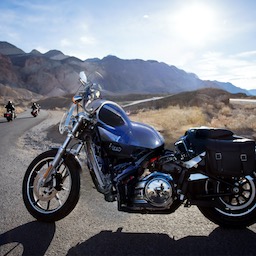}&
    \includegraphics[width=\tmpwidth]{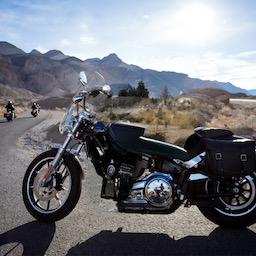}&
    \includegraphics[width=\tmpwidth]{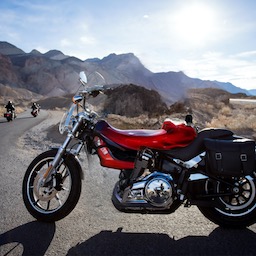}

    \\
    
    \includegraphics[width=\tmpwidth]{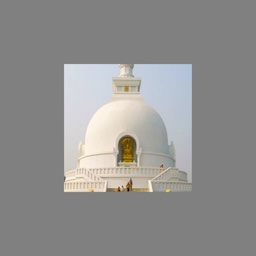}&
    \includegraphics[width=\tmpwidth]{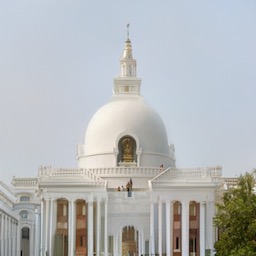} &
    \includegraphics[width=\tmpwidth]{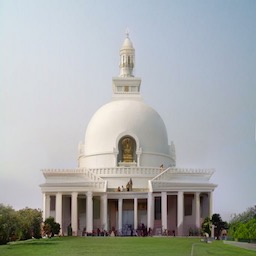} &
    \includegraphics[width=\tmpwidth]{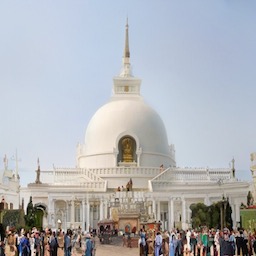} 
    
    \\
    \includegraphics[width=\tmpwidth]{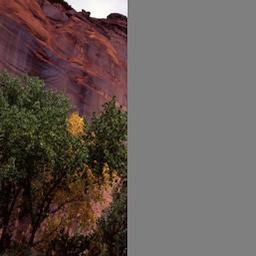}&
    \includegraphics[width=\tmpwidth]{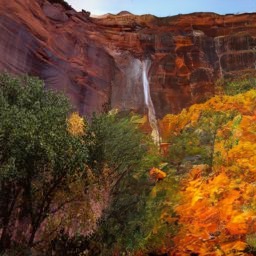} &
    \includegraphics[width=\tmpwidth]{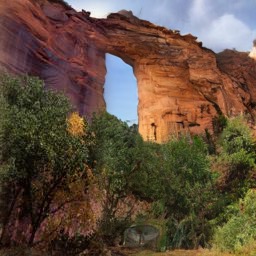} &
    \includegraphics[width=\tmpwidth]{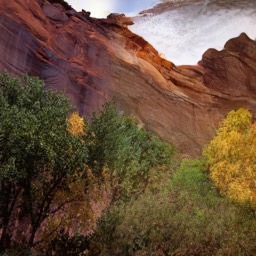} 
   \\
    \vspace{-1mm}
    Input  & \multicolumn{3}{c}{------ \model (Our Samples) ------} \\    
    % Input & DeepFill & HiFill  & \multicolumn{3}{c}{------ \model (Our Samples) ------} & Groundtruth \\    
    \end{tabular}
    \vspace{-1mm}
    \caption{\textbf{Inpainting and outpainting.} Given a single input image, \model synthesizes diverse results for inpainting (first row) and outpainting in different directions (last two rows).}
    \label{fig:uncrop_diversity}
    \vspace{-3mm}
\end{figure}

\setlength{\tabcolsep}{4pt}
\begin{table}[h]
\small
    \centering
    \resizebox{.92\linewidth}{!}{
    \begin{tabular}{llc ccc}
    \toprule
    {Task} & {Model} &
    & {FID}  $\downarrow$ & {IS} $\uparrow$ & %{L1} $\downarrow$ 
   % {Fool Rate} $\uparrow$ 
    \\
    \midrule
    \textit{\textbf{Outpainting}}
    & 
    Boundless \cite{teterwak2019boundless}$^{\square}$ & 
    & 35.02 & 6.15 & 
    %.064%
    \\
    % NS-Outpaint \cite{}$^{\square}$ &
    % & 50.68 & 4.70 &   \\   
     {\footnotesize Right 50\%} & In\&Out \cite{cheng2021inout}$^{\square}$ &
    & 23.57 & 7.18 &  \\
    & InfinityGAN \cite{lin2021infinitygan} &
    & 10.60 & 5.57 &
    \\
    & Boundless \cite{teterwak2019boundless} TF $^{\blackdiamond}$ & 
    & 7.80 & 5.99 &
    %.064% 
    \\
    %  & CoModGAN \cite{zhao2021comodgan}$^{512}$ &
    % & 7.67 & 9.09 &
    % %.064% 
    % \\
    & \textbf{\model (Ours)} $^{512}$& 
    & \textbf{6.78} & \textbf{11.69} &  \\ % [1pt]
    \midrule
    \textit{\textbf{Inpainting}} & 
     DeepFill \cite{yu2019free} &
    & 11.51 & 22.55 & \\
    {\footnotesize Center 50\%$\times$50\%} 
    & ICT\cite{wan2021ict}$^\dagger$ & 
    & 13.63 & 17.70 &  \\    
    & HiFill \cite{yi2020contextual}$^{512}$ & 
    & 16.60 & 19.93 &  \\
    & CoModGAN\cite{zhao2021comodgan}$^{512}$ &
    & \textbf{7.13} & 21.82 & \\    
    & \textbf{\model (Ours)$^{512}$} &
    & 7.92 & \textbf{22.95} & \\
    \bottomrule
    \end{tabular}
    }
\vspace{-2mm}
    \caption{\textbf{Quantitative Comparisons for Inpainting and Outpainting on Places2.} \footnotesize{$^{512}$ evaluated on 512$\times$512 samples while others evaluated on the corresponding 256$\times$256 ones, consistent with their training; $^\square$~taken from the prior work; $^\dagger$~evaluated using the released model trained on a subset of Places2; $^\blackdiamond$~evaluated using the TFHub model\cite{boundless_tfhub}.} }
    \label{tab:inpaint_uncrop}
    \vspace{-3mm}
\end{table}

\noindent\textbf{Speed.} 
We evaluate model speed by assessing the number of steps, \ie forward passes, each model requires to generate a sample. As shown in Table~\ref{tab:maintable}, \model requires the fewest steps among all non-GAN-based models on both resolutions.

To further substantiate the speed difference between \model and autoregressive models, we perform a runtime comparison between \model and VQGAN's decoding processes. As illustrated in Figure~\ref{fig:speed}, \model significantly accelerates VQGAN by $30$-$64$x, with a speedup that gets more pronounced as the image resolution (and thus the input token length) grows.

\noindent\textbf{Diversity.} We consider Classification Accuracy Score (CAS) \cite{Ravuri19CAS} and Precision/Recall~\cite{KynkaanniemiKLL19} as two metrics for evaluating sample diversity, in addition to sample quality.

CAS involves first training a ResNet-50 classifier\cite{ResNet} solely on the samples generated by the candidate model, and then measuring the classifier's classification accuracy on the ImageNet validation set.
The last two columns in Table~\ref{tab:maintable} present the CAS results, where the scores of the classifier trained on real ImageNet training data are included for reference (76.6\% and 93.1\% for the top-1 and top-5 accuracy). For image resolution 256$\times$256,  we follow the common practice of using data augmentation RandAugment\cite{cubuk2019randaugment}, and report the scores trained without augmentation in the appendix ~\ref{sec:supp_diversity_comparison}. We find that \model significantly outperforms prior work VQVAE-2 and VQGAN, establishing a new state-of-the-art of CAS on the ImageNet benchmark on both resolutions.

The Precision/Recall results in Table~\ref{tab:maintable} show that \model achieves better coverage (Recall) compared to BigGAN, and better sample quality (Precision) compared to likelihood-based models such as VQVAE-2 and diffusion models. Compared to our baseline VQGAN, we improve the diversity as measured by recall while slightly boosting its precision. 

In contrast to BigGAN’s samples, \model's samples are more diverse with more varied lighting, poses, scales and context as shown in Figure~\ref{fig:diversity}. More comparisons are available in the appendix ~\ref{sec:supp_diversity_comparison}.

\subsection{Image Editing Applications}
\label{ssec:applications}
In this subsection, we present direct applications of \model on three image editing tasks: class-conditional image editing, image inpainting, and outpainting. All three tasks can be almost trivially translated to ones that \model can handle if we consider the task as just a constraint on the initial binary mask $\unknownmask$ \model uses in its iterative decoding, as discussed in ~\ref{ssec:decoding}. We show that without modifications to the architecture or any task-specific training, \model is capable of generating very compelling results on all three applications. Furthermore, \model obtains comparable performance to dedicated models on both inpainting and outpainiting, even though it is not designed specifically for either task.

\noindent\textbf{Class-conditional Image Editing.}
We define a new class-conditional image editing task to showcase \model's flexibility. In this task, the model regenerates content specified inside a bounding box on the given class while preserving the context, \ie content outside of the box. It is infeasible for autoregressive methods due to the violation to their prediction orders.

For \model, however, it is a trivial task if we consider the bounding box region as the input of initial mask to the iterative decoding algorithm. Figure~\ref{fig:editing} shows a few example results. More can be found in the appendix ~\ref{sec:supp_image_editing}.

In these examples, we observe that \model can reasonably replace the selected object while preserving, or to some extend even completing, the context in the background. Furthermore, we find that \model seems to be capable of synthesizing unnatural yet plausible combinations unseen in the ImageNet training set, \eg a flying cat, cat in a soup bowl, and cat in a flower. This suggests that \model has incidentally learned useful representations for composition, which may be further exploited in related tasks in future works.

\vspace{2mm}
\noindent\textbf{Image Inpainting.}
\label{ssec:inpainting}
Image inpainting or image completion is a fundamental image editing task to synthesize contents in missing regions so that the completion looks visually realistic. Traditional patch-based methods\cite{Barnes:2009:patchmatch} work well on texture regions, while deep learning based methods\cite{yu2019free, yi2020contextual, zhao2021comodgan, saharia2021palette, esser2021imagebart} have been demonstrated to synthesize images requiring better semantic coherence. Both approaches have been are extensively studied in computer vision.

We extend \model to this problem by tokenizing the masked image and interpreting the inpainting mask as the initial mask in our iterative decoding. We then composite the output image by linearly blending it with the input based on the masking boundary following \cite{cheng2021inout}.
To match the training of our baselines, we train \model on the 512$\times$512 center-cropped images from the Places2\cite{zhou2017places} dataset. All hyperparameters are kept the same as the \model model trained on ImageNet.

We compare \model against common GAN-based baselines, including DeepFillv2\cite{yu2019free} and HiFill\cite{yi2020contextual}, on inpainting with a central 50\% $\times$ 50\% mask, which are evaluated on the Places2 validation set. Table~\ref{tab:inpaint_uncrop} summarizes the quantitative comparisons. \model beats both DeepFill and HiFill in FID and IS by a significant margin, while achieving scores close to the state-of-the-art inpainting approach CoModGAN~\cite{zhao2021comodgan}. We show more qualitative comparisons with CoModGAN in the appendix ~\ref{sec:supp_inpainting_and_outpainting_comparison_with_gans}.

\vspace{2mm}
\noindent\textbf{Image Outpainting.}
\label{ssec:outpainting}
Outpainting, or image extrapolation, is an image editing task that has received increased attention recently. It is seen as a more challenging task than inpainting due to the fewer constraints from surrounding pixels and thus more uncertainty in the predicted regions. Our adaptation of the problem and the model used in the following evaluation is the same as in inpainting.

We compare against common GAN-based baselines, including Boundless~\cite{teterwak2019boundless}, In\&Out~\cite{cheng2021inout}, InfinityGAN\cite{lin2021infinitygan}, and CoModGAN\cite{zhao2021comodgan} on extrapolating rightward with a 50\% ratio. We evaluate on the image set generously provided by the authors of InfinityGAN\cite{lin2021infinitygan} and In\&Out\cite{cheng2021inout}.

Table~\ref{tab:inpaint_uncrop} summarizes the quantitative comparisons. \model beats all baselines and achieves state-of-the-art FID and IS. As the examples in Figure~\ref{fig:uncrop_diversity} illustrate, \model is also capable of synthesizing diverse results given the same input with different seeds. We observe that \model completes objects and global structures particularly well, and hypothesize that this is thanks to the model learning useful representations with the global attentions in the transformer.

\subsection{Ablation Studies}
\setlength{\tabcolsep}{5pt}
\begin{table}[!t]
\small
    \centering
    {\small
    \resizebox{0.7\linewidth}{!}{
        \begin{tabular}{lc cccc}
            \toprule
            {\large$\gamma$} &  %\bfseries{Sampling method}  &
              & $T$ & \bfseries{FID}  $\downarrow$ & \bfseries{IS} $\uparrow$ & \bfseries{NLL} \\
            \midrule
            Exponential & 
            % & 7  & 7.63 & 160.9 & 4.83 \\
            & 8  & 7.89 & 156.3 & 4.83 \\
            Cubic & 
            & 9 & 7.26 & 165.2 & 4.63  \\ 
            Square & 
            & 10 & 6.35 & 179.9 & 4.38  \\ 
            \textbf{Cosine} & 
            & 10 & \textbf{6.06} & \textbf{181.5} & 4.22 \\    
            Linear & 
            & 16 & 7.51 & 113.2 & 3.75  \\
            Square Root &
            & 32 & 12.33 & 99.0 & 3.34 \\  
            Logarithmic  &
            & 60 & 29.17 & 47.9 &  3.08\\
            % Exponential & 
            % & 16  & 6.62 & 166.5 & 4.83 \\
            % Cubic & 
            % & 32 & 6.12 & 170.9 & 4.63  \\ 
            % Square & 
            % & 64 & 5.55 & 179.8 & 4.38  \\ 
            % \textbf{Cosine} & 
            % & 64 & \textbf{5.14} & \textbf{184.0} & 4.22 \\    
            % Linear & 
            % & 256 & 6.09 & 132.6 & 3.75  \\
            % Square Root &
            % & 256 & 9.26 & 120.7 & 3.34 \\  
            % Logarithmic  &
            % & 256 & 21.18 & 63.2 &  3.08\\
    \bottomrule
    \end{tabular}
    }
    }
    \vspace{-.1cm}
    \caption{\textbf{Ablation results on the mask scheduling functions.} We report the best FID, IS, and Negative Log-Likelihood loss for each candidate scheduling function. }
    \label{tab:ablation}
\vspace{-3mm}
\end{table}

\label{ssec:ablation}
We conduct ablation experiments using the default setting on ImageNet 256$\times$256.

\begin{figure}[!t]
    \centering
\includegraphics[width=1.\linewidth]{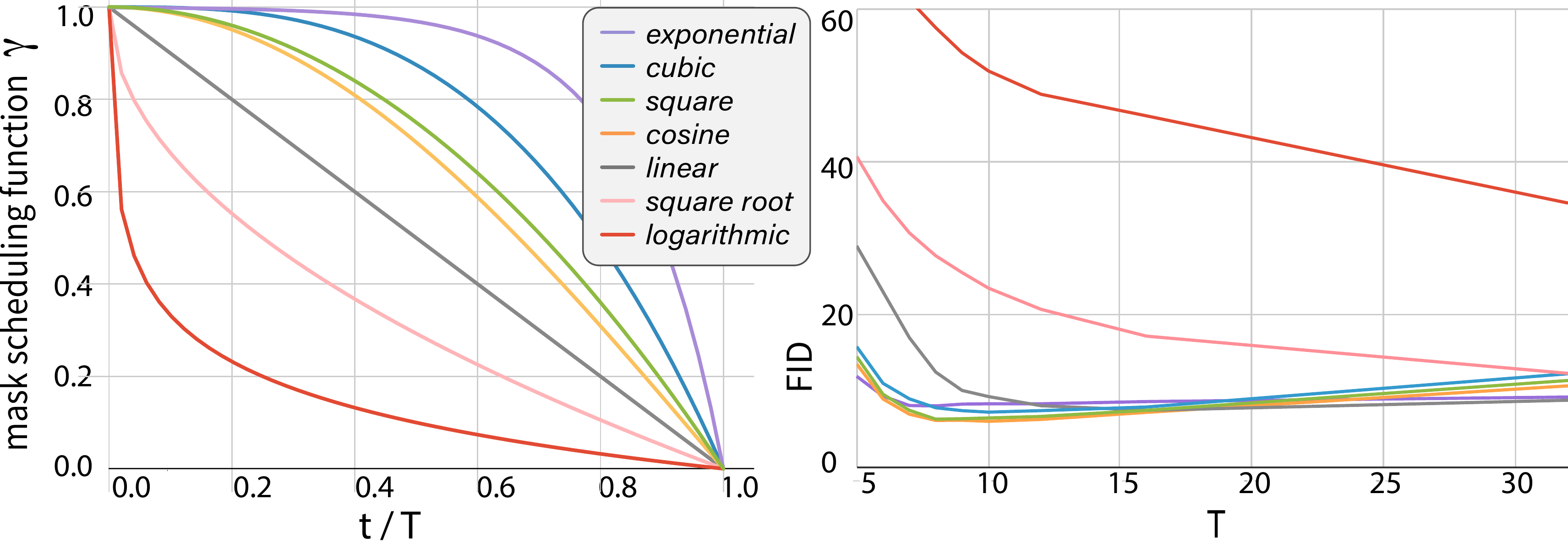}
    \vspace{-6mm}
    \caption{\textbf{Choices of Mask Scheduling Functions} $\gamma (\frac{t}{T})$, and 
    \textbf{number of iterations \textit{T}}. On the left, we visualize seven functions we consider for $\gamma$. On the right, we show line graphs of models' FID scores against the number of decoding iterations $T$. Among the candidates, we find that cosine achieves the best FID.}
    \label{fig:scheduling}
    \vspace{-3mm}
\end{figure}

\noindent{\textbf{Mask scheduling.}}
A key design of \model is the mask scheduling function used in both training and iterative decoding. We compare the functions discussed in ~\ref{ssec:masking}, visualize them in Figure~\ref{fig:scheduling}, and summarize the results in Table~\ref{tab:ablation}.

We observe that concave functions generally obtain better FID and IS than linear, followed by the convex functions. While cosine and square perform similarly relative to other functions, cosine slightly edges out square in all scores, making cosine the default in our model.

We hypothesize that concave functions perform favorably because they 1) challenge training with more difficult cases (\ie encouraging larger mask ratios), and 2) appropriately prioritize the less-to-more prediction throughout the decoding.
That said, over-prioritization seems to be costly as well, as shown by the cubic function being worse than square, and exponential being much worse than all other concave functions.

\noindent{\textbf{Iteration number.}}
We study the effect of the number of iterations ($T$) on our model by running all candidate masking functions with different $T$s. As shown in Figure~\ref{fig:scheduling}, under the same setting, more iterations are not necessarily better: as $T$ increases, aside from the logarithmic function which performs poorly throughout, all other functions hit a ``sweet spot" where the model's performance peaks before it worsens again. The sweet spot also gets ``delayed" as functions get less concave. As shown, among functions that achieve strong FIDs (\ie cosine, square, and linear), cosine not only has the strongest overall score, but also the earliest sweet spot at a total of $8$ to $12$ iterations.
We hypothesize that such sweet spots exist because too many iterations may discourage the model from keeping less confident predictions, which worsens the token diversity. We think further study on the masking design would be interesting for future work.

\section{Conclusion}

In this paper, we propose \model, a novel image synthesis paradigm using a bidirectional transformer decoder. Trained on Masked Visual Token Modeling, \model learns to generate samples using an iterative decoding process within a constant number of iterations. Experimental results show that \model significantly outperforms the state-of-the-art transformer model on conditional image generation, and our model is readily extendable to various image manipulation tasks. As \model achieves competitive performance with state-of-the-art GANs, applying our approach to other synthesis tasks is a promising direction for future work. Please see the appendix \ref{sec:supp_limitations} for the limitations and future work.

\noindent \textbf{Acknowledgement} 
The authors would like to thank Xiang Kong for inspiring related works and anonymous reviewers for helpful comments.

%%%%%%%%% REFERENCES

\bibliographystyle{ieee}
\bibliography{main}

% \vfill
% \pagebreak

%%%%%%%%%%%%%%%%%%%%%%%%%%%%%%%%%%%%%%%%%%%%%%%%%%%%%%%%%%%%

\appendix
\newcommand{\suppsection}[1]{\section{{\fontsize{12}{13}\selectfont #1}}}

\renewcommand{\tmpwidth}{18mm}
\setlength{\tabcolsep}{1pt}

\twocolumn[
\begin{@twocolumnfalse}
{
\begin{center}
\begin{tabular}{m{20mm}m{2mm} cccc c cccc}

Original & 
&
&&& \parbox[c]{\tmpwidth}{\includegraphics[width=\tmpwidth]{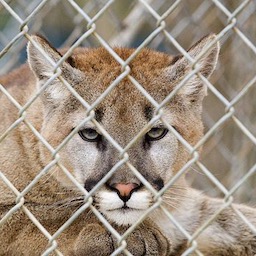}} &
&
&&& \parbox[c]{\tmpwidth}{\includegraphics[width=\tmpwidth]{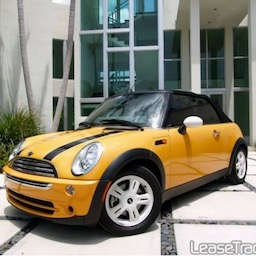}} 
\\
\\[-4pt]
&
& Mask 95\% & Mask 90\% & Mask 85\% & Mask 75\% &
& Mask 95\% & Mask 90\% & Mask 85\% & Mask 75\% 
\\ 
\parbox[l]{15mm}{\small{Example Input Mask}} &
&
\parbox[c]{\tmpwidth}{\includegraphics[width=\tmpwidth]{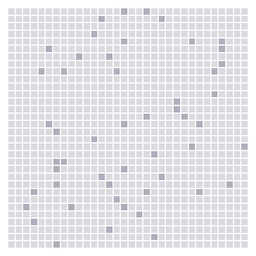}}&
\parbox[c]{\tmpwidth}{\includegraphics[width=\tmpwidth]{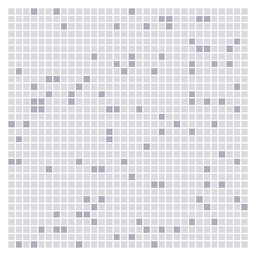}}&
\parbox[c]{\tmpwidth}{\includegraphics[width=\tmpwidth]{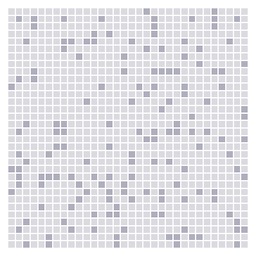}}&
\parbox[c]{\tmpwidth}{\includegraphics[width=\tmpwidth]{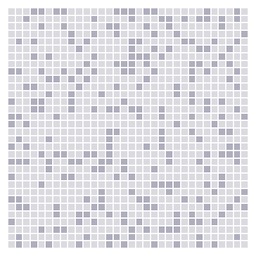}}&
&
\parbox[c]{\tmpwidth}{\includegraphics[width=\tmpwidth]{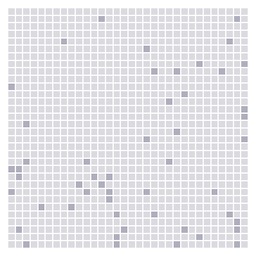}}&
\parbox[c]{\tmpwidth}{\includegraphics[width=\tmpwidth]{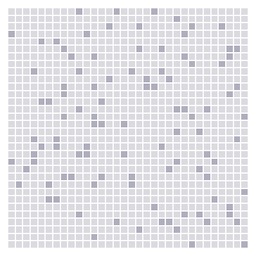}}&
\parbox[c]{\tmpwidth}{\includegraphics[width=\tmpwidth]{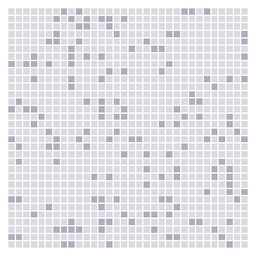}}&
\parbox[c]{\tmpwidth}{\includegraphics[width=\tmpwidth]{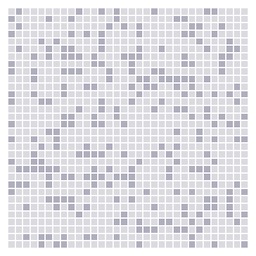}}

\\
\parbox[l]{15mm}{\small{Reconstruction Sample}} &
&
\parbox[c]{\tmpwidth}{\includegraphics[width=\tmpwidth]{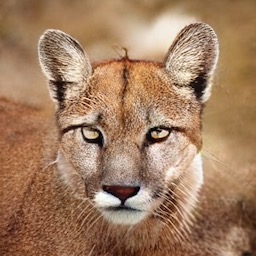}}&
\parbox[c]{\tmpwidth}{\includegraphics[width=\tmpwidth]{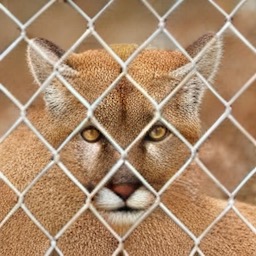}}&
\parbox[c]{\tmpwidth}{\includegraphics[width=\tmpwidth]{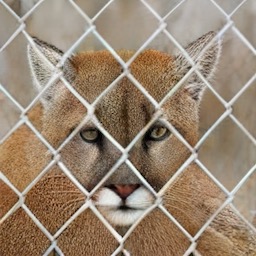}}&
\parbox[c]{\tmpwidth}{\includegraphics[width=\tmpwidth]{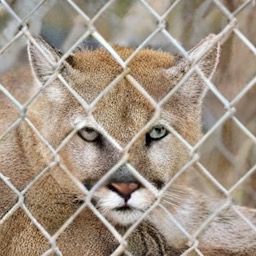}}&
&
\parbox[c]{\tmpwidth}{\includegraphics[width=\tmpwidth]{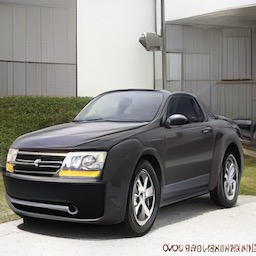}}&
\parbox[c]{\tmpwidth}{\includegraphics[width=\tmpwidth]{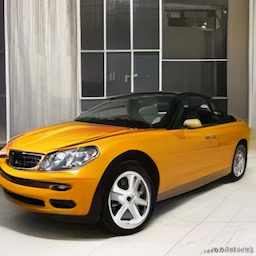}}&
\parbox[c]{\tmpwidth}{\includegraphics[width=\tmpwidth]{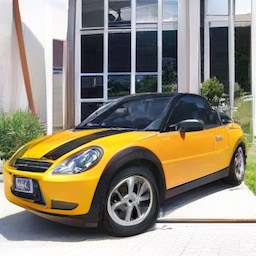}}&
\parbox[c]{\tmpwidth}{\includegraphics[width=\tmpwidth]{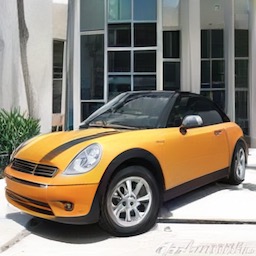}}

\\
\parbox[l]{20mm}{\small{Median of $100$ Samples}}
 & 
&
\parbox[c]{\tmpwidth}{\includegraphics[width=\tmpwidth]{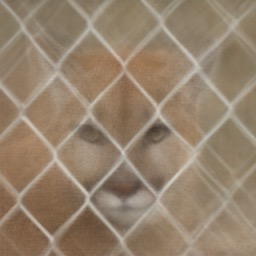}}&
\parbox[c]{\tmpwidth}{\includegraphics[width=\tmpwidth]{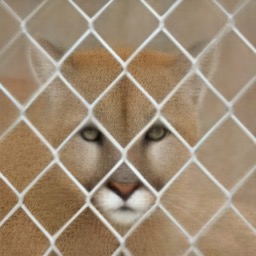}}&
\parbox[c]{\tmpwidth}{\includegraphics[width=\tmpwidth]{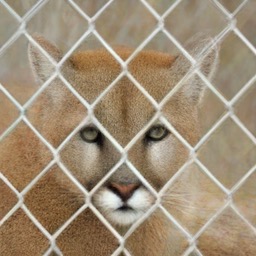}}&
\parbox[c]{\tmpwidth}{\includegraphics[width=\tmpwidth]{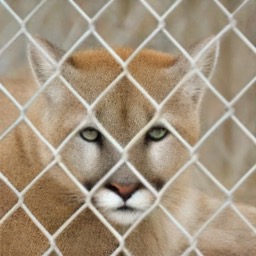}}&
&
\parbox[c]{\tmpwidth}{\includegraphics[width=\tmpwidth]{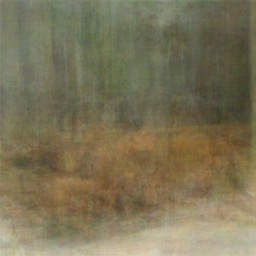}}&
\parbox[c]{\tmpwidth}{\includegraphics[width=\tmpwidth]{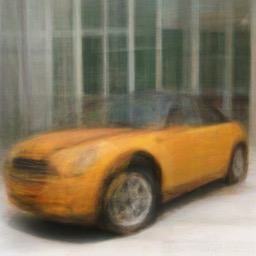}}&
\parbox[c]{\tmpwidth}{\includegraphics[width=\tmpwidth]{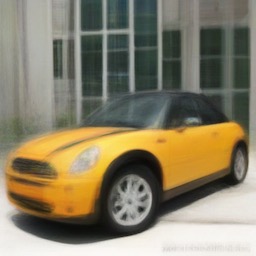}}&
\parbox[c]{\tmpwidth}{\includegraphics[width=\tmpwidth]{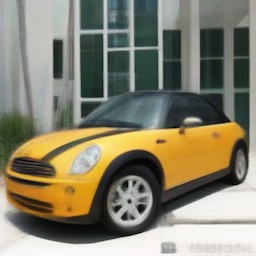}}
\\
\end{tabular}
\captionof{figure}{\textbf{Examples of \model on Image Reconstruction.} \model takes in masked tokens extracted from original images (row one) using random input masks (row two, with unknown tokens marked in light gray), and outputs reconstructed images (row three). We then randomly sample 100 masks with the same mask ratio, and illustrate the median of the 100 reconstructed samples in row four.}
% The sharpness of the median may be interpreted as the consensus among the reconstruction samples.}
\label{fig:supp_reconstruction}
\end{center}
\vspace{10mm}
}
\end{@twocolumnfalse}
]

\suppsection{Discussion on Image Reconstruction}
\label{sec:supp_reconstruction_discussion}

\begin{figure}[!h]
    \centering
    \begin{tabular}{cc}
     \includegraphics[width=41mm]{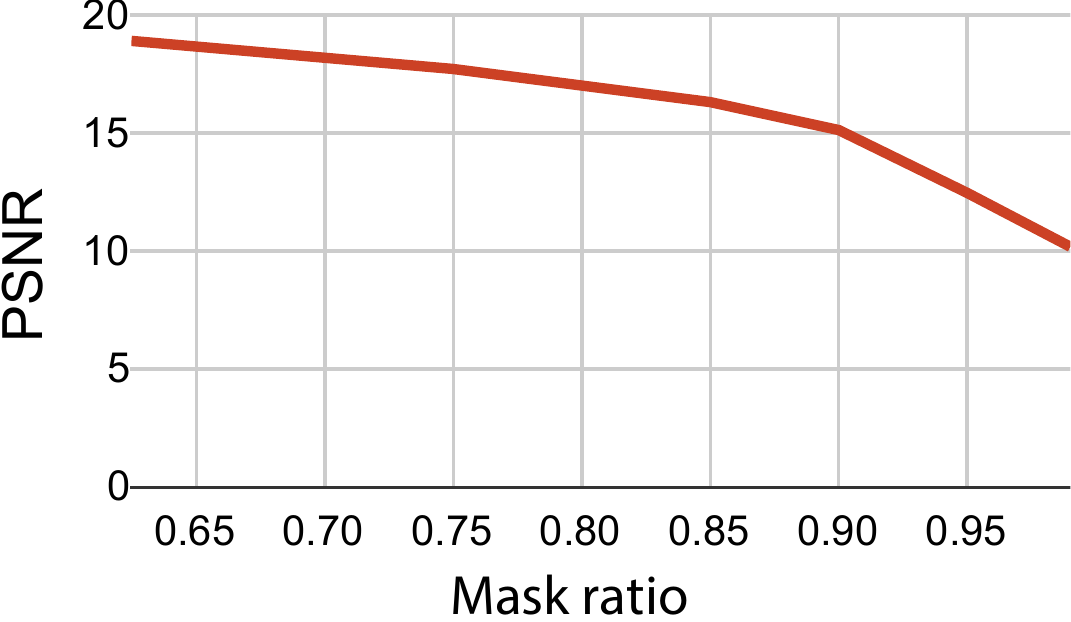} & 
      \includegraphics[width=41mm]{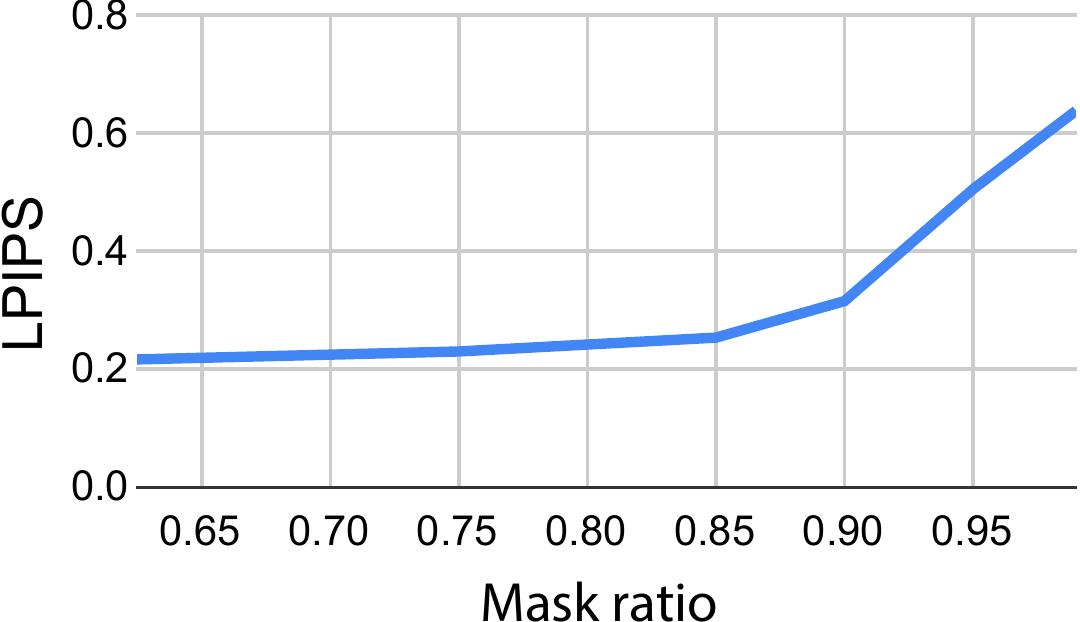} 
    \end{tabular}
      \vspace{-3mm}
    \caption{Reconstruction quality and diversity measured by PSNR and LPIPS\cite{zhang2018lpips}.}
    \label{fig:analysis_psnr}
    \vspace{-3mm}
\end{figure}

In \ref{ssec:class_conditional_synthesis}, we primarily evaluate \model on class-conditional image generation tasks. Here we offer more discussion on its performance on image reconstruction. We set up by first randomly sampling input mask $M$ with a mask ratio $r$ of the visual tokens masked out, and then running \model's iterative decoding algorithm to reconstruct images. Figure~\ref{fig:analysis_psnr} shows the PSNR and LPIPS\cite{zhang2018lpips} of the reconstructed samples as functions of $r$, whereas Figure~\ref{fig:supp_reconstruction} visualizes two examples of this process with $r$ ranging from $95\%$ to $75\%$.

We observe that \model reconstructs holistic information (\eg pose and shape of the foreground objects) even with a very high percentage (\eg 95\%) of tokens masked out. More importantly, there seems to exist an inflection point around 90\%: while both reconstruction quality and consistency improve drastically as the mask ratio decreases until 90\%, after 90\% further improvements are slowed down. This observation is corroborated by the large jump in the visual similarity between reconstruction samples and the original images from 95\% to 90\% in Figure~\ref{fig:supp_reconstruction}, \eg the fence in front of the tiger and the car's color are consistently captured once the mask ratio is below 90\%, but not at 95\%.

In other words, we find that visual tokens are highly redundant. For a holistic reconstruction, only a very small portion (\eg 10\%) of the tokens are essential; the remaining ones merely improve the recovery of finer appearance or details. This echos our intuition behind the masking design laid out in \ref{ssec:masking} that the prediction of the first few tokens is key to image generation. Similar observations on the spatial redundancy of images are discussed in a concurrent paper MAE~\cite{he2021mae}. In their work, they find that masking a high proportion of the input image yields a nontrivial and meaningful self-supervisory task for image representation learning.

\setlength{\tabcolsep}{2pt}

\suppsection{Additional Class-conditional Image Generation Results}
\label{sec:supp_diversity_comparison}
 
In this section, we report additional results on class-conditional image generation.

We follow prior transformer-based methods\cite{Razavi19vqvae2,Esser21vqgan} to employ the classifier-based rejection sampling to improve the sample quality scores. Specifically, we use a pre-trained ResNet classifier\cite{ResNet} to score output samples based on the predicted probability and keep samples with an acceptance rate of $0.05$, as in VQGAN~\cite{Esser21vqgan}. 
As shown in Table \ref{tab:supp_cc_classifier}, MaskGIT demonstrates consistent improvement over VQGAN, and is comparable with ADM with classifier guidance~\cite{dhariwal2021diffusion}. More importantly, by adding the rejection sampling, \model achieves state-of-the-art Inception Scores ($355.6$ on 256$\times$256 and $342.0$ on 512$\times$512).

In Table~\ref{tab:supp_cas}, we report Precision and Recall scores calculated using Inception features~\cite{christian16inception}. In contrast to the VGG\cite{simonyan2015vgg} feature-based scores, which we report in Table~\ref{tab:maintable} for a more direct comparison with prior work~\cite{KynkaanniemiKLL19,dhariwal2021diffusion}, we find that the Inception feature-based scores are more consistent with our qualitative observations that VQGAN’s samples are more diverse than BigGAN’s. Under both measures, \model’s recall scores outperform those of BigGAN and VQGAN. We also report CAS evaluated on classifiers trained without augmentation from RandAugment\cite{cubuk2019randaugment}. Consistent with our main results, \model outperforms BigGAN and our baseline VQGAN by a large margin.

Finally, we show a few comparisons of the class-conditional samples generated by \model with the samples generated by BigGAN-deep and VQVAE-2 in Figure~\ref{fig:supp_cc_diversity}, \ref{fig:supp_cc_diversity_1}, and \ref{fig:supp_cc_diversity_2}.

\setlength{\tabcolsep}{4pt}
\begin{table}[h]
\small
    \centering
    
    \resizebox{0.97\linewidth}{!}{
        \begin{tabular}{lc l lc cc}
        \toprule
Dataset && Model & Classifier guidance && FID & IS          \\ 
\midrule
% ADM~[9]  & & & 23.24& 58.06 && 10.94 & 101.0  \\
% % VQ-GAN~[11]       & & & n/a & n/a && 15.78 & 78.3  \\
% VQGAN\cite{Esser21vqgan} & & & 26.52$^\ast$ & 66.18$^\ast$ && 15.78 & 78.3 \\
% MaskGIT                         & & & 7.32 & 156.0 && 6.18 & 182.1 \\ \cline{2-8}

ImageNet && ADM~\cite{dhariwal2021diffusion}  & 1.0 guidance  && 4.59 & 186.70 \\
256$\times$256 && VQGAN~\cite{Esser21vqgan} & 0.05 acceptance rate    &&  5.88 & 304.8 \\
&& MaskGIT & 0.05 acceptance rate    &          & \textbf{4.02} & \textbf{355.6} \\ \midrule
ImageNet && ADM~\cite{dhariwal2021diffusion}  & 1.0 guidance  && 7.72 & 172.71 \\
% 512x512 && VQGAN~\cite{Esser21vqgan}$^\ast$ & 0.05 acceptance rate && &  \\
512$\times$512 && MaskGIT & 0.05 acceptance rate    && \textbf{4.46} & \textbf{342.0} \\
  \bottomrule
    \end{tabular}
    
    }
    \vspace{-1mm}
    \caption{\small{Class-conditional image synthesis on ImageNet for methods with classifier guidance.
    %\small{$^\ast$ denotes the model we train with the same architecture and setup with ours.}
    }}
    \label{tab:supp_cc_classifier}
\vspace{-5mm}
\end{table}

\newcolumntype{N}{@{}m{0pt}@{}}
\setlength{\tabcolsep}{2.5pt}
\begin{table}[!th]
\small
    \centering
    \resizebox{.93\linewidth}{!}{
    \begin{tabular}{lc ccc ccc }
    \toprule
     {Model} & 
     &  {Prec}  $\uparrow$ & {Rec}   $\uparrow$ &
     & \multicolumn{2}{c}{{{CAS} $\times 100$} $\uparrow$}  
     \\
     \cline{6-7}
    & &&& &{{Top-1 (73.1)}} & {{Top-5 (91.5)}} \\
   % \textbf{ImageNet 256$\times$256}  &&& &&& &
   %\\[-2pt]
    \midrule
     
    BigGAN-deep~\cite{biggan}  & %, 1.0 trunc. &
    & \textbf{0.82} & 0.27 &
    & 42.65 &65.92 
    \\
    VQ-GAN$^\ast$ &
    & 0.61 & 0.47 &
    & 47.50 & 68.90
    \\
    \bfseries{\model (Ours)} &
    & 0.78 & \bfseries{0.50} &
   & \bfseries{58.20} & \bfseries{79.65} 

    \\
    % \textbf{ImageNet 512$\times$512} &&& &&& &\\[-2pt]
    % \midrule
    
    %  BigGAN-deep~\cite{biggan} & 
    %  & \textbf{0.88} & 0.29 &
    %  &44.02&68.22
    %  \\
    %  VQGAN$^\ast$ &
    %  & 0.73 & 0.31 &
    %   &51.29&74.24 \\
      
    %  \bfseries{\model (Ours)} & 
    %  &  0.78 & 0.50 &
    %  &63.43&84.79 \\
    %  % \bfseries{Ours}, 0.5 RS & ? & ? & &  \\
    \bottomrule
    
    \end{tabular}
    }
    \vspace{-1mm}
    \caption{More quantitative comparison with BigGAN-deep and our baseline VQGAN on ImageNet 256$\times$256.
    \footnotesize{ $^\ast$ denotes the model we train with the same architecture and setup with ours.} 
    }
    \vspace{-5mm}
    \label{tab:supp_cas}
    % On average 10-15ms per step

\end{table}
\renewcommand{\tmpwidth}{50mm}
\setlength{\tabcolsep}{2pt}
\begin{figure*}[!ht]
    \centering
    \begin{tabular}{c c c}

    BigGAN-deep (FID=$6.95$) & VQVAE-2$^\dagger$ (FID=$31$) & \model (FID=\bestfid) \\
    \includegraphics[ width=\tmpwidth]{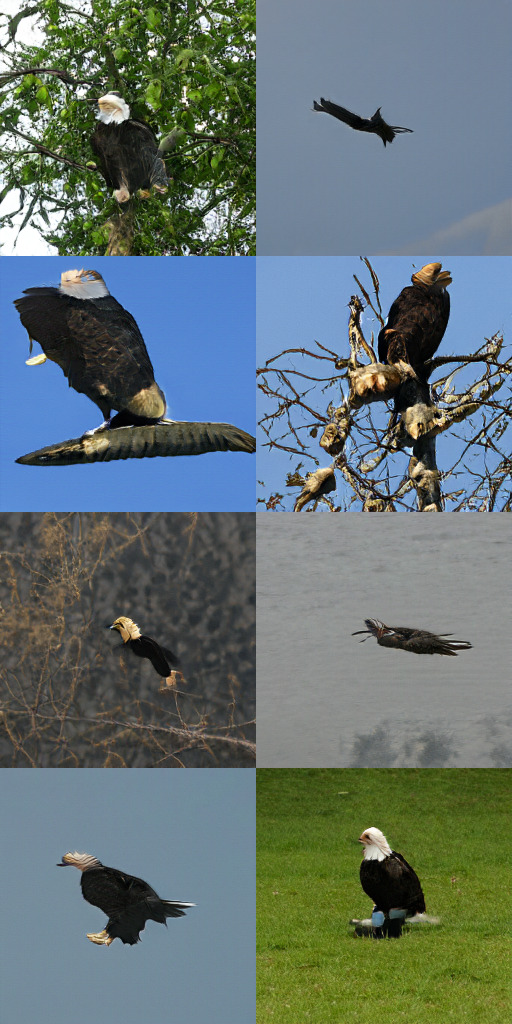} & 
    \includegraphics[ width=\tmpwidth]{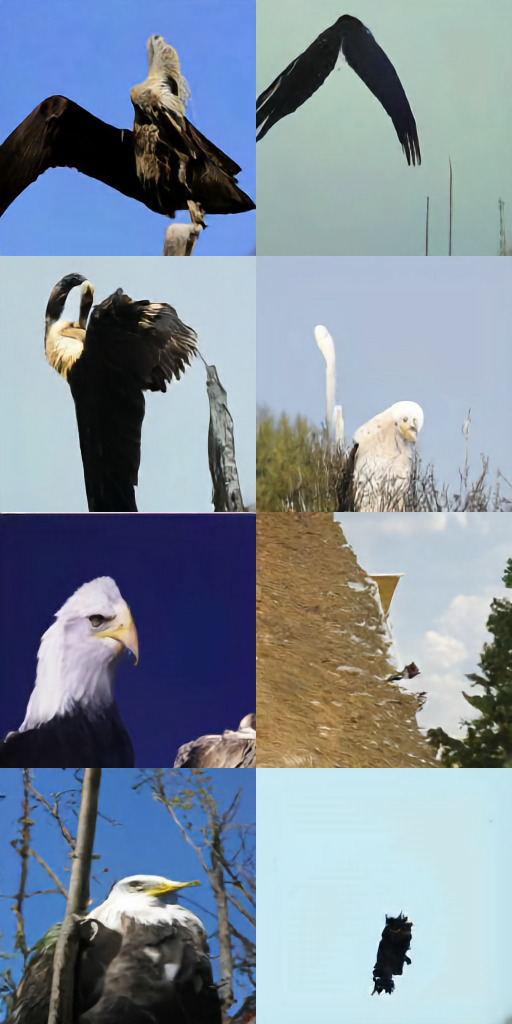}& 
    \includegraphics[ width=\tmpwidth]{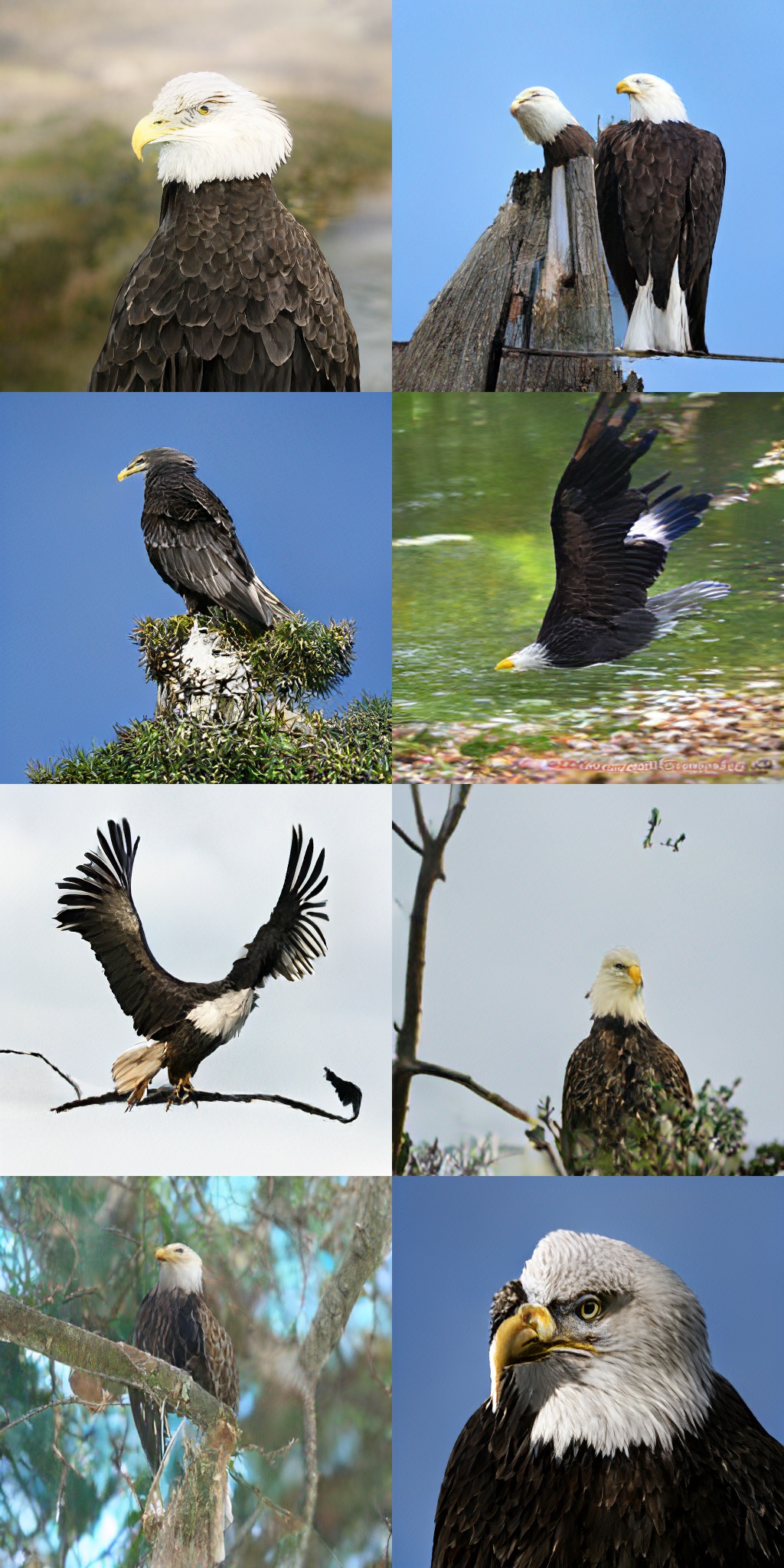} \\ 
       
    \includegraphics[ width=\tmpwidth]{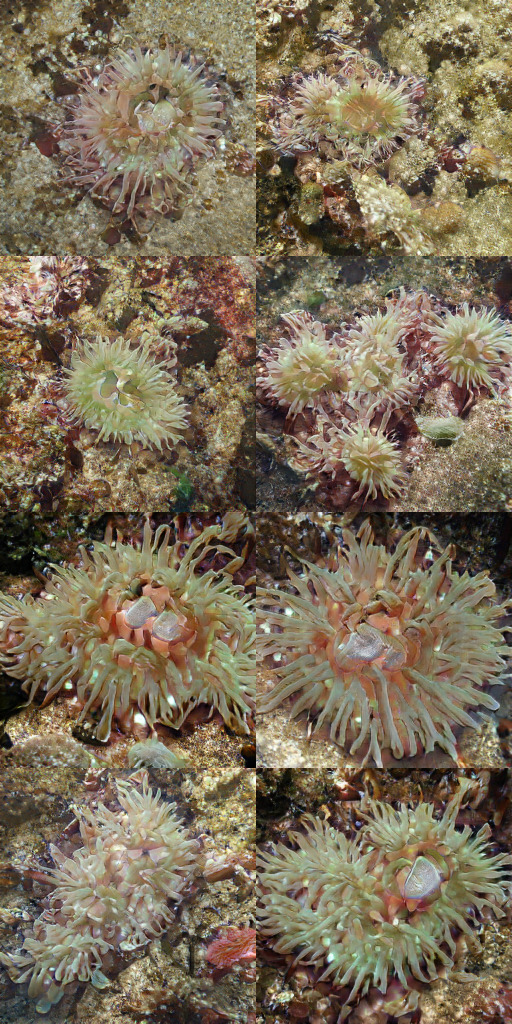} & \
    \includegraphics[ width=\tmpwidth]{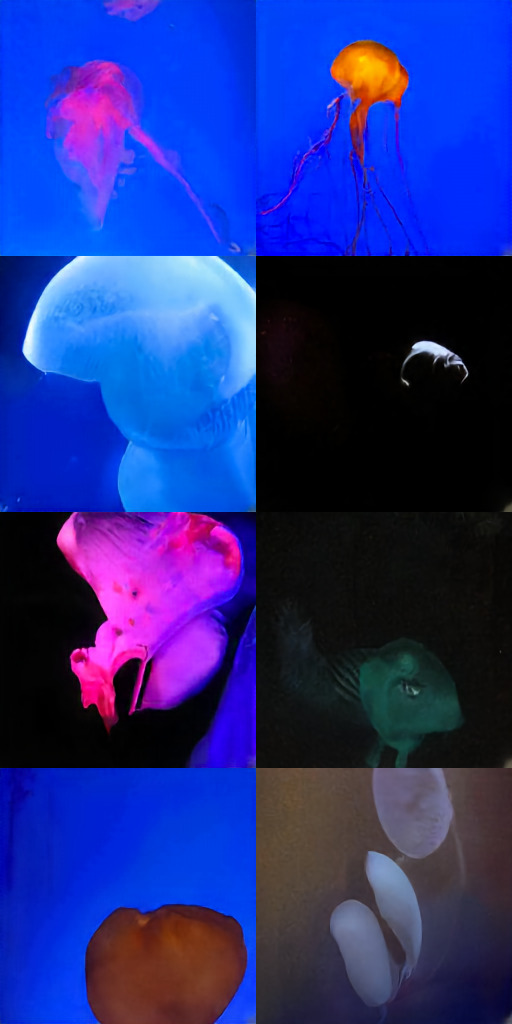}& \
    \includegraphics[ width=\tmpwidth]{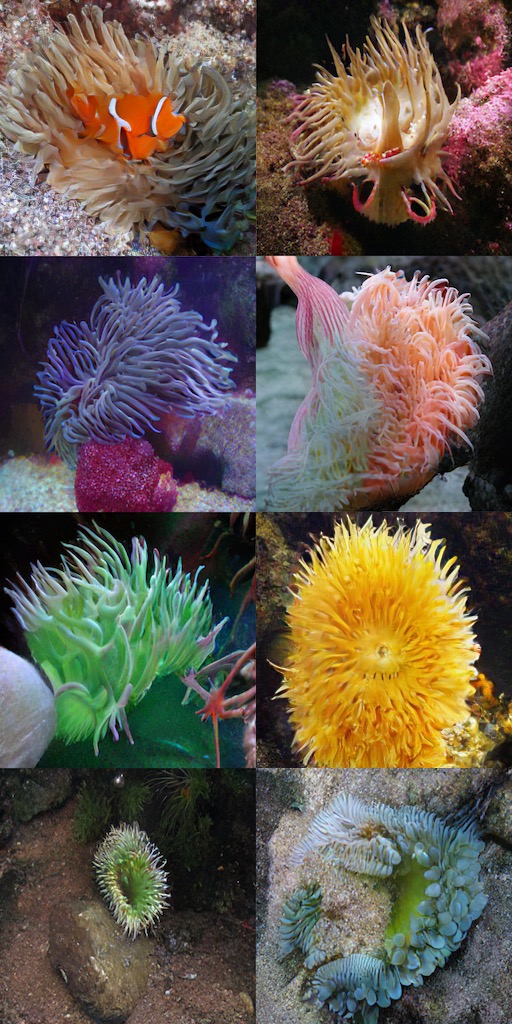} \\

    \end{tabular}
    \caption{\textbf{More diversity comparisons} between BigGAN-deep with truncation $1.0$, VQVAE-2\cite{Razavi19vqvae2}, and our proposed method \model on ImageNet. $^\dagger$ represents extracted samples from the paper.}
    \vspace{-6mm}
    \label{fig:supp_cc_diversity}
\end{figure*}
    
\begin{figure*}[!ht]
    \centering
    \begin{tabular}{c c c}
    
    BigGAN-deep (FID=$6.95$) & VQVAE-2 (FID=$31$) & \model (FID=\bestfid) \\
    \includegraphics[ width=\tmpwidth]{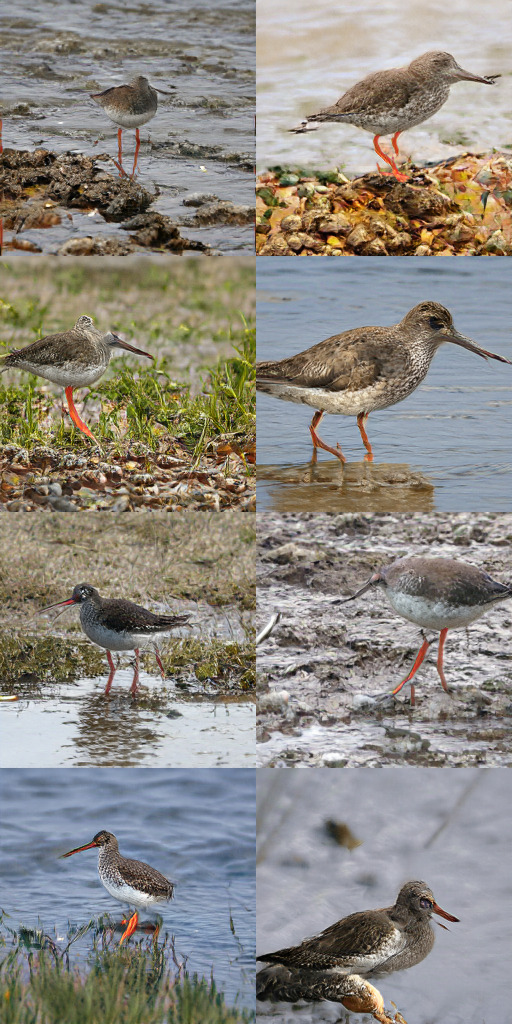} & \
    \includegraphics[ width=\tmpwidth]{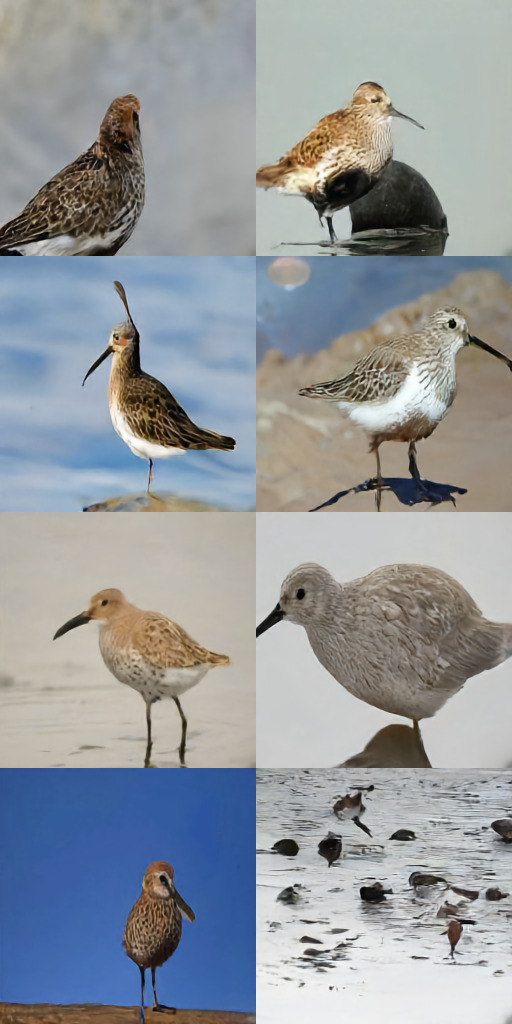}& \
    \includegraphics[ width=\tmpwidth]{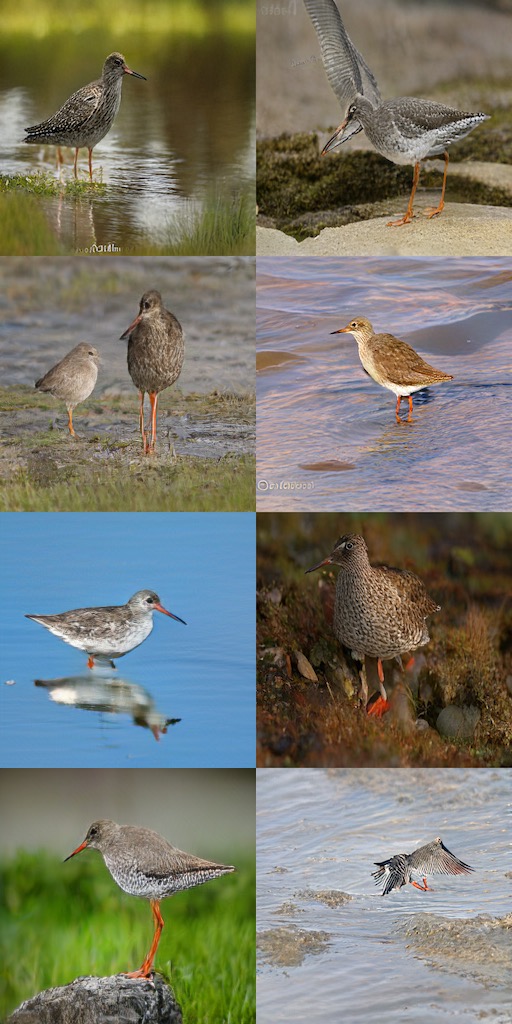} \\ 
    
    \includegraphics[ width=\tmpwidth]{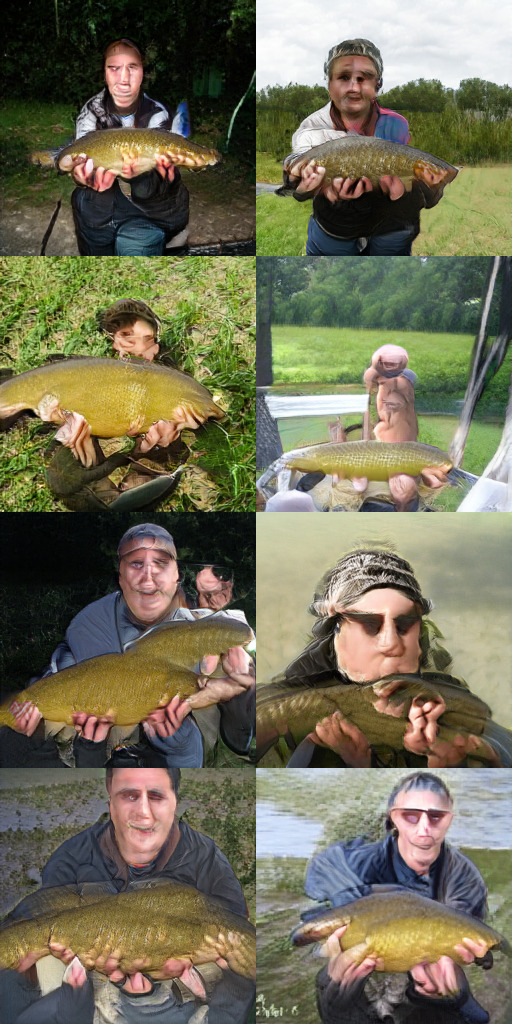} & \
    \includegraphics[ width=\tmpwidth]{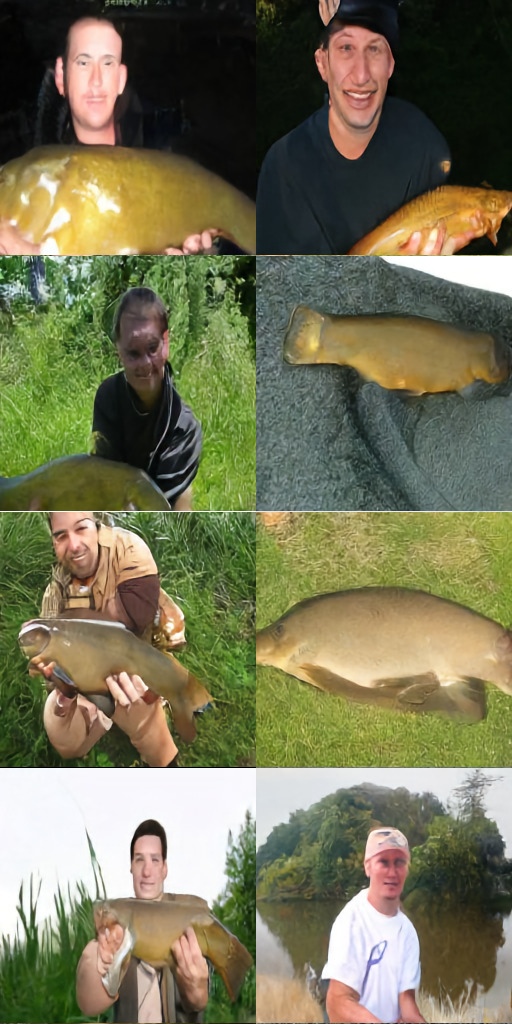}& \
    \includegraphics[ width=\tmpwidth]{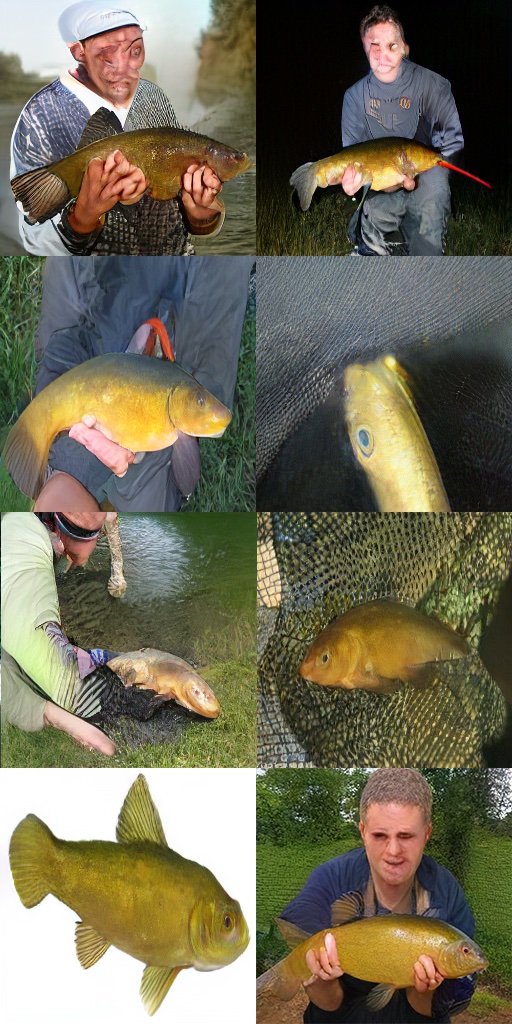} \\ 
     
    \end{tabular}
    \caption{\textbf{More diversity comparisons} between BigGAN-deep with truncation $1.0$, VQVAE-2\cite{Razavi19vqvae2}, and our proposed method \model on ImageNet. $^\dagger$ represents extracted samples from the paper.}
    \vspace{-6mm}
    \label{fig:supp_cc_diversity_1}
\end{figure*}

\begin{figure*}[!ht]
    \centering
    \begin{tabular}{c c c}
    BigGAN-deep (FID=$6.95$) & VQVAE-2 (FID=$31$) & \model (FID=\bestfid)  \\
    
    \includegraphics[ width=\tmpwidth]{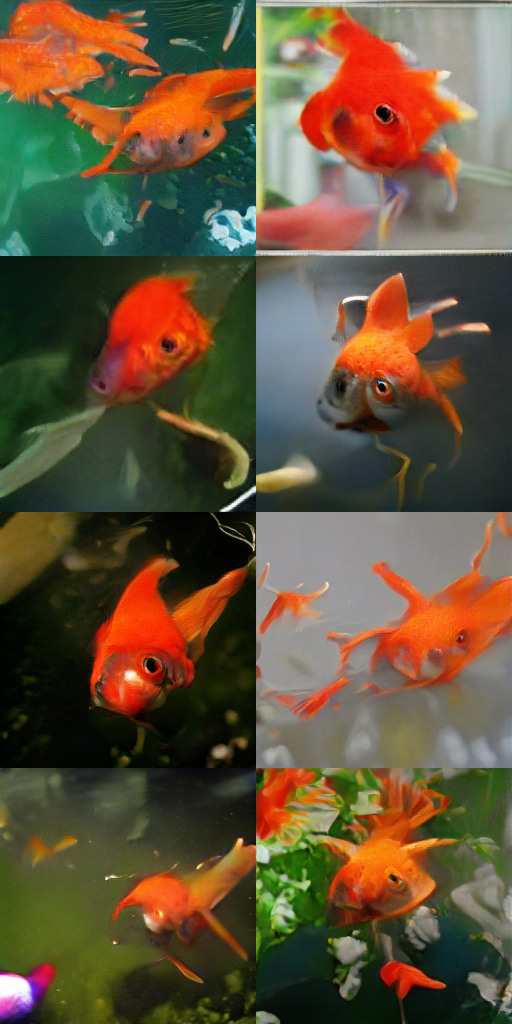} & \
    \includegraphics[ width=\tmpwidth]{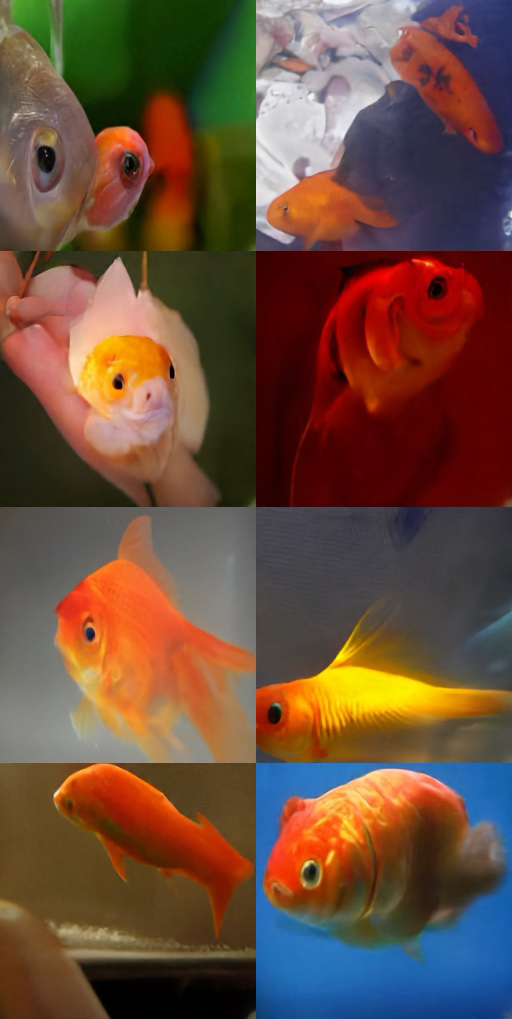}& \
    \includegraphics[ width=\tmpwidth]{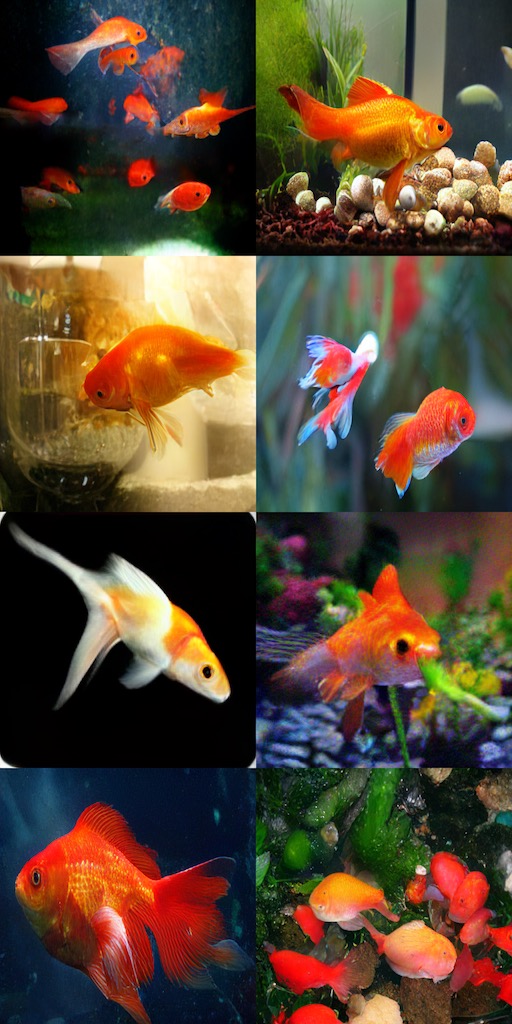} \\ 
    
    \includegraphics[ width=\tmpwidth]{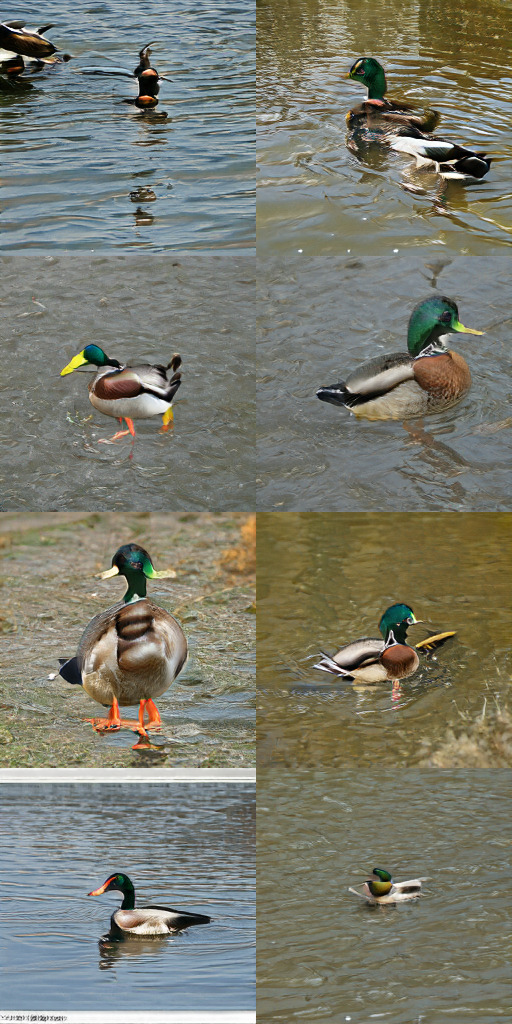} & \
    \includegraphics[ width=\tmpwidth]{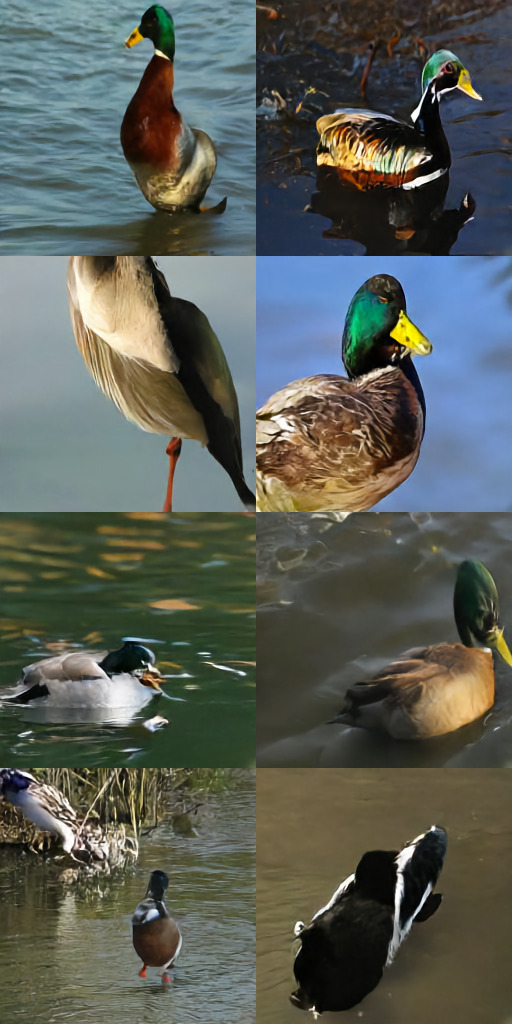}& \
    \includegraphics[ width=\tmpwidth]{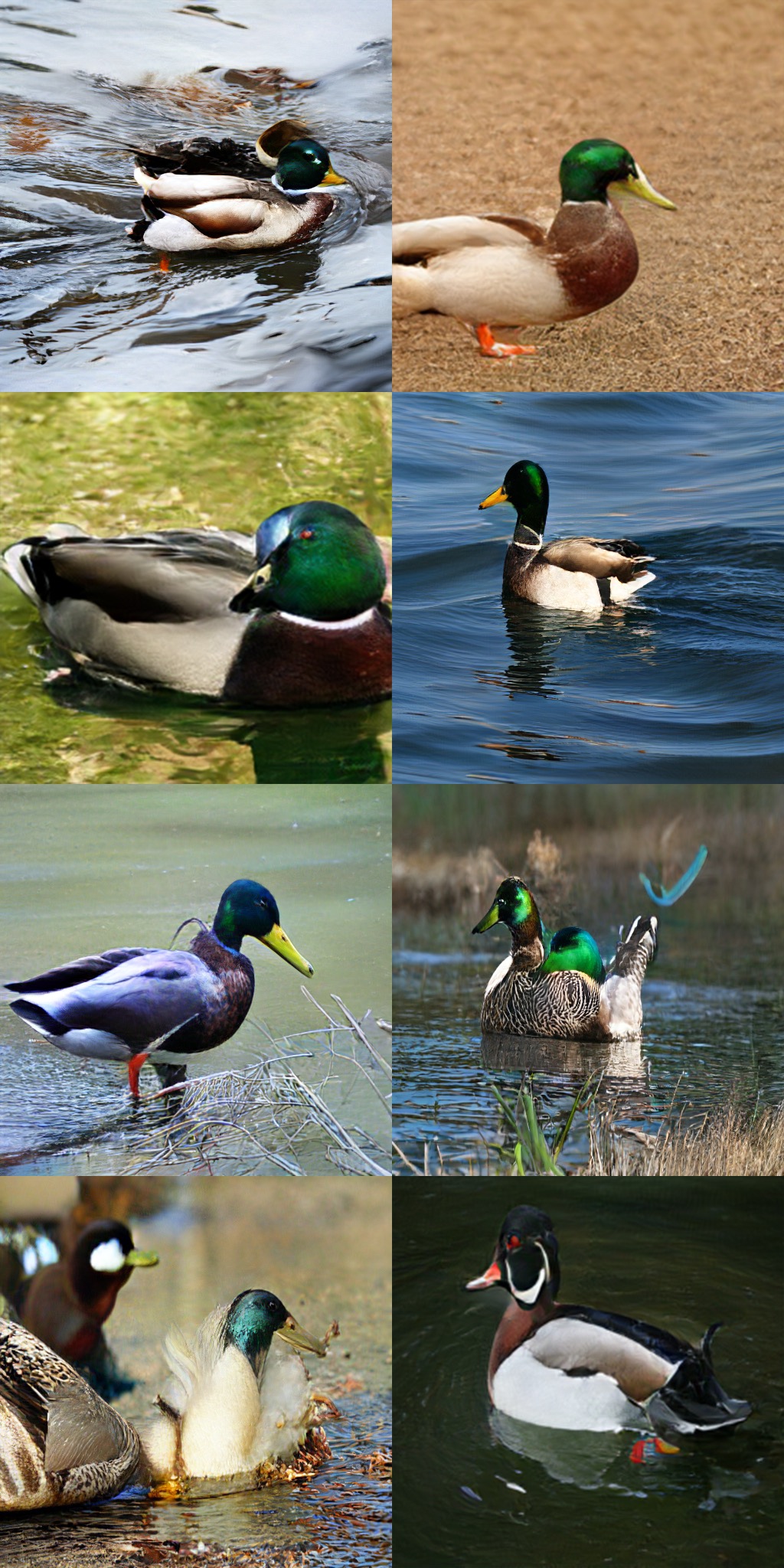} \\ 
    
\end{tabular}
    \caption{More Diversity Comparisons among BigGAN-deep with truncation $1.0$, VQVAE-2\cite{Razavi19vqvae2}, and our proposed method \model on ImageNet. $^\dagger$ represents extracted samples from the paper.}
    \vspace{-6mm}
    \label{fig:supp_cc_diversity_2}
\end{figure*}

\suppsection{Additional Examples of Class-conditional Image Editing Applications}
\label{sec:supp_image_editing}

\renewcommand{\tmpwidth}{28mm}

\setlength{\tabcolsep}{2pt}
\begin{figure*}[!ht]
    \centering
    \begin{tabular}{m{12mm}m{5mm} c c c c c}
Input Image &&  
\parbox[c]{\tmpwidth}{\includegraphics[ height=\tmpwidth]{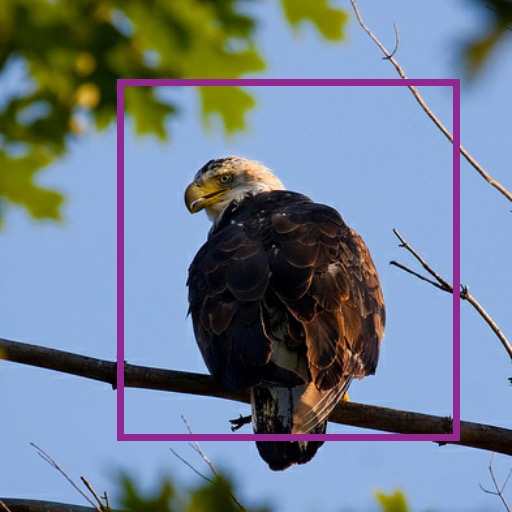}} & 
 \parbox[c]{\tmpwidth}{\includegraphics[ height=\tmpwidth]{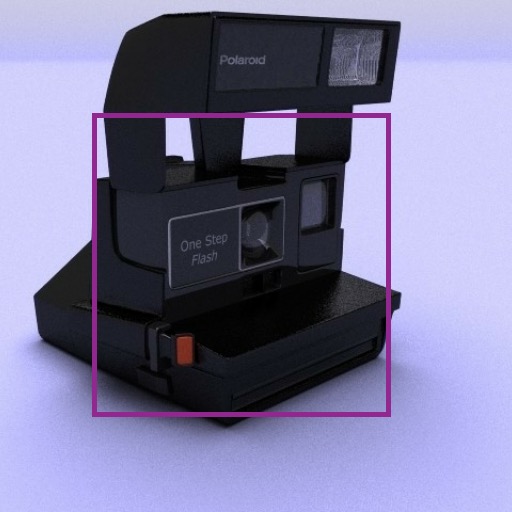}} & 
  \parbox[c]{\tmpwidth}{\includegraphics[ height=\tmpwidth]{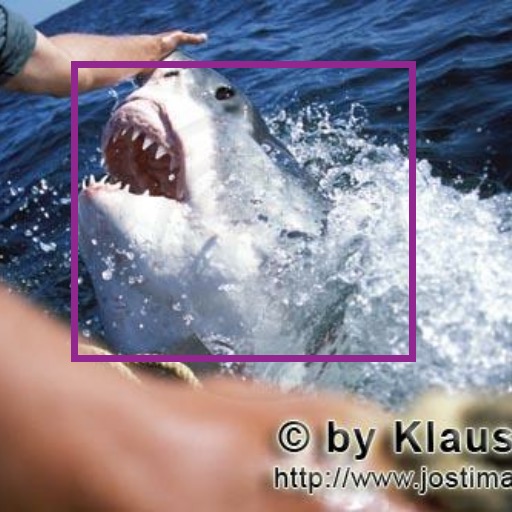}} & 
\parbox[c]{\tmpwidth}{\includegraphics[ height=\tmpwidth]{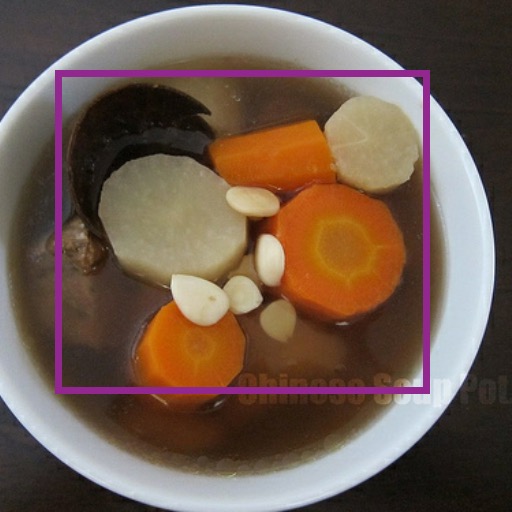}} & 
 \parbox[c]{\tmpwidth}{\includegraphics[ height=\tmpwidth]{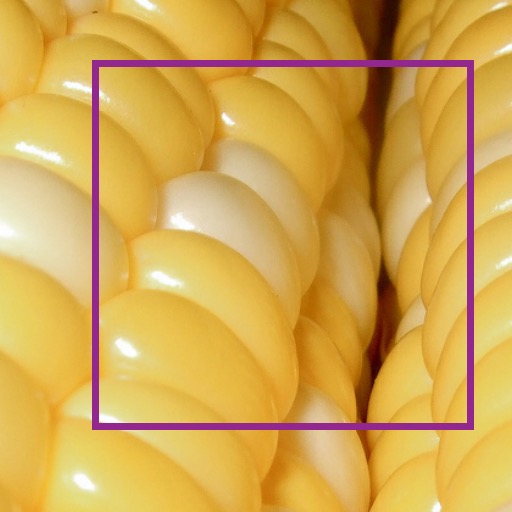}} 
% \\ 
% Input Mask &&  
%  \parbox[c]{\tmpwidth}{\includegraphics[ width=\tmpwidth]{figures/supp_class_modification/tree_mask.jpeg}} & 
% \parbox[c]{\tmpwidth}{\includegraphics[ width=\tmpwidth]{figures/supp_class_modification/008784_mask.jpeg}} &
% \parbox[c]{\tmpwidth}{\includegraphics[ width=\tmpwidth]{figures/supp_class_modification/water_mask.jpeg}} &
%  \parbox[c]{\tmpwidth}{\includegraphics[ height=\tmpwidth]{figures/supp_class_modification/soup_mask.jpeg}} & 
%  \parbox[c]{\tmpwidth}{\includegraphics[ width=\tmpwidth]{figures/supp_class_modification/corn_mask.jpeg}} 

\\  \addlinespace[3pt]

Goldfish \texttt{[001]} && 
\parbox[c]{\tmpwidth}{\includegraphics[ width=\tmpwidth]{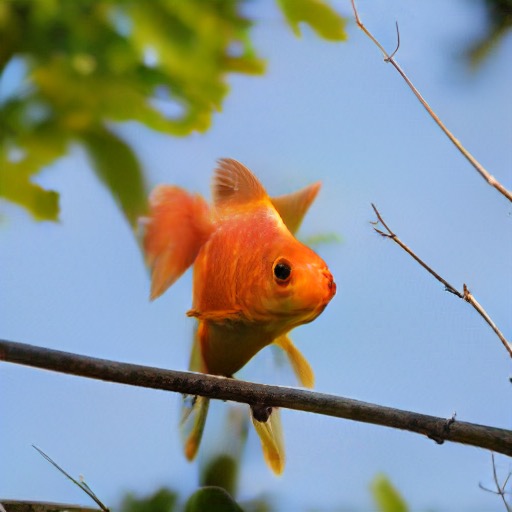}}&
\parbox[c]{\tmpwidth}{\includegraphics[ width=\tmpwidth]{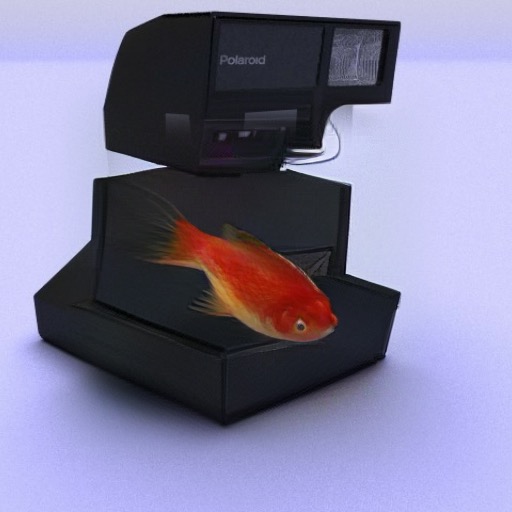}} & 
\parbox[c]{\tmpwidth}{\includegraphics[ width=\tmpwidth]{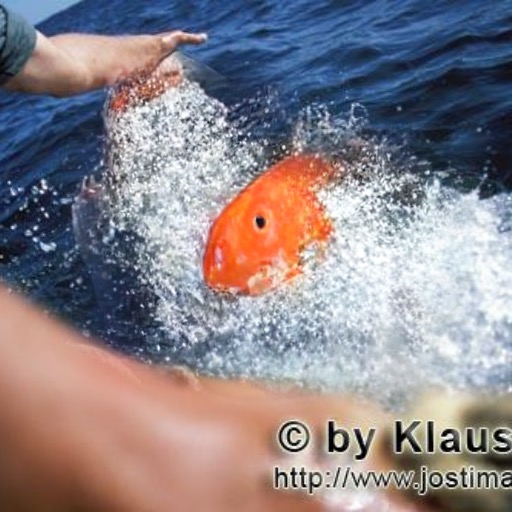}} & 
% \parbox[c]{\tmpwidth}{\includegraphics[ height=\tmpwidth]{figures/supp_class_modification/water1_to_001025_0090_output.jpeg}} & 
\parbox[c]{\tmpwidth}{\includegraphics[ height=\tmpwidth]{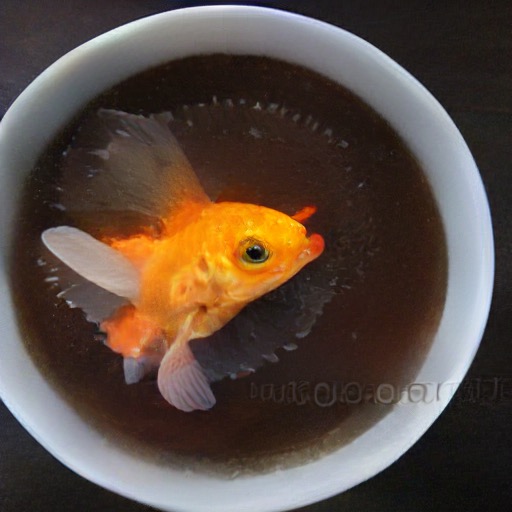}} & 
\parbox[c]{\tmpwidth}{\includegraphics[ width=\tmpwidth]{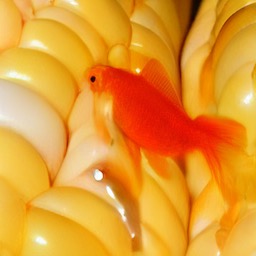}} 

\\

Ice Bear \texttt{[296]} && 
\parbox[c]{\tmpwidth}{\includegraphics[ width=\tmpwidth]{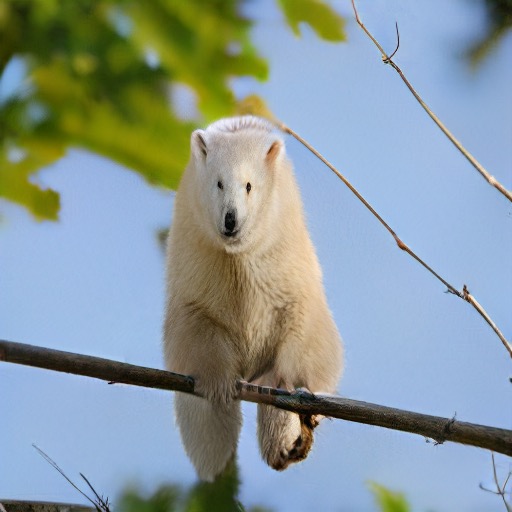}} & 
\parbox[c]{\tmpwidth}{\includegraphics[ width=\tmpwidth]{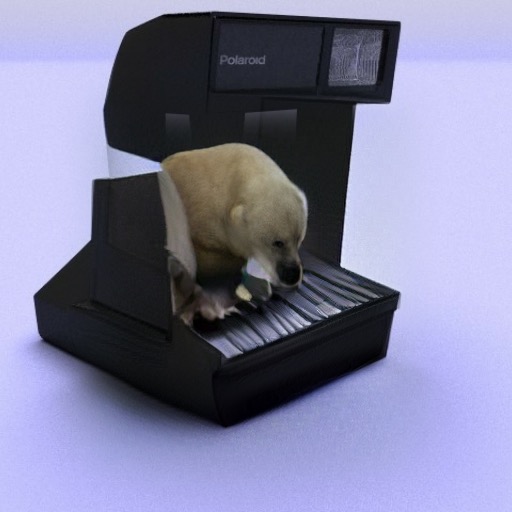}} & 
\parbox[c]{\tmpwidth}{\includegraphics[ width=\tmpwidth]{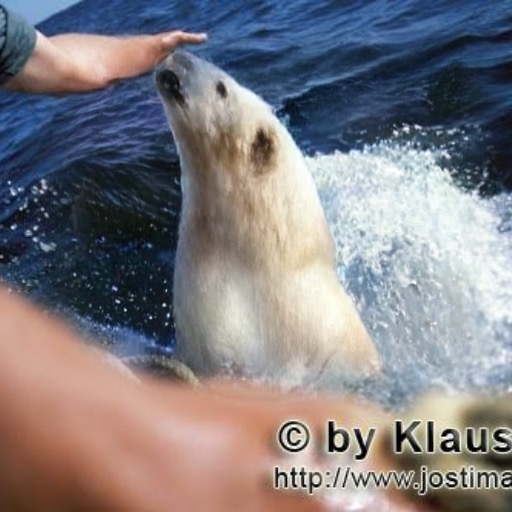}} & 
\parbox[c]{\tmpwidth}{\includegraphics[ height=\tmpwidth]{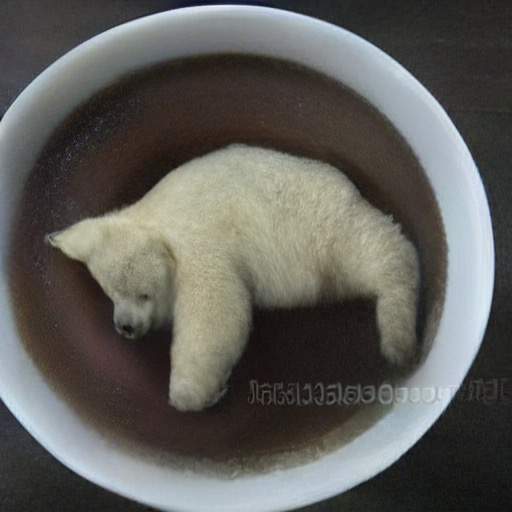}} & 
\parbox[c]{\tmpwidth}{\includegraphics[ width=\tmpwidth]{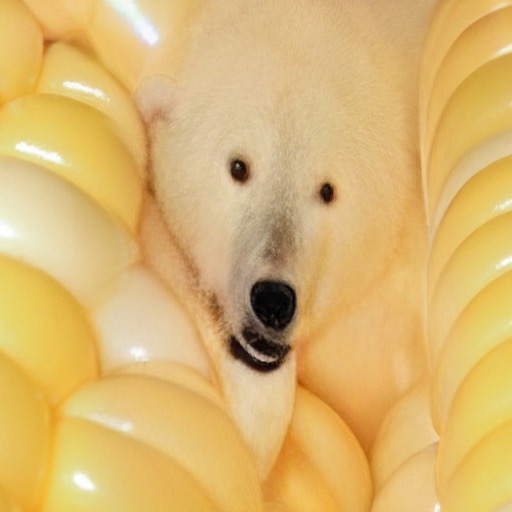}} 
\\
Argaric 
\texttt{[992]} &&
\parbox[c]{\tmpwidth}{\includegraphics[ width=\tmpwidth]{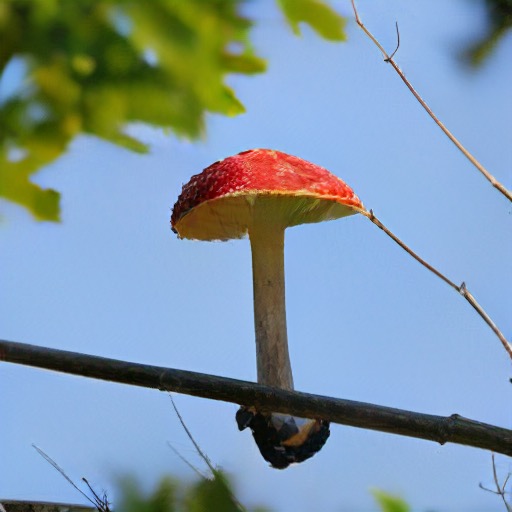}} & 
\parbox[c]{\tmpwidth}{\includegraphics[ width=\tmpwidth]{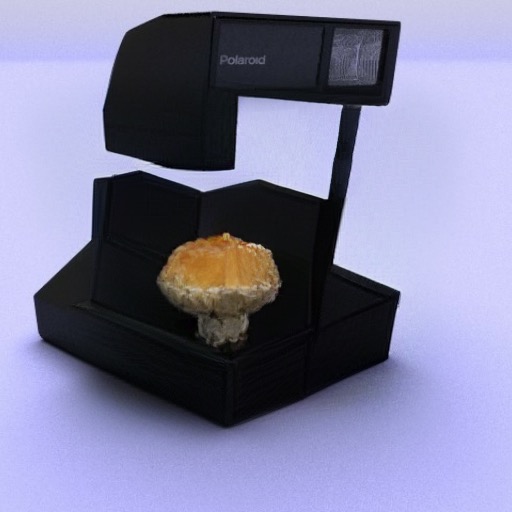}} & 
\parbox[c]{\tmpwidth}{\includegraphics[ width=\tmpwidth]{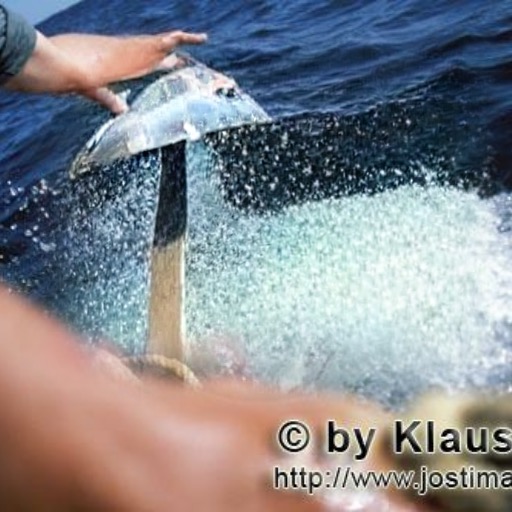}} & 
\parbox[c]{\tmpwidth}{\includegraphics[ height=\tmpwidth]{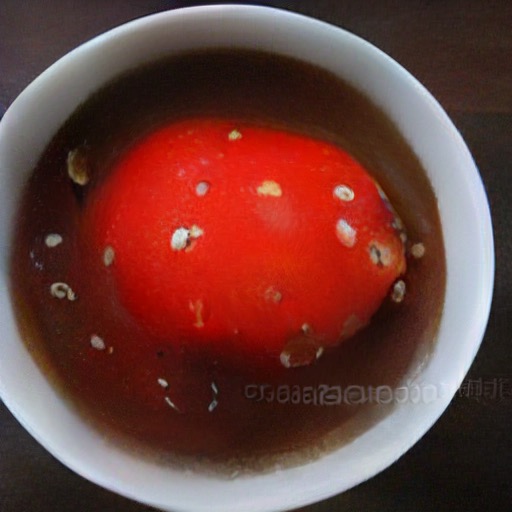}} & 
\parbox[c]{\tmpwidth}{\includegraphics[ width=\tmpwidth]{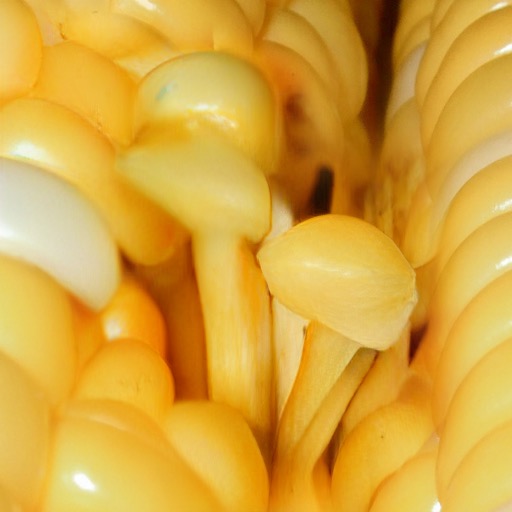}} 

\\

Lorikeet 
\texttt{[90]} &&
 \parbox[c]{\tmpwidth}{\includegraphics[ width=\tmpwidth]{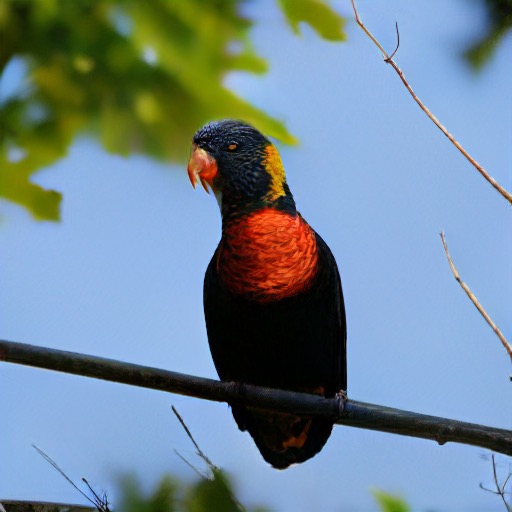}} & 
  \parbox[c]{\tmpwidth}{\includegraphics[ width=\tmpwidth]{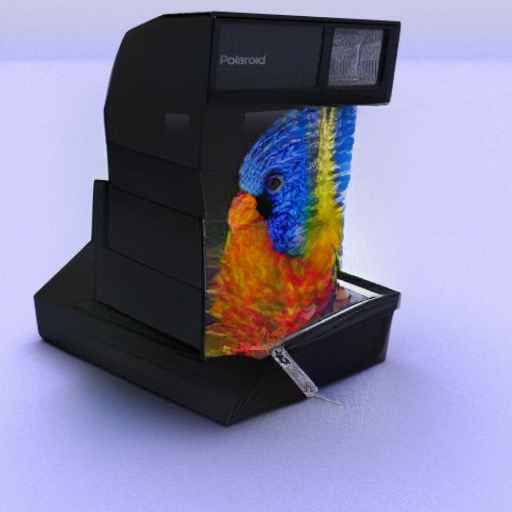}} & 
   \parbox[c]{\tmpwidth}{\includegraphics[ width=\tmpwidth]{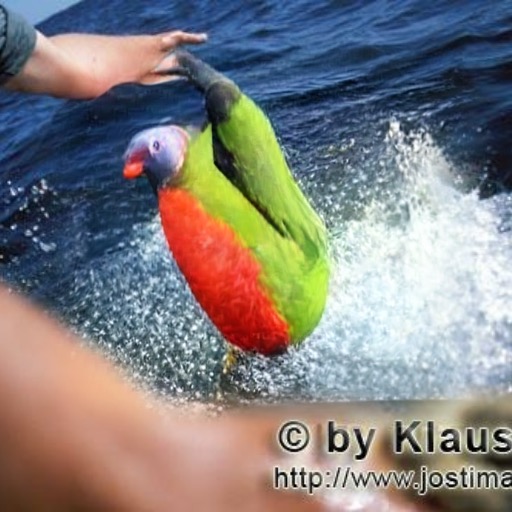}} & 
 \parbox[c]{\tmpwidth}{\includegraphics[ height=\tmpwidth]{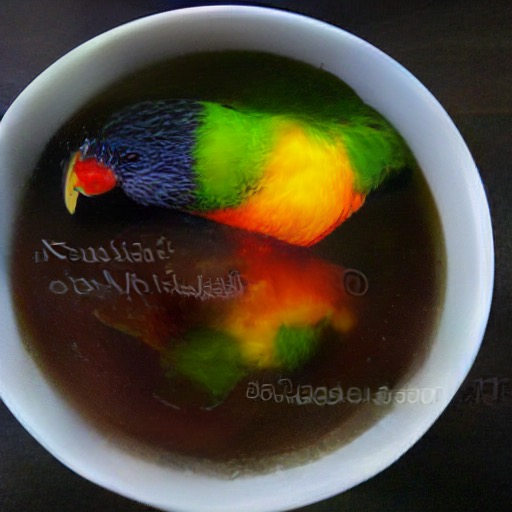}} & 
 \parbox[c]{\tmpwidth}{\includegraphics[ width=\tmpwidth]{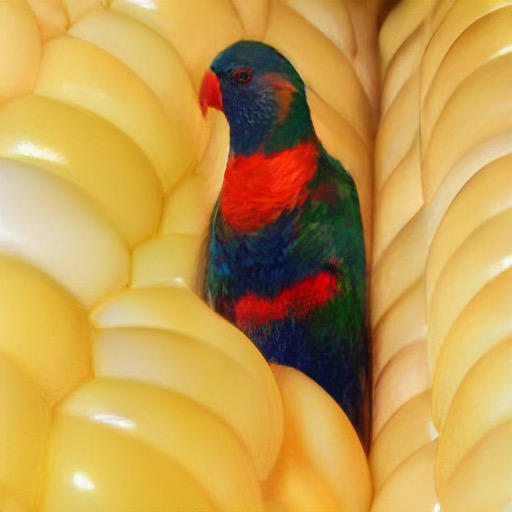}}

\\

% Red Fox \texttt{[145]} && 
%  \parbox[c]{\tmpwidth}{\includegraphics[ width=\tmpwidth]{figures/supp_class_modification/008857_to_001301_0000_output.jpeg}} & 
%  \parbox[c]{\tmpwidth}{\includegraphics[ width=\tmpwidth]{figures/supp_class_modification/000398_to_001303_0006_output.jpeg}} & 
%  \parbox[c]{\tmpwidth}{\includegraphics[ width=\tmpwidth]{figures/supp_class_modification/008784_to_001301_0003_output.jpeg}} & 
% \\

Train \texttt{[829]} && 
\parbox[c]{\tmpwidth}{\includegraphics[ width=\tmpwidth]{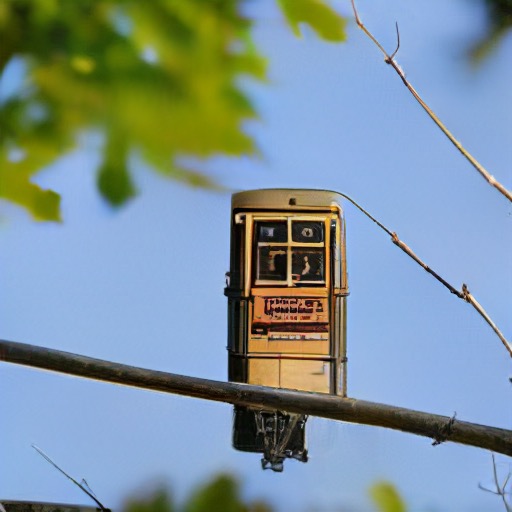}} & 
\parbox[c]{\tmpwidth}{\includegraphics[ width=\tmpwidth]{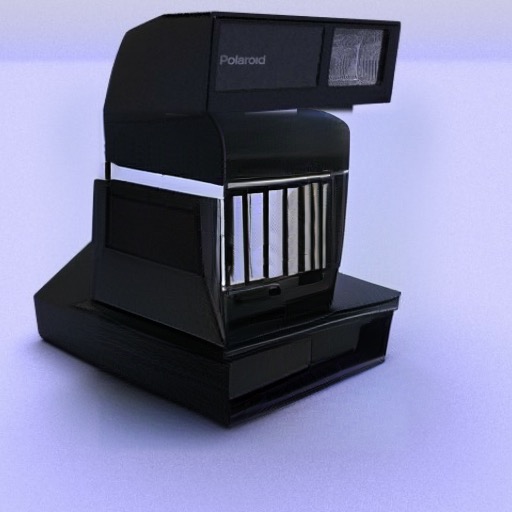}} & 
\parbox[c]{\tmpwidth}{\includegraphics[ width=\tmpwidth]{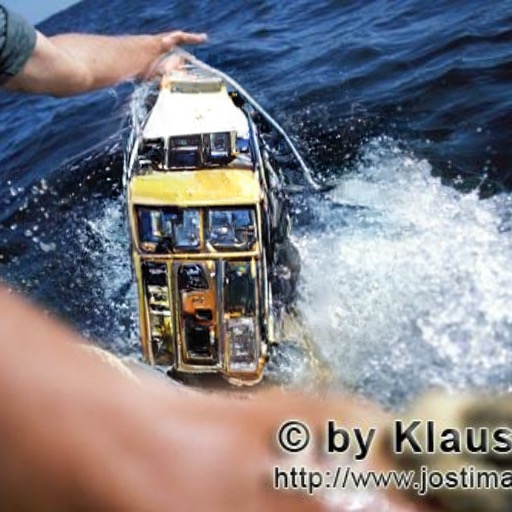}} & 
\parbox[c]{\tmpwidth}{\includegraphics[ height=\tmpwidth]{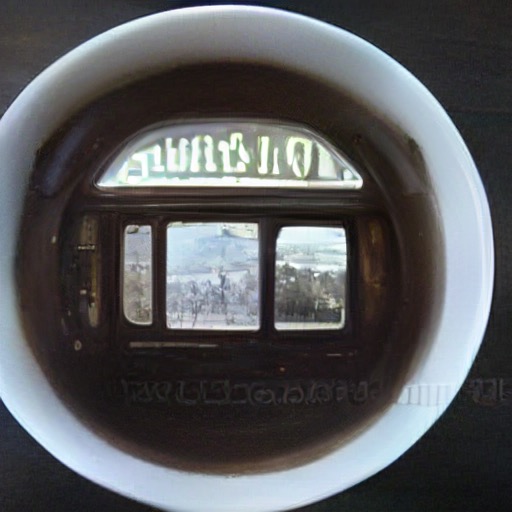}} & 
\parbox[c]{\tmpwidth}{\includegraphics[ width=\tmpwidth]{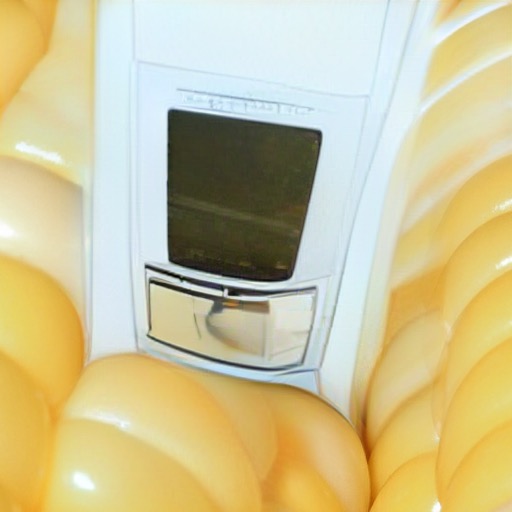}} 
\\

Tiger 
\texttt{[292]} && 
\parbox[c]{\tmpwidth}{\includegraphics[ width=\tmpwidth]{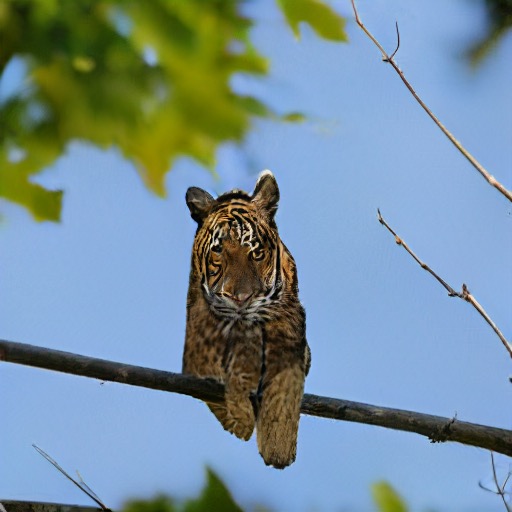}}&
\parbox[c]{\tmpwidth}{\includegraphics[ width=\tmpwidth]{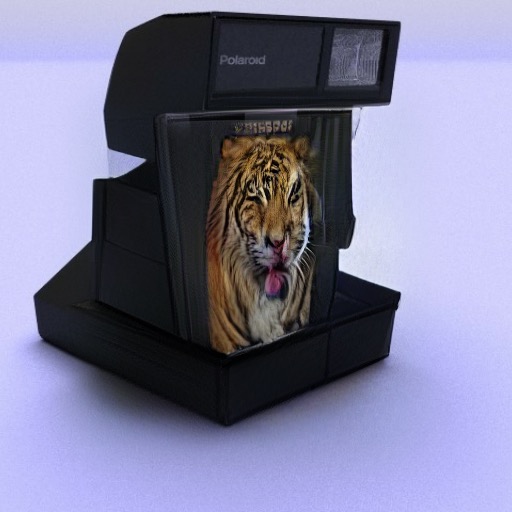}} & 
\parbox[c]{\tmpwidth}{\includegraphics[ width=\tmpwidth]{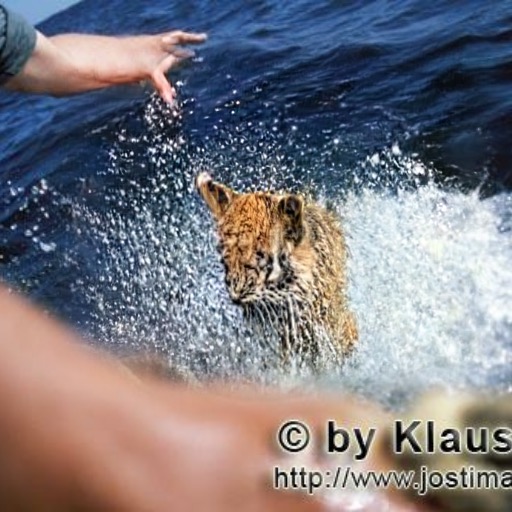}} & 
\parbox[c]{\tmpwidth}{\includegraphics[ height=\tmpwidth]{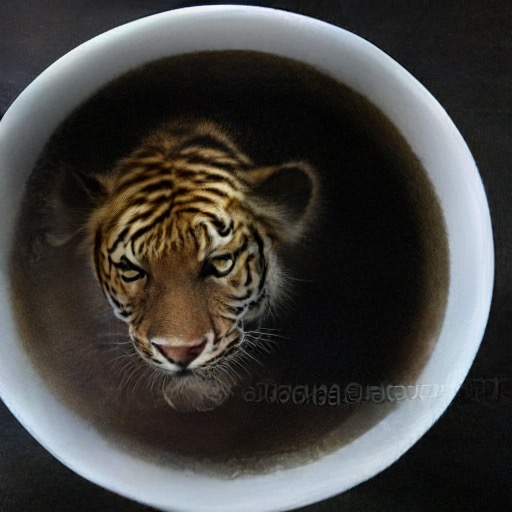}} & 
\parbox[c]{\tmpwidth}{\includegraphics[ width=\tmpwidth]{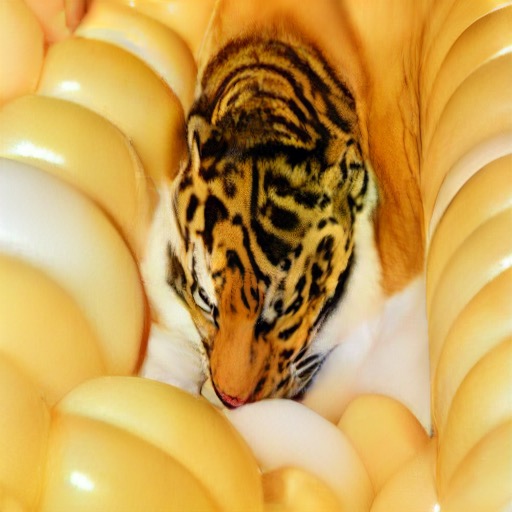}} 

\\

%  \parbox[c]{\tmpwidth}{\includegraphics[ width=\tmpwidth]{figures/supp_class_modification/000398_to_001320_0002_output.jpeg}} & 
%  \parbox[c]{\tmpwidth}{\includegraphics[ width=\tmpwidth]{figures/supp_class_modification/corn_to_001320_0018_output.jpeg}} &

    \end{tabular}
    \caption{\textbf{More Examples of Class-conditional Image Editing.} In each column, the bottom images are synthesized using the image on the top, ImageNet class labels on the left, and a bounding box of the main object downsampled into latent space (as shown in the second row).}
    \vspace{-12mm}
    \label{fig:supp_editing}
\end{figure*}

\renewcommand{\tmpwidth}{32mm}
\setlength{\tabcolsep}{2pt}
\begin{figure*}[!ht]
    \centering
    \begin{tabular}{cc}
Input & \model (Ours)
\\ 

\includegraphics[ height=\tmpwidth]{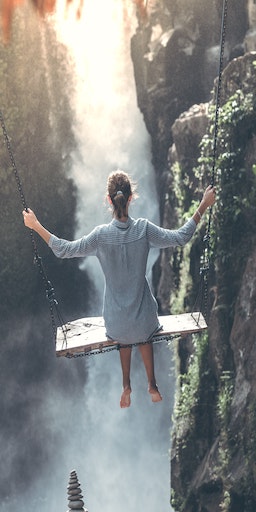} &
\includegraphics[ height=\tmpwidth]{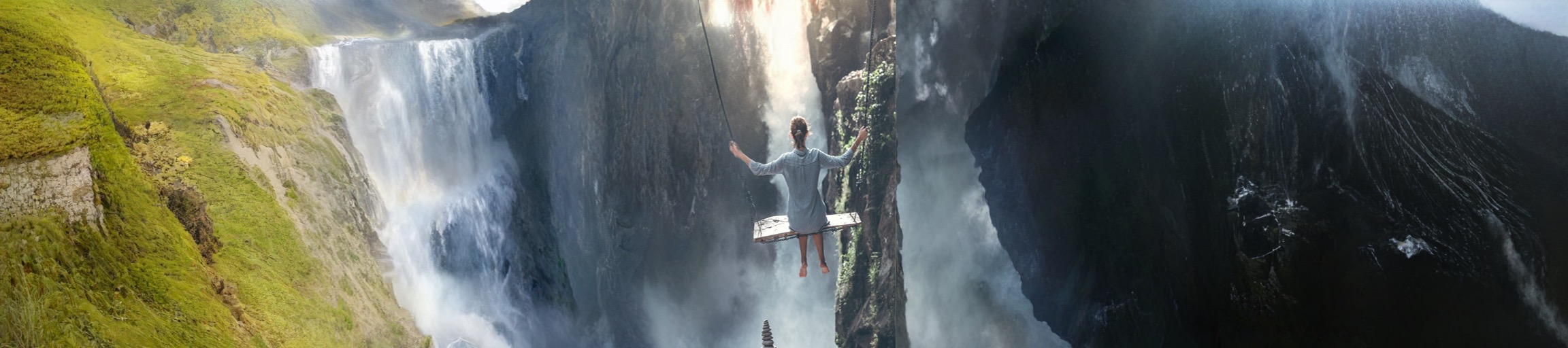} \\

\includegraphics[ height=\tmpwidth]{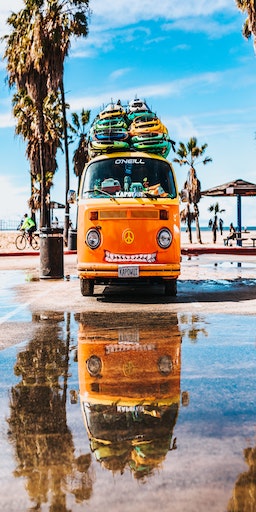} &
\includegraphics[ height=\tmpwidth]{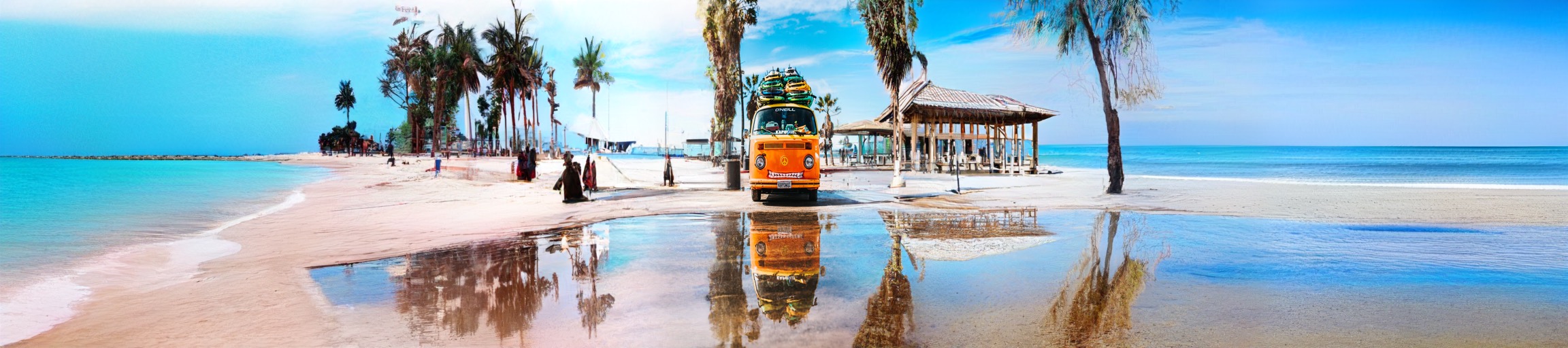} \\ 
\includegraphics[ height=\tmpwidth]{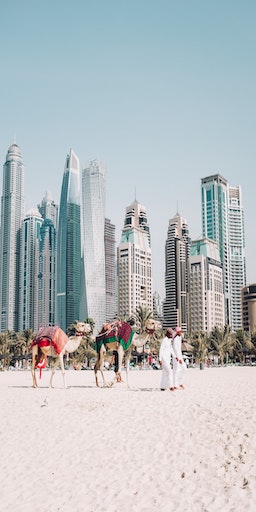} &
\includegraphics[ height=\tmpwidth]{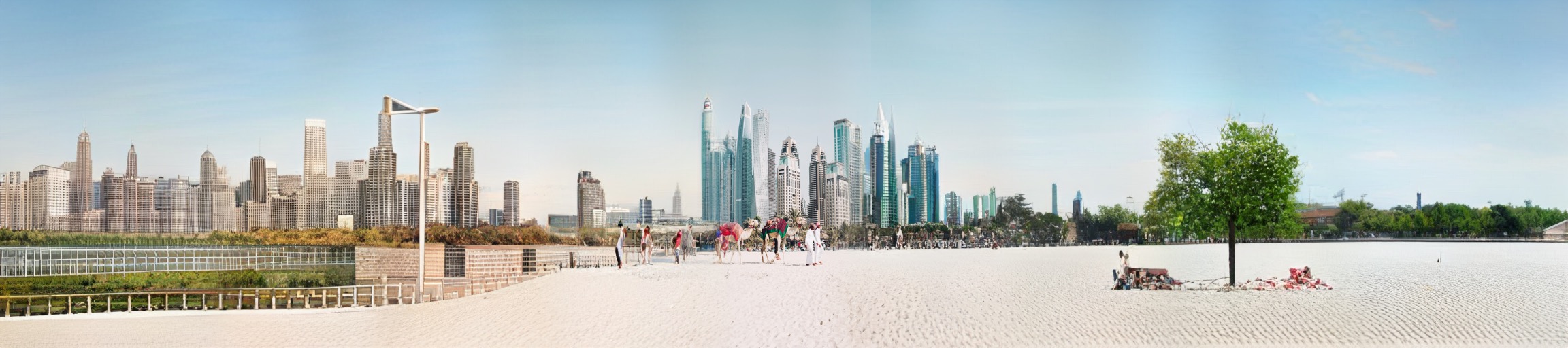} \\ 

\includegraphics[ height=\tmpwidth]{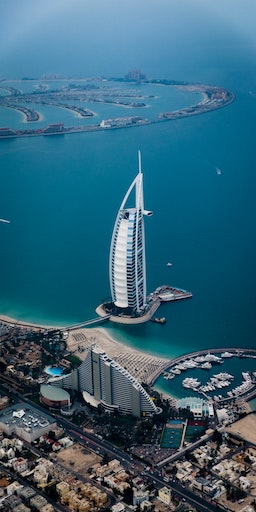} &
\includegraphics[ height=\tmpwidth]{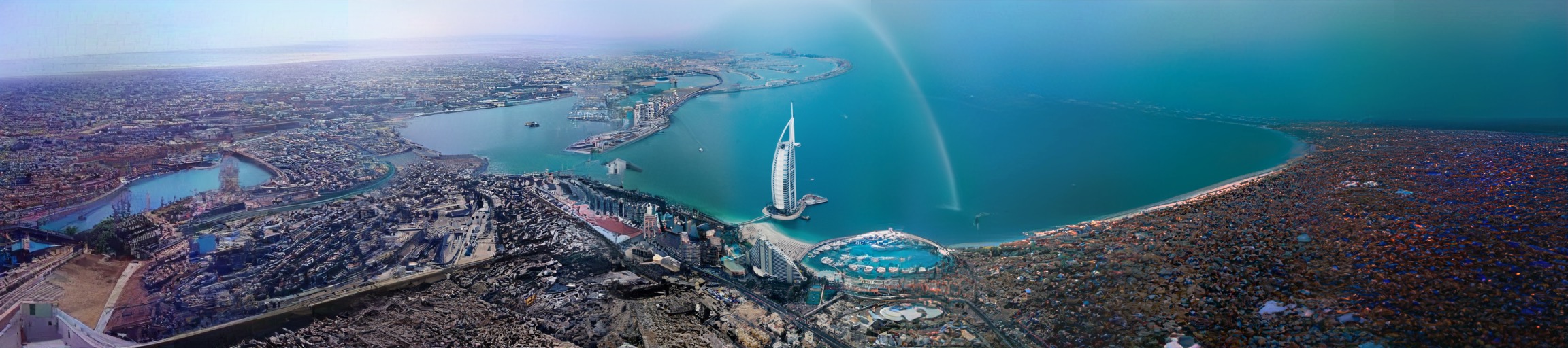} \\ 
\includegraphics[ height=\tmpwidth]{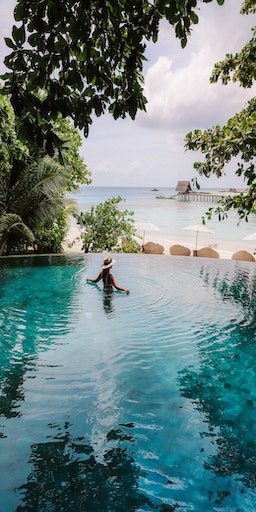} &
\includegraphics[ height=\tmpwidth]{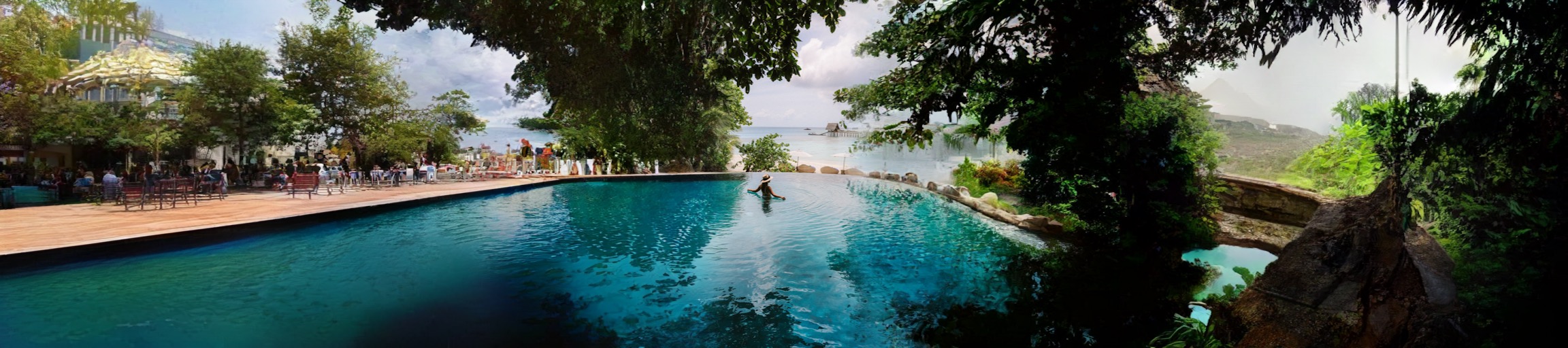} \\ 
\includegraphics[ height=\tmpwidth]{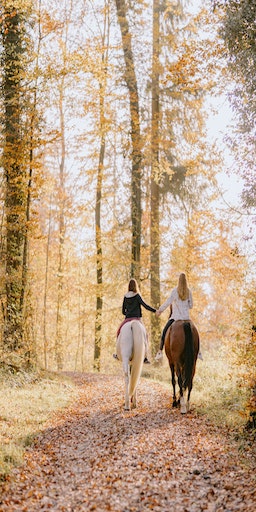} &
\includegraphics[ height=\tmpwidth]{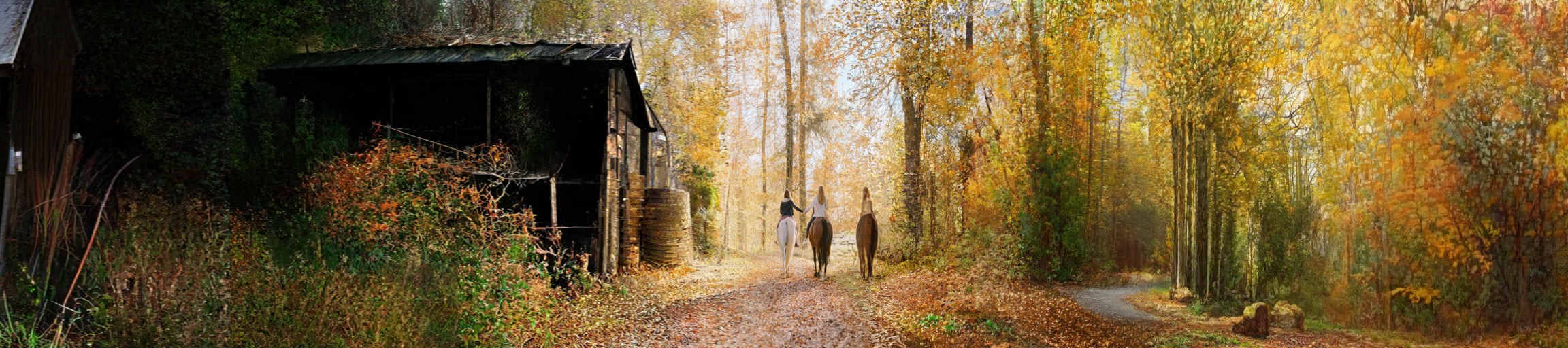} \\ 

    \end{tabular}
    \vspace{-3mm}
    \caption{More Samples of Horizontal Image Extrapolation (from 512$\times$256 to 512$\times$2304). The synthesized "panoramas" are created by repeatedly applying \model's outpainting abilities horizontally in both directions.}
    %\vspace{-6mm}
    \label{fig:supp_panorama}
\end{figure*}

We show more examples of class-conditional image editing in Figure~\ref{fig:supp_editing}, and examples of image-conditional panorama synthesis in Figure~\ref{fig:supp_panorama}. 

\suppsection{Image Outpainting Comparisons with SOTA Transformer-based Approaches}
\label{sec:supp_outpainting_comparison_with_transformers}
In Figure~\ref{fig:supp_uncropgpt-1} and \ref{fig:supp_uncropgpt-2}, we show a few outpainting comparisons among \model, ImageGPT\cite{chen2020imagegpt}, and VQGAN\cite{Esser21vqgan}. In each set of images, we show the groundtruth (left), extrapolated samples using only the top half of the groundtruth (middle), and extrapolated samples using only the bottom half of the groundtruth (right). 

\model and VQGAN can both perform on higher resolutions by taking advantage of tokenization and thus achieve higher sample fidelity than ImageGPT, which runs on a maximum resolution of $192\times192$. At the same time, \model demonstrates stronger flexibility than ImageGPT and VQGAN in that it can outpaint in arbitrary directions (\eg both upward and downward), while ImageGPT and VQGAN can only handle outpainting in one direction with a single model due to their autoregressive natures.

\renewcommand{\tmpwidth}{22mm}
\setlength{\tabcolsep}{2pt}
\begin{figure*}[!ht]
    \centering
    \begin{tabular}{cc cccc ccc}
% \multirow{5}{*}{\includegraphics[width=\tmpwidth]{figures/supp_uncropgpt/1_input.jpg}}
\toprule
Groundtruth  & 
 &\multicolumn{3}{c}{ ------ Outpaint bottom 50\% ------ } & &\multicolumn{3}{c}{ ------ Outpaint top 50\% ------ }  
\\
& 
 \raisebox{0.85\tmpwidth }{\rotatebox[origin=l]{90}{ ImageGPT\cite{chen2020imagegpt}}} & 

\includegraphics[width=\tmpwidth]{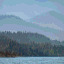}&
\includegraphics[width=\tmpwidth]{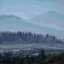}&
\includegraphics[width=\tmpwidth]{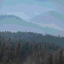} &
\\

\includegraphics[width=\tmpwidth]{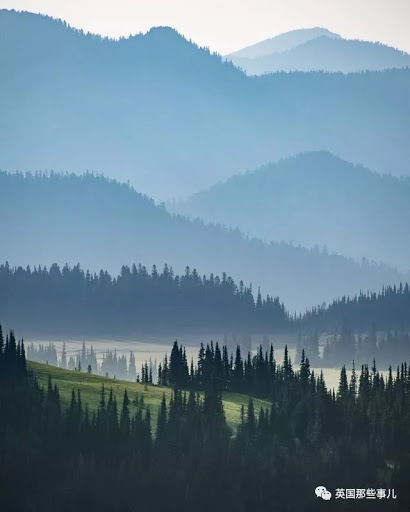} &
 \raisebox{0.85\tmpwidth }{\rotatebox[origin=l]{90}{ VQGAN\cite{Esser21vqgan}}} & 

\includegraphics[width=\tmpwidth]{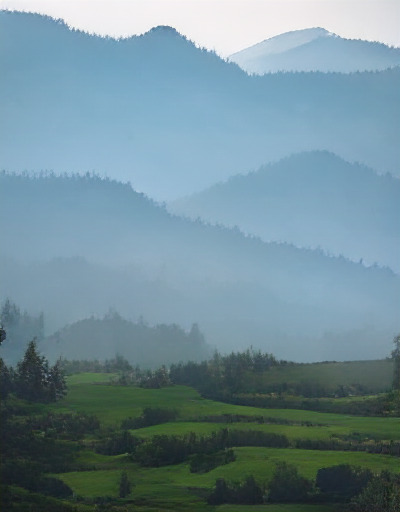}&
\includegraphics[width=\tmpwidth]{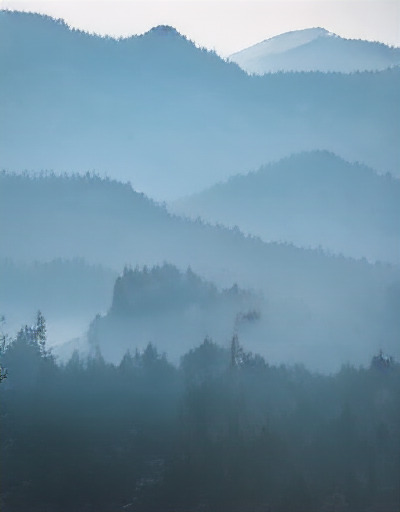}&
\includegraphics[width=\tmpwidth]{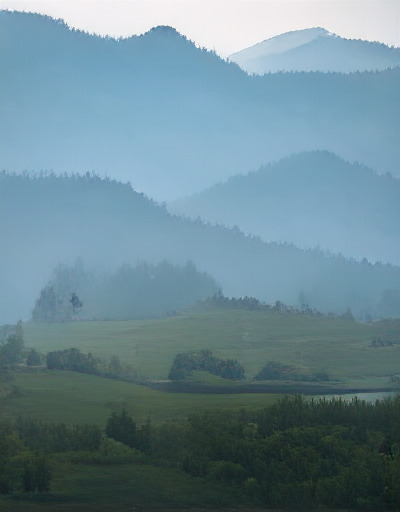} &
\\
& 
 \raisebox{0.85\tmpwidth }{\rotatebox[origin=l]{90}{\model (Ours)}} & 
\includegraphics[width=\tmpwidth]{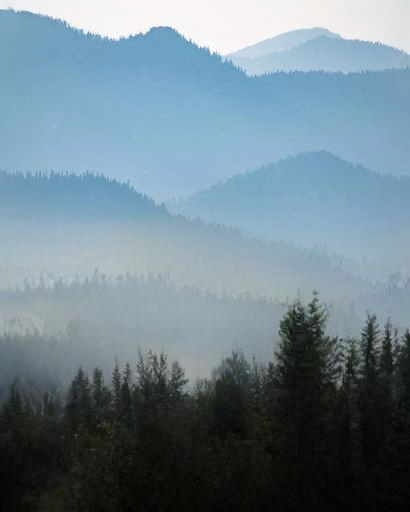}&
\includegraphics[width=\tmpwidth]{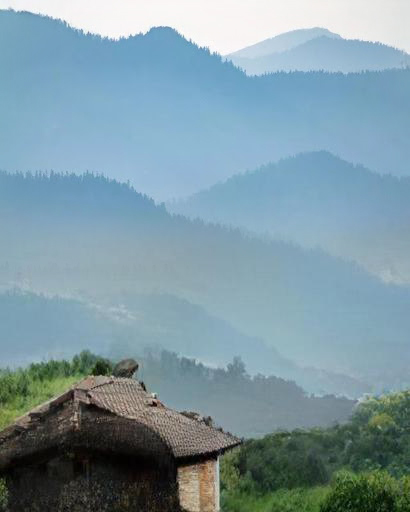}&
\includegraphics[width=\tmpwidth]{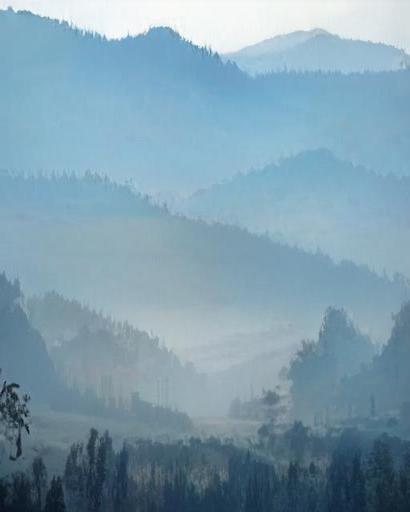} &
&
\includegraphics[width=\tmpwidth]{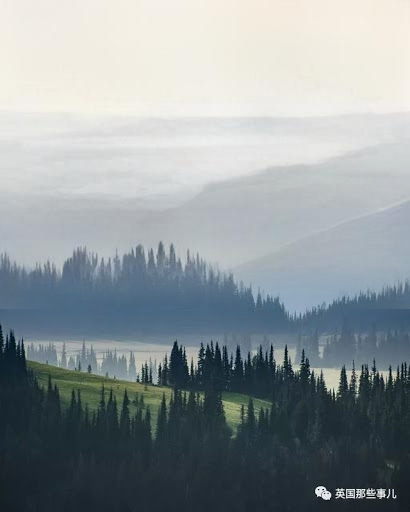}&
\includegraphics[width=\tmpwidth]{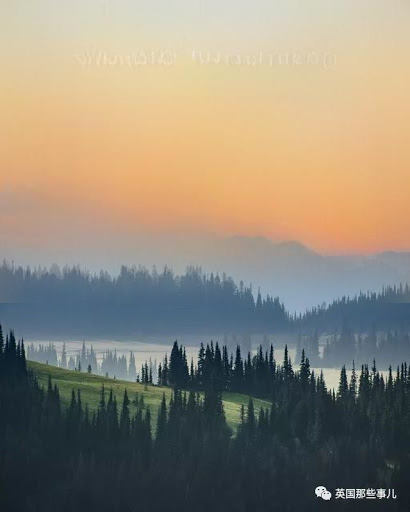}&
\includegraphics[width=\tmpwidth]{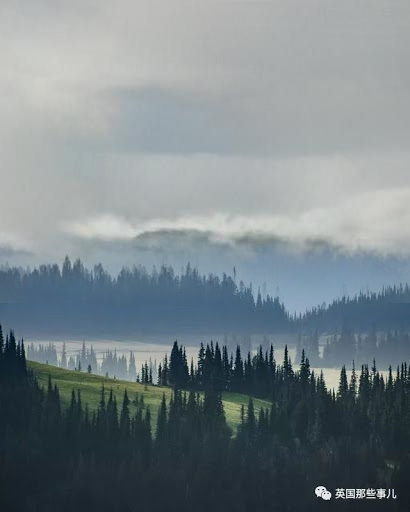}

% &\multicolumn{3}{c}{--- MaskGIT's Samples (Outpainting Bottom Half) ---} &\multicolumn{3}{c}{--- MaskGIT's Samples (Outpainting Top Half) ---}  
\\
& 
 \raisebox{0.85\tmpwidth }{\rotatebox[origin=l]{90}{ ImageGPT\cite{chen2020imagegpt}}} & 

\includegraphics[width=\tmpwidth]{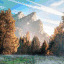}&
\includegraphics[width=\tmpwidth]{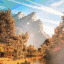}&
\includegraphics[width=\tmpwidth]{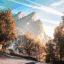} &
\\
% & \multicolumn{3}{c}{------ ImageGPT\cite{chen2020imagegpt}'s Samples ------}  &
% \\
\includegraphics[width=\tmpwidth]{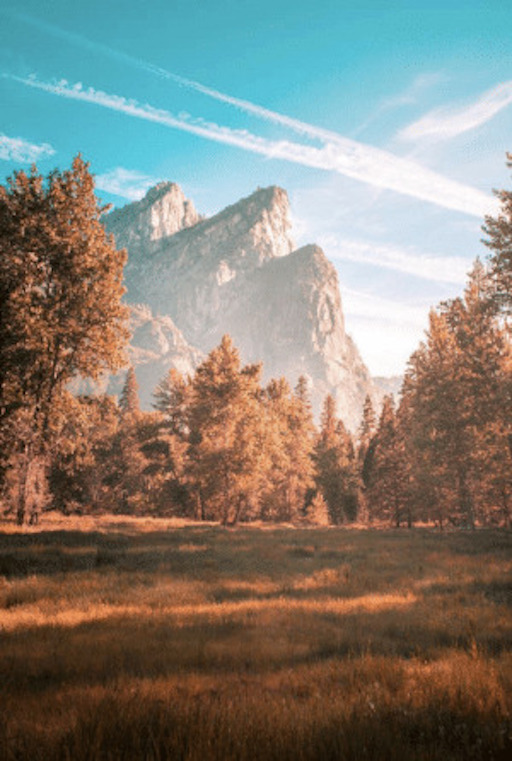} &
 \raisebox{0.85\tmpwidth }{\rotatebox[origin=l]{90}{VQGAN\cite{Esser21vqgan}}} & 
\includegraphics[width=\tmpwidth]{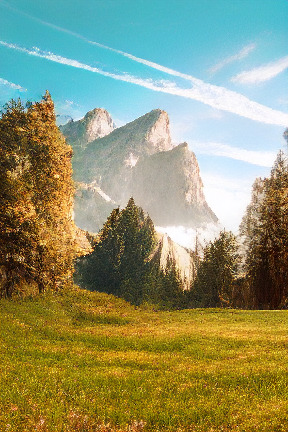}&
\includegraphics[width=\tmpwidth]{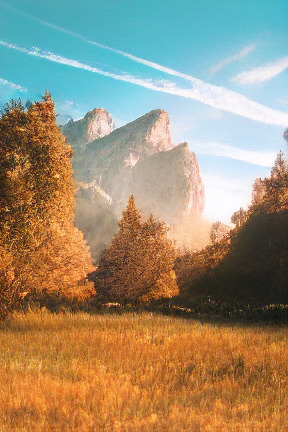}&
\includegraphics[width=\tmpwidth]{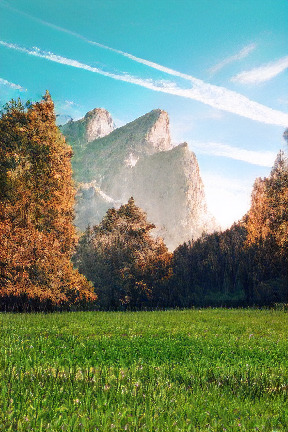} &
\\
& 
 \raisebox{0.85\tmpwidth }{\rotatebox[origin=l]{90}{\model (Ours)}} & 
 \includegraphics[width=\tmpwidth]{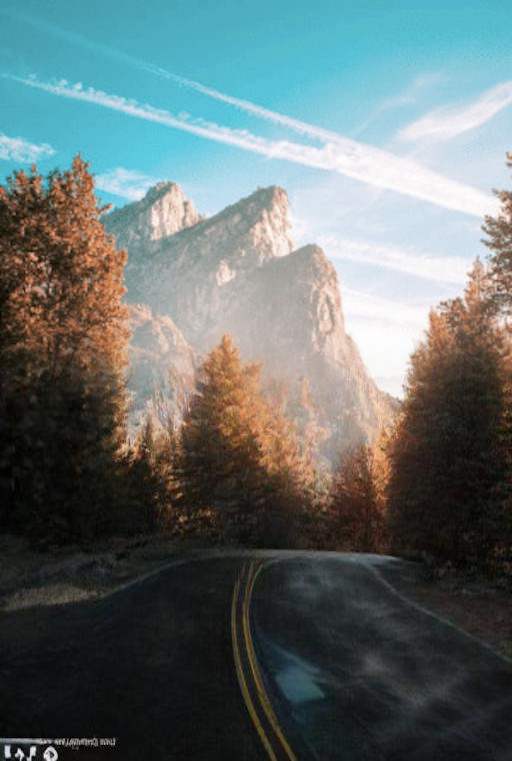}&
\includegraphics[width=\tmpwidth]{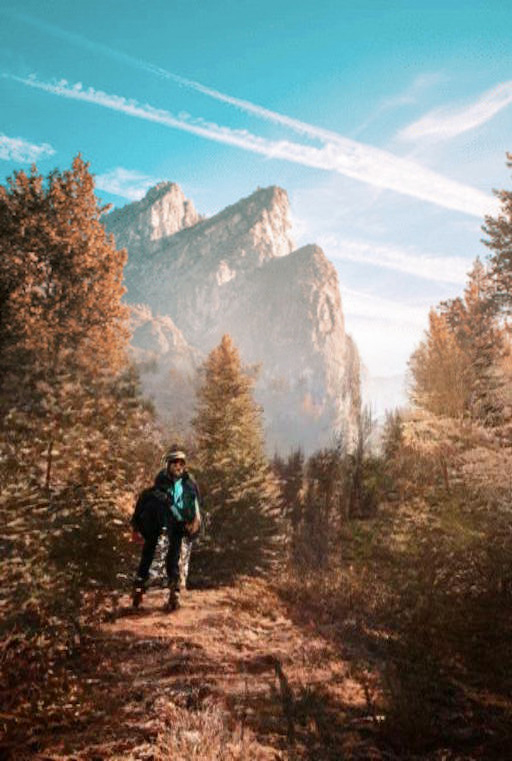}&
\includegraphics[width=\tmpwidth]{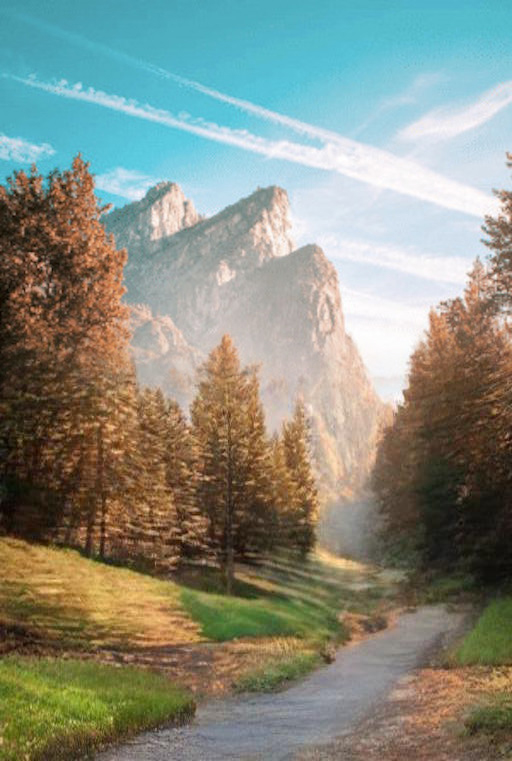} &
&
 \includegraphics[width=\tmpwidth]{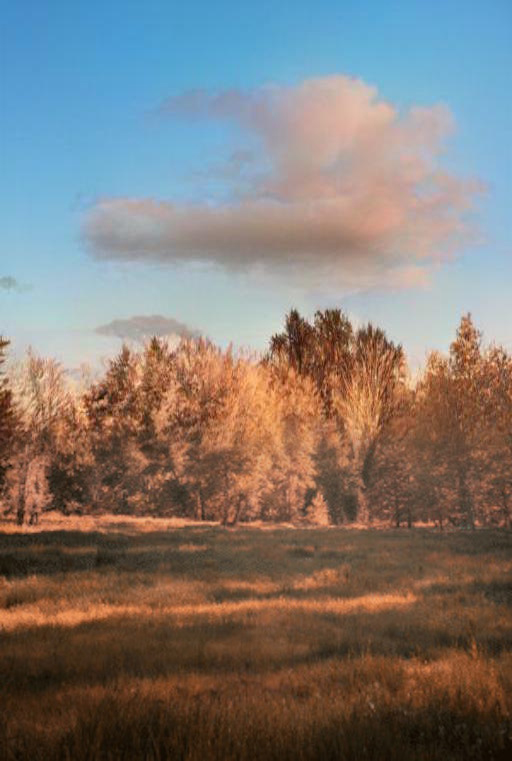}&
\includegraphics[width=\tmpwidth]{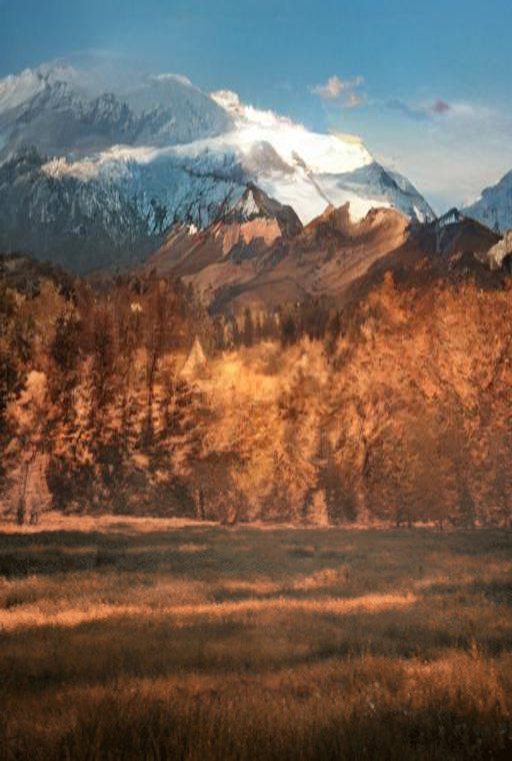}&
\includegraphics[width=\tmpwidth]{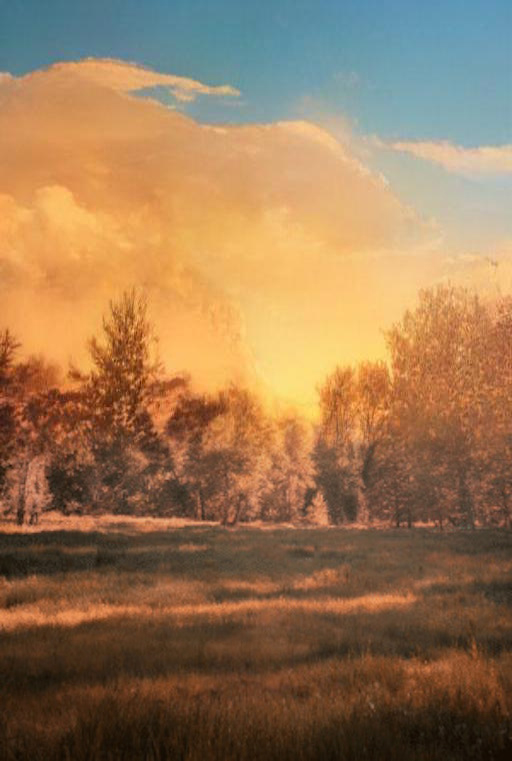}
\\
      
    \end{tabular}
    \vspace{-2mm}
    \caption{Outpainting comparisons with the pixel-based approach ImageGPT\cite{chen2020imagegpt} and the transformer-based approach VQGAN\cite{Esser21vqgan}.
    }
    % \vspace{-10mm}
    \label{fig:supp_uncropgpt-1}
\end{figure*}

\renewcommand{\tmpwidth}{22mm}
\setlength{\tabcolsep}{2pt}
\begin{figure*}[!ht]
    \centering
    \begin{tabular}{cc cccc ccc}
\toprule
Groundtruth  & 
 &\multicolumn{3}{c}{ ------ Outpaint bottom 50\% ------ } & &\multicolumn{3}{c}{ ------ Outpaint top 50\% ------ }  
\\
& 
 \raisebox{0.85\tmpwidth }{\rotatebox[origin=l]{90}{ ImageGPT\cite{chen2020imagegpt}}} & 

\includegraphics[width=\tmpwidth]{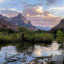}&
\includegraphics[width=\tmpwidth]{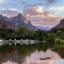}&
\includegraphics[width=\tmpwidth]{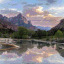}&
% \\
% & \multicolumn{3}{c}{------ ImageGPT\cite{chen2020imagegpt}'s Samples ------}  &
\\
\includegraphics[width=\tmpwidth]{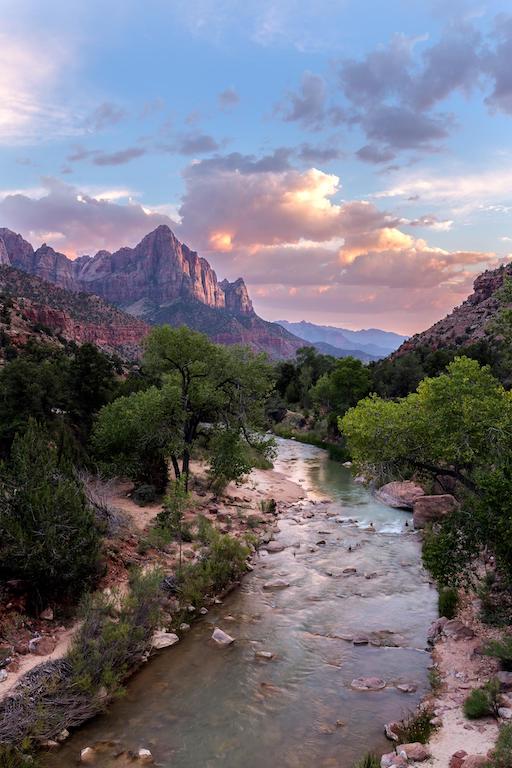} &
 \raisebox{0.85\tmpwidth }{\rotatebox[origin=l]{90}{ VQGAN\cite{Esser21vqgan}}} & 
\includegraphics[width=\tmpwidth]{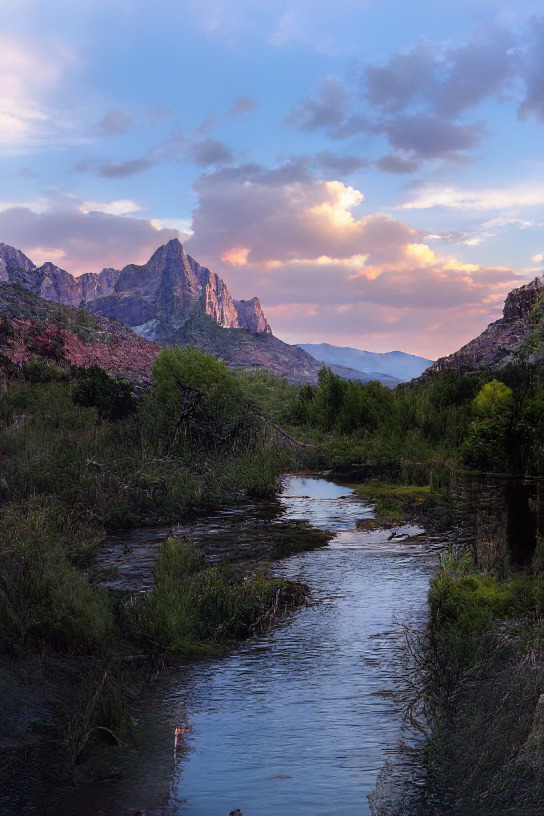}&
\includegraphics[width=\tmpwidth]{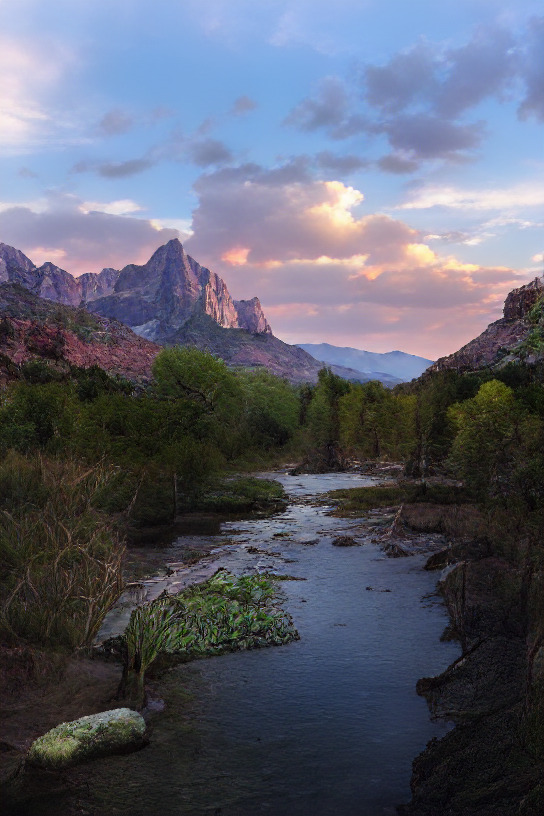}&
\includegraphics[width=\tmpwidth]{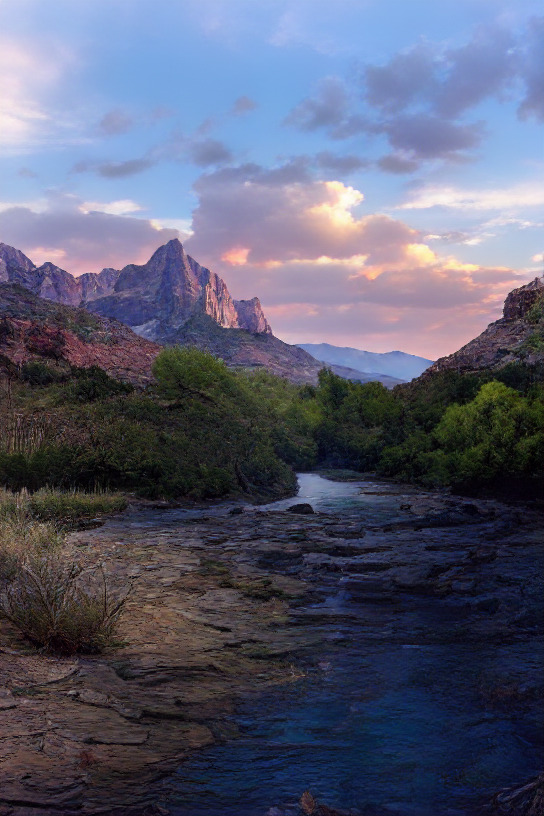} &
\\
% Groundtruth  & \multicolumn{3}{c}{------ VQGAN\cite{Esser21vqgan}'s Samples ------}  &
% \\
% Groundtruth  & \multicolumn{3}{c}{------ VQGAN\cite{Esser21vqgan}'s Samples ------}  &
% Groundtruth  & \multicolumn{3}{c}{------ VQGAN\cite{Esser21vqgan}'s Samples ------}  
& 
 \raisebox{0.85\tmpwidth }{\rotatebox[origin=l]{90}{\model (Ours)}} & 
\includegraphics[width=\tmpwidth]{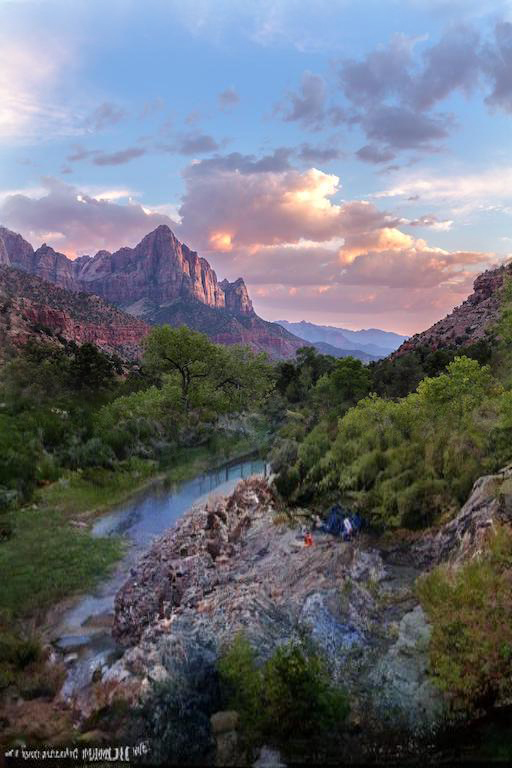}&
\includegraphics[width=\tmpwidth]{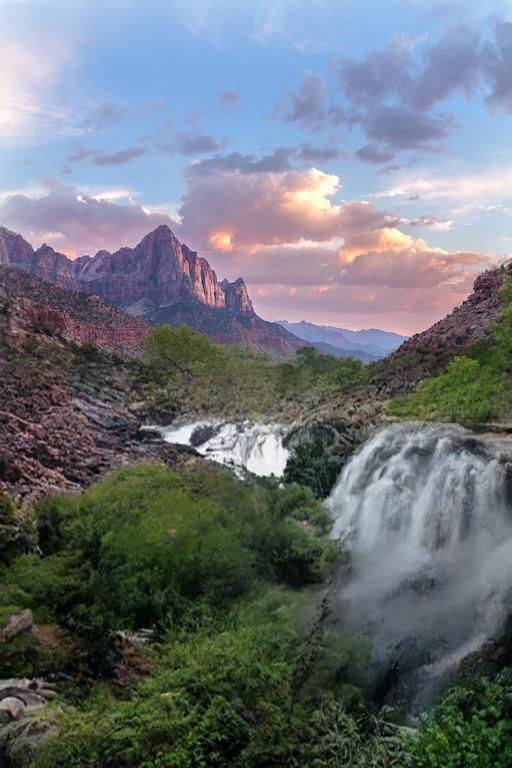}&
\includegraphics[width=\tmpwidth]{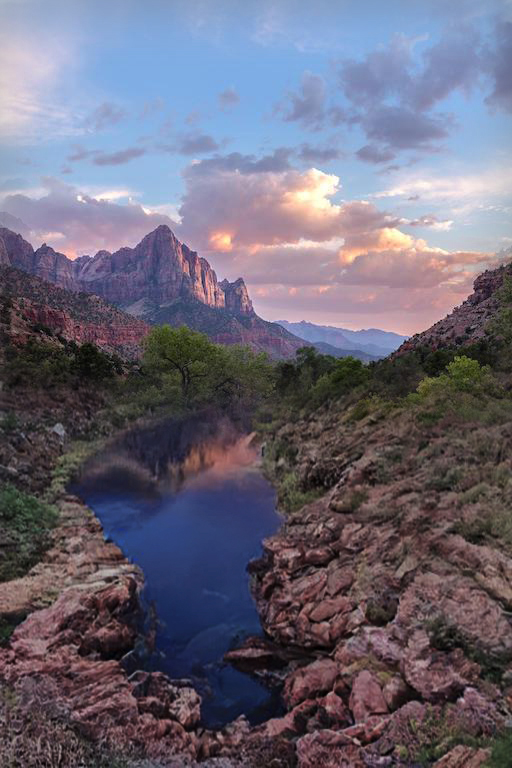} &
&
\includegraphics[width=\tmpwidth]{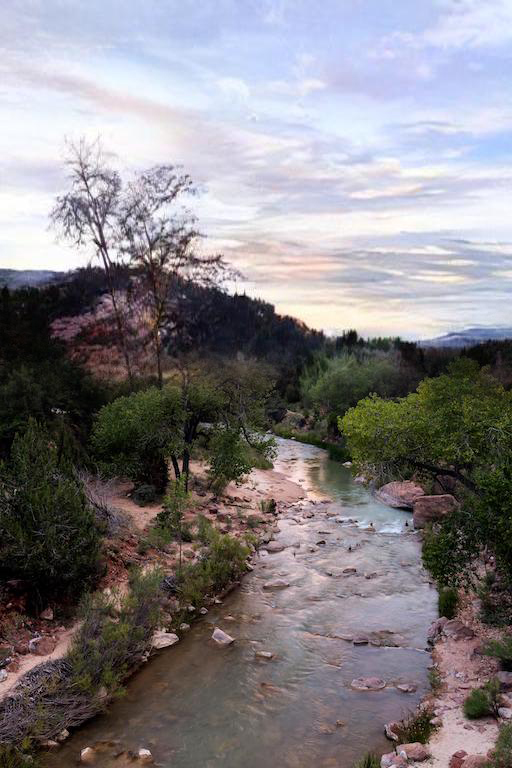}&
\includegraphics[width=\tmpwidth]{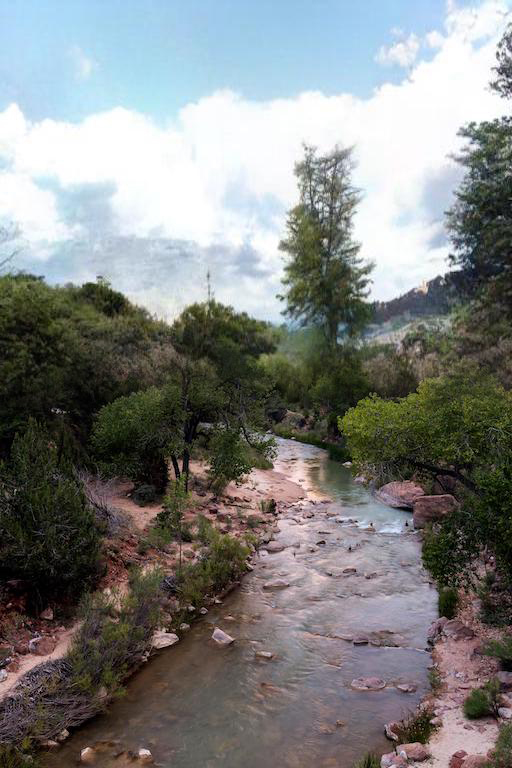}&
\includegraphics[width=\tmpwidth]{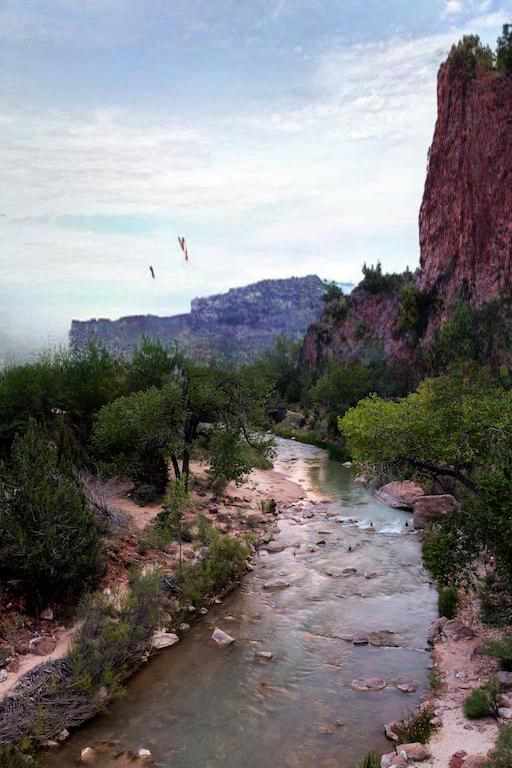}

\\
%  &\multicolumn{3}{c}{--- MaskGIT's Samples (Outpainting Bottom Half) ---} &\multicolumn{3}{c}{--- MaskGIT's Samples (Outpainting Top Half) ---}  
& 
 \raisebox{0.85\tmpwidth }{\rotatebox[origin=l]{90}{ ImageGPT\cite{chen2020imagegpt}}} & 
\includegraphics[width=\tmpwidth]{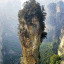}&
\includegraphics[width=\tmpwidth]{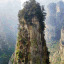}&
\includegraphics[width=\tmpwidth]{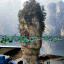} &
\\
% & \multicolumn{3}{c}{------ ImageGPT\cite{chen2020imagegpt}'s Samples ------}  &
% \\
\includegraphics[width=\tmpwidth]{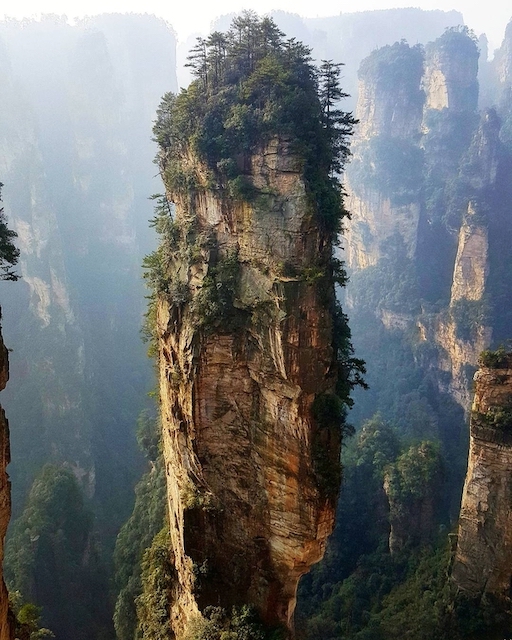} &
 \raisebox{0.85\tmpwidth }{\rotatebox[origin=l]{90}{VQGAN\cite{Esser21vqgan}}} & 
\includegraphics[width=\tmpwidth]{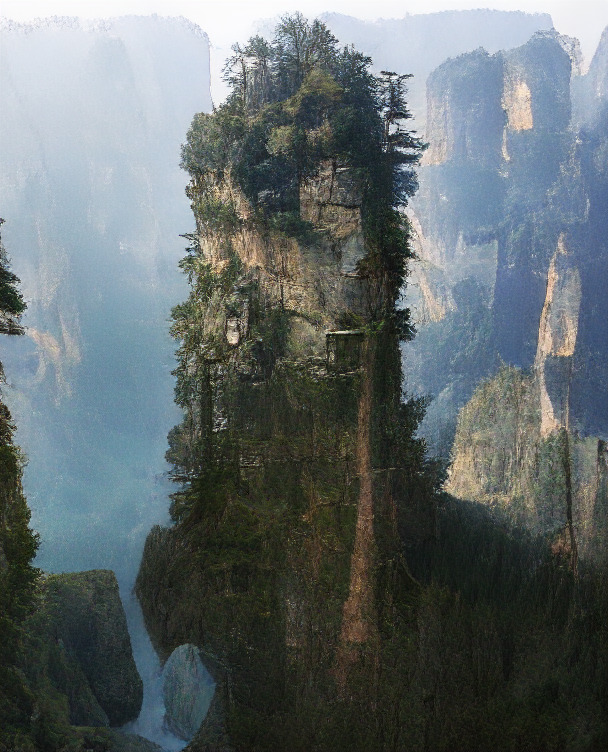}&
\includegraphics[width=\tmpwidth]{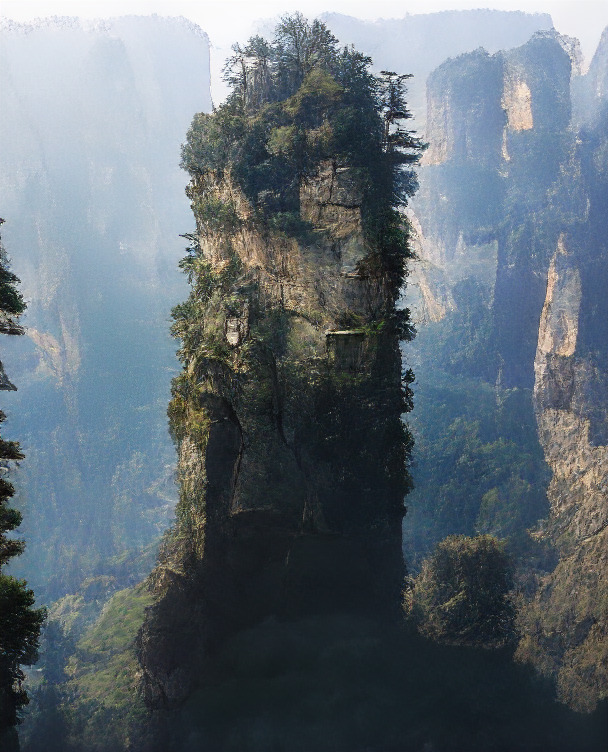}&
\includegraphics[width=\tmpwidth]{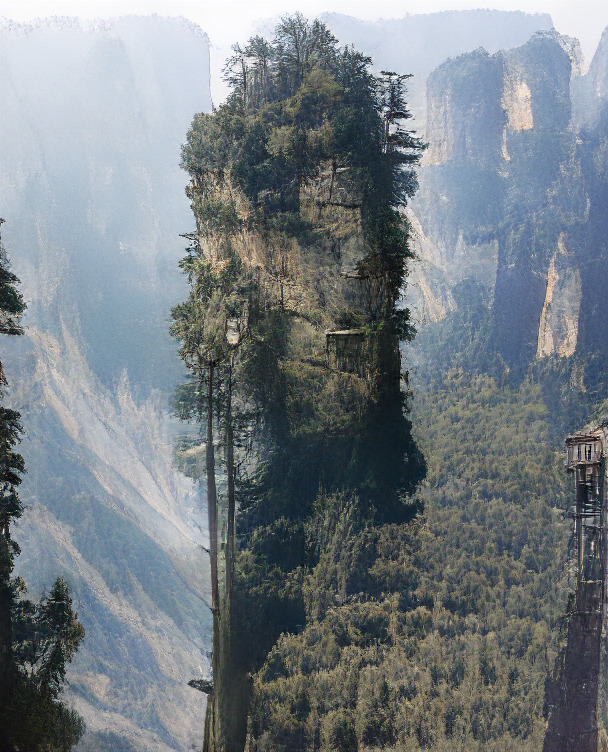} &
 \\   
% Groundtruth  & \multicolumn{3}{c}{------ VQGAN\cite{Esser21vqgan}'s Samples ------}  &
% \\

& 
 \raisebox{0.85\tmpwidth }{\rotatebox[origin=l]{90}{\model (Ours)}} & 
\includegraphics[width=\tmpwidth]{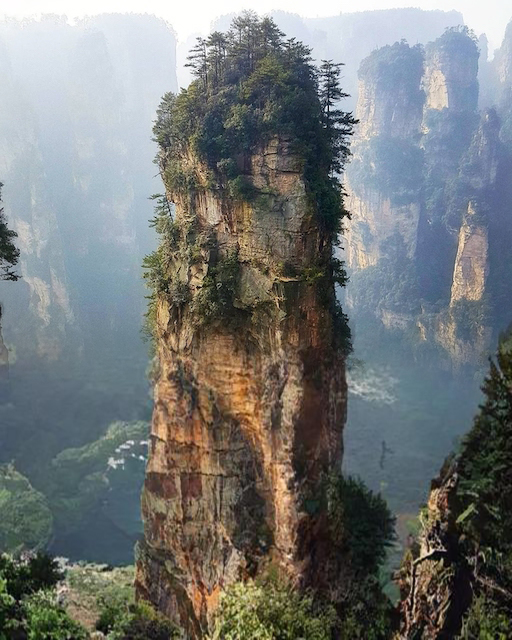}&
\includegraphics[width=\tmpwidth]{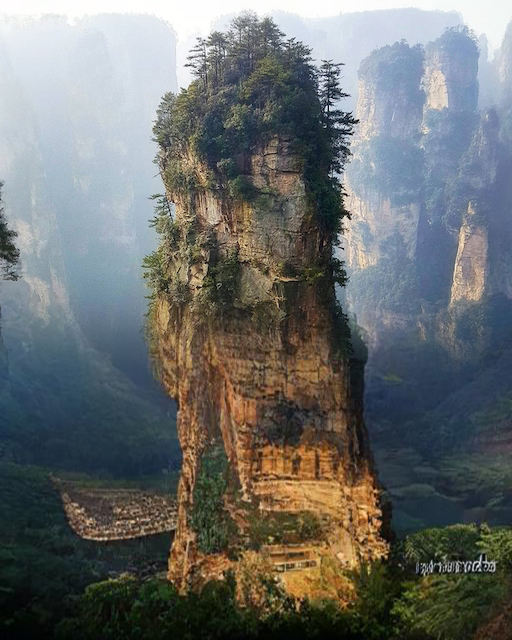}&
\includegraphics[width=\tmpwidth]{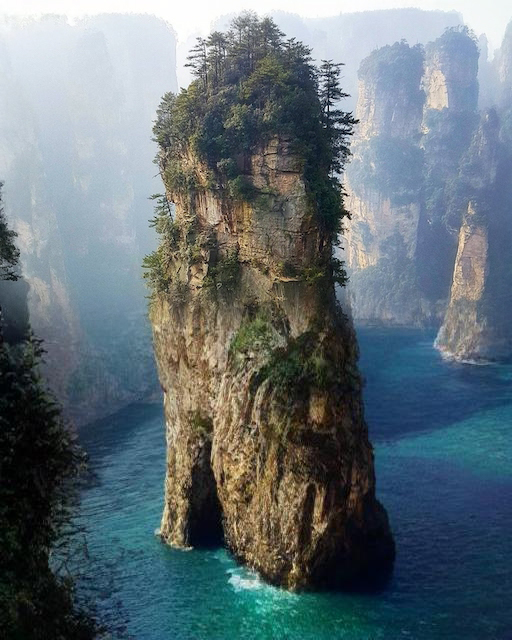} &
&
\includegraphics[width=\tmpwidth]{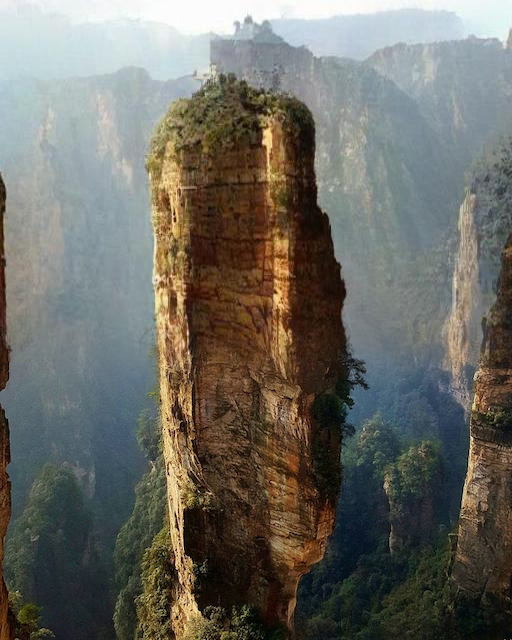}&
\includegraphics[width=\tmpwidth]{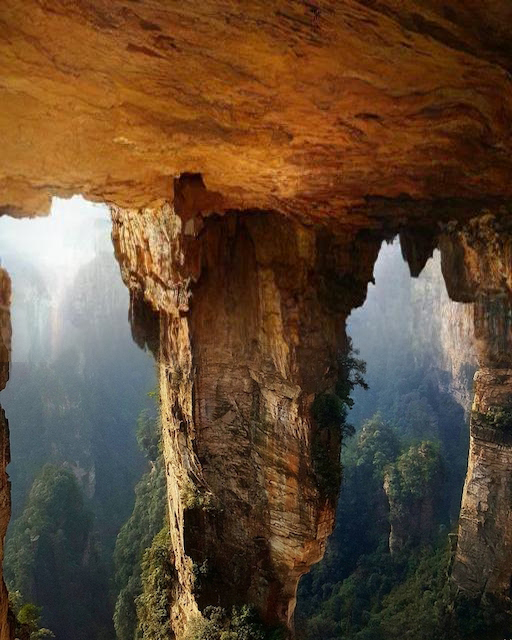}&
\includegraphics[width=\tmpwidth]{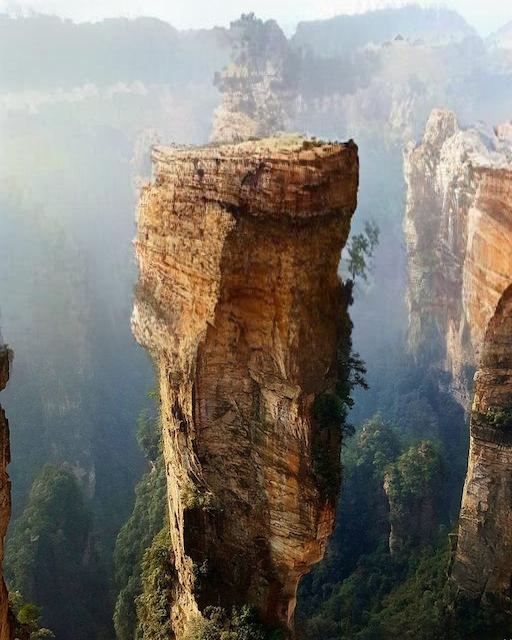}
 \\
% &\multicolumn{3}{c}{--- MaskGIT's Samples (Outpainting Bottom Half) ---} &\multicolumn{3}{c}{--- MaskGIT's Samples (Outpainting Top Half) ---}  
% \\ 
       
    \end{tabular}
    \caption{Outpainting comparisons with the pixel-based approach ImageGPT\cite{chen2020imagegpt} and the transformer-based approach VQGAN\cite{Esser21vqgan}.
    }
    % \vspace{-6mm}
    \label{fig:supp_uncropgpt-2}
\end{figure*}

\suppsection{Image Inpainting and Outpainting Comparisons with SOTA GAN-based Approaches}
\label{sec:supp_inpainting_and_outpainting_comparison_with_gans}

In this section, we show more qualitative comparisons with state-of-the-art GAN-based image completion methods in Figure~\ref{fig:supp_inpainting} and Figure~\ref{fig:supp_uncrop_comparison_right}. Quantitative results have been discussed in \ref{ssec:inpainting}. 

We find that compared to prior GAN-based methods, \model demonstrates a stronger capability of completing structures coherently, and its samples contain fewer artifacts. In Figure~\ref{fig:supp_uncrop_comparison_right}, \model completes the bridge in row two and the building in the second to last row, which all GAN methods struggle to do in comparison. 

\renewcommand{\tmpwidth}{27mm}
\setlength{\tabcolsep}{1pt}

\begin{figure*}[!ht]
    \centering
    \begin{tabular}{cccccc}

    Input & DeepFillv2\cite{yu2019free}  & HiFill\cite{yi2020contextual}  & CoModGAN\cite{zhao2021comodgan} & \textbf{\model (Ours)} & Groundtruth \\   
\includegraphics[width=\tmpwidth]{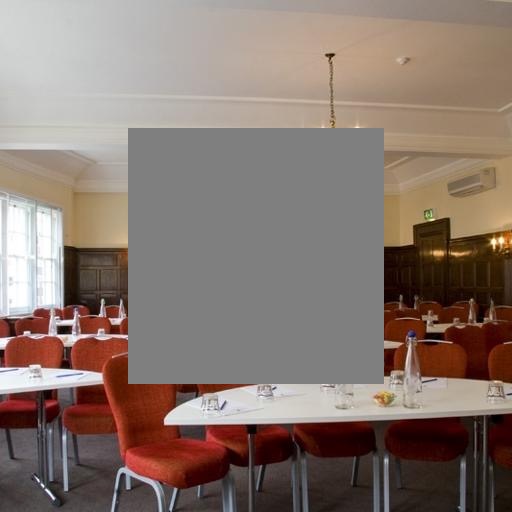}&
\includegraphics[width=\tmpwidth]{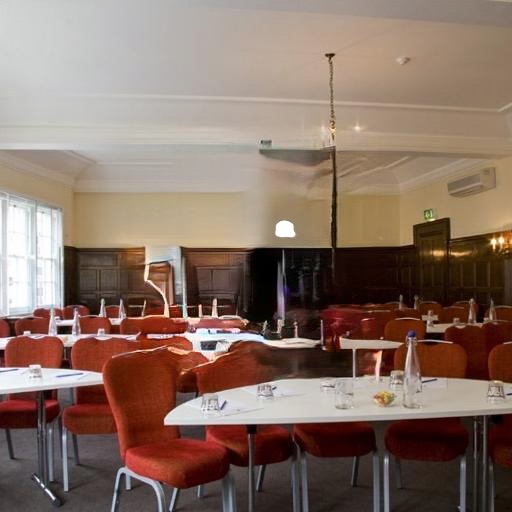}&
\includegraphics[width=\tmpwidth]{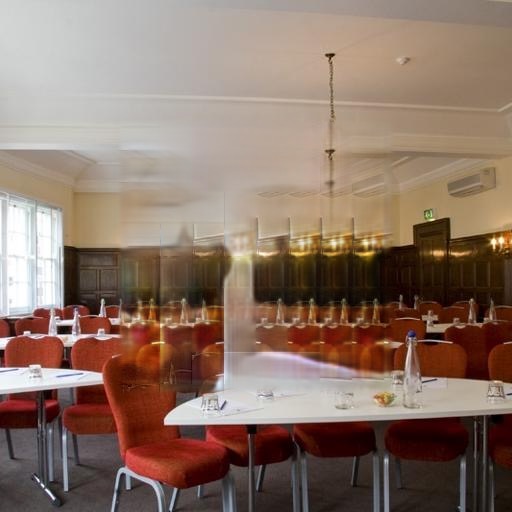}&
\includegraphics[width=\tmpwidth]{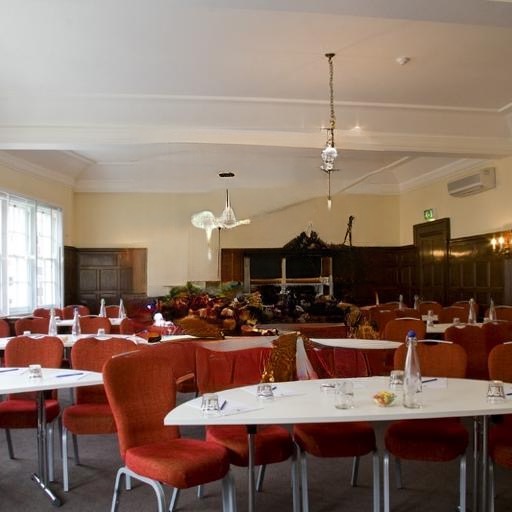}&
\includegraphics[width=\tmpwidth]{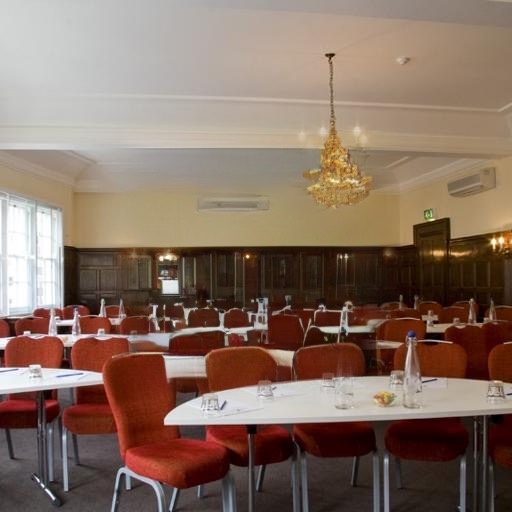}&
\includegraphics[width=\tmpwidth]{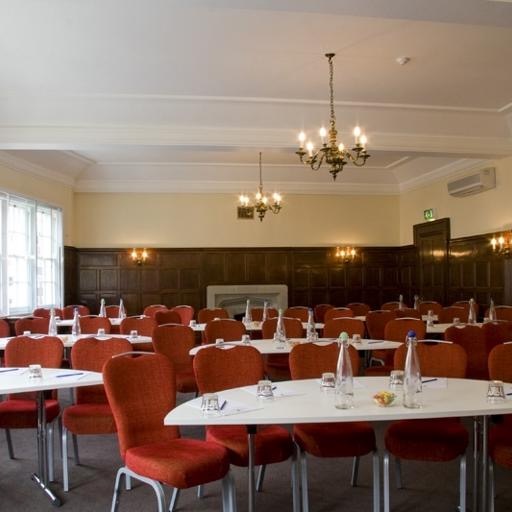}
\\

\includegraphics[width=\tmpwidth]{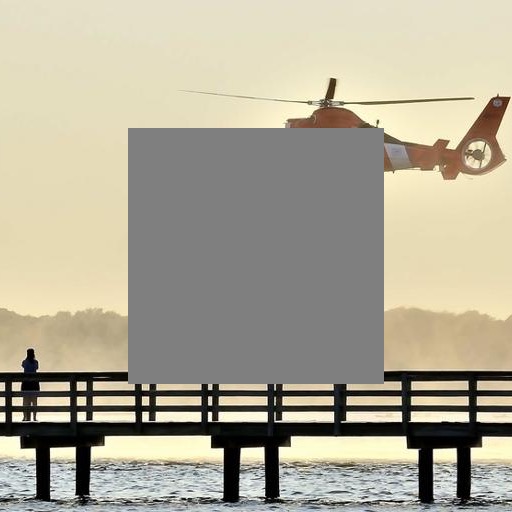}&
\includegraphics[width=\tmpwidth]{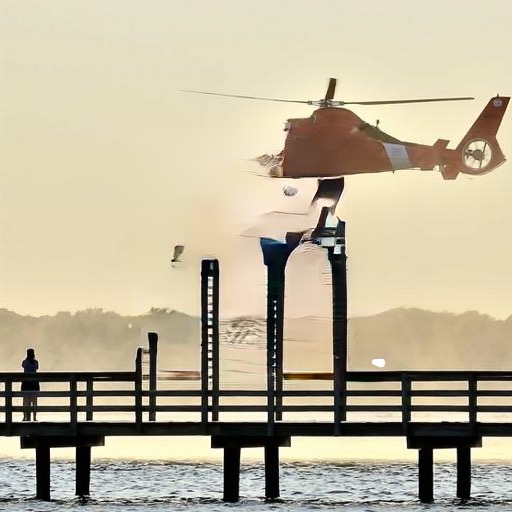}&
\includegraphics[width=\tmpwidth]{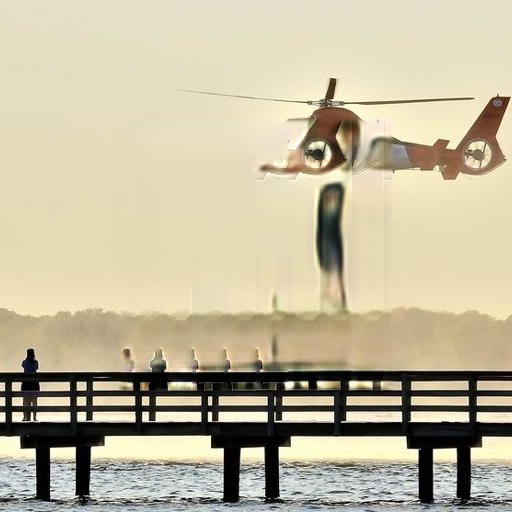}&
\includegraphics[width=\tmpwidth]{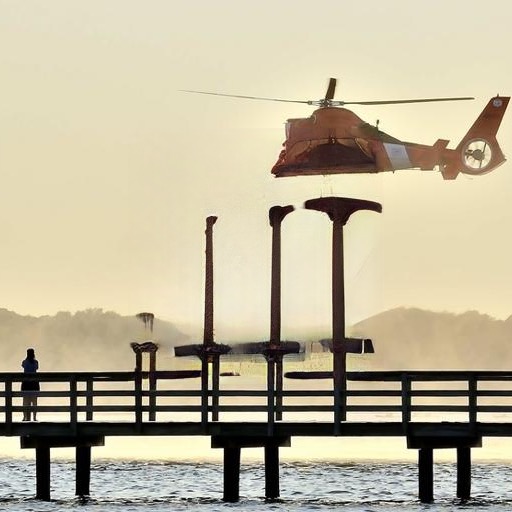}&
\includegraphics[width=\tmpwidth]{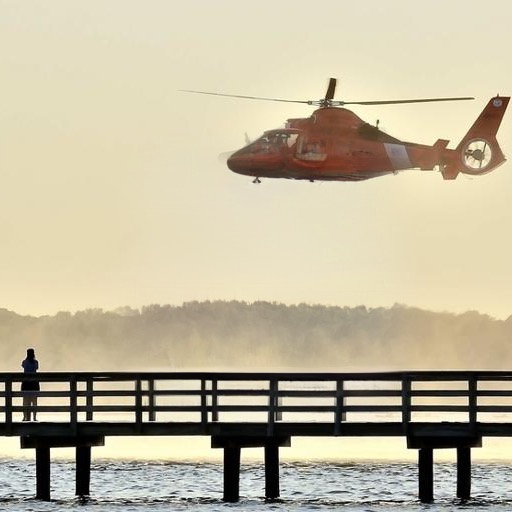}&
\includegraphics[width=\tmpwidth]{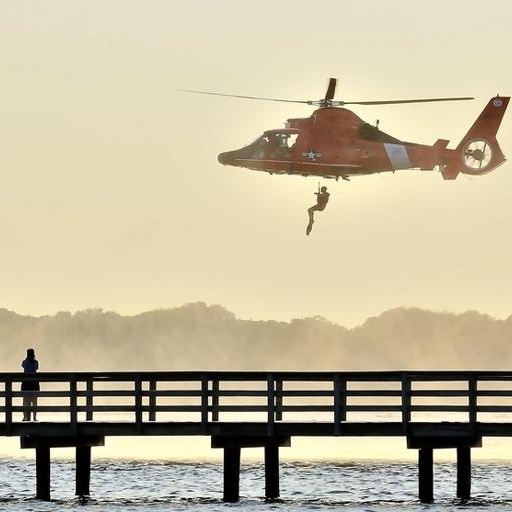}
\\

% \includegraphics[width=\tmpwidth]{figures/inpaint/000990_input.jpeg}&
% \includegraphics[width=\tmpwidth]{figures/inpaint/000990_deepfillv2.jpeg}&
% \includegraphics[width=\tmpwidth]{figures/inpaint/000990_hifill.jpeg}&
% \includegraphics[width=\tmpwidth]{figures/inpaint/000990_comodgan.jpeg}&
% \includegraphics[width=\tmpwidth]{figures/inpaint/000990_ours.jpeg}&
% \includegraphics[width=\tmpwidth]{figures/inpaint/000990_gt.jpeg}
% \\

% \includegraphics[width=\tmpwidth]{figures/inpaint/000158_input.jpeg}&
% \includegraphics[width=\tmpwidth]{figures/inpaint/000158_deepfillv2.jpeg}&
% \includegraphics[width=\tmpwidth]{figures/inpaint/000158_hifill.jpeg}&
% \includegraphics[width=\tmpwidth]{figures/inpaint/000158_comodgan.jpeg}&
% \includegraphics[width=\tmpwidth]{figures/inpaint/000158_ours.jpeg}&
% \includegraphics[width=\tmpwidth]{figures/inpaint/000158_gt.jpeg}
% \\

% \includegraphics[width=\tmpwidth]{figures/inpaint/000966_input.jpeg}&
% \includegraphics[width=\tmpwidth]{figures/inpaint/000966_deepfillv2.jpeg}&
% \includegraphics[width=\tmpwidth]{figures/inpaint/000966_hifill.jpeg}&
% \includegraphics[width=\tmpwidth]{figures/inpaint/000966_comodgan.jpeg}&
% \includegraphics[width=\tmpwidth]{figures/inpaint/000966_ours.jpeg}&
% \includegraphics[width=\tmpwidth]{figures/inpaint/000966_gt.jpeg}
% \\

\includegraphics[width=\tmpwidth]{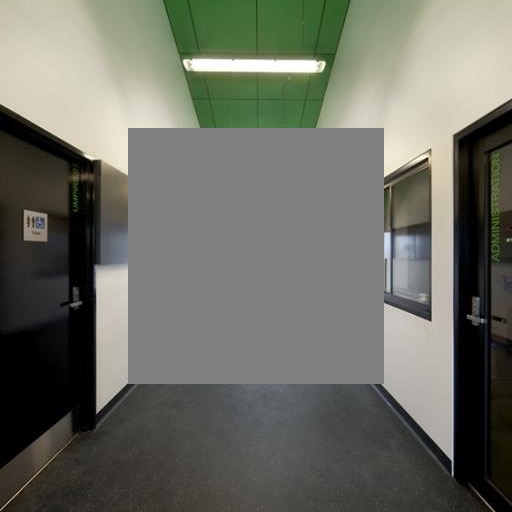}&
\includegraphics[width=\tmpwidth]{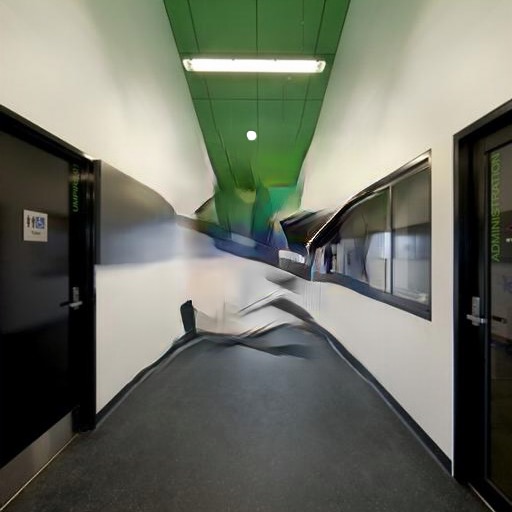}&
\includegraphics[width=\tmpwidth]{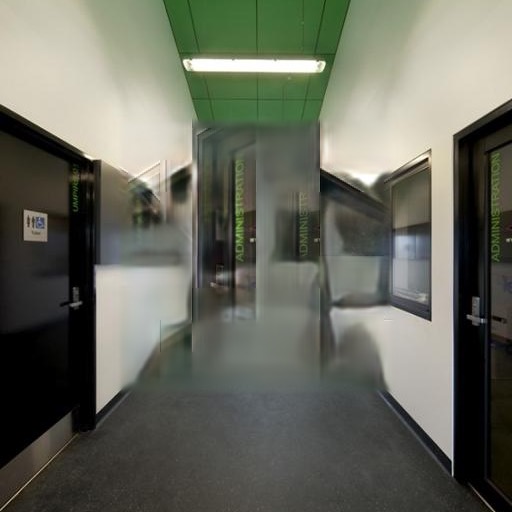}&
\includegraphics[width=\tmpwidth]{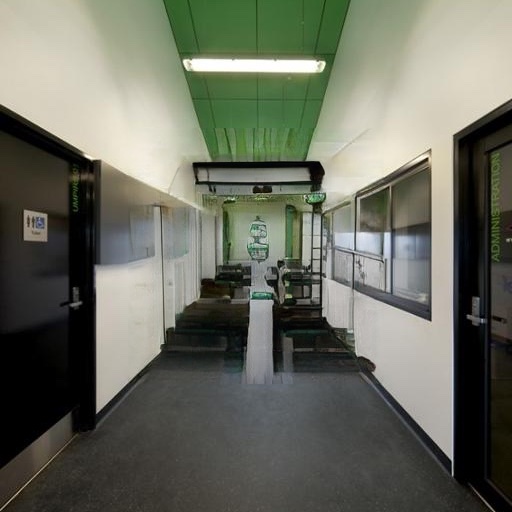}&
\includegraphics[width=\tmpwidth]{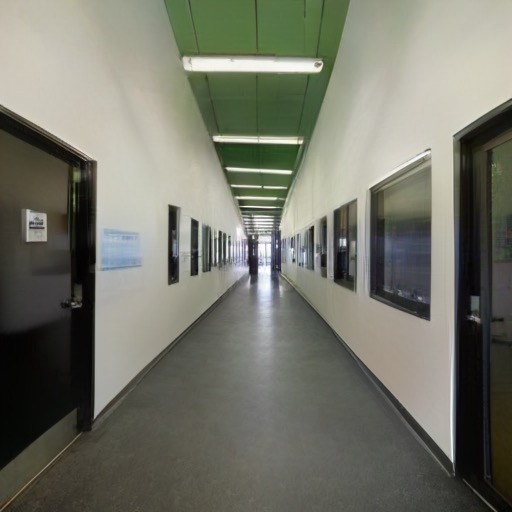}&
\includegraphics[width=\tmpwidth]{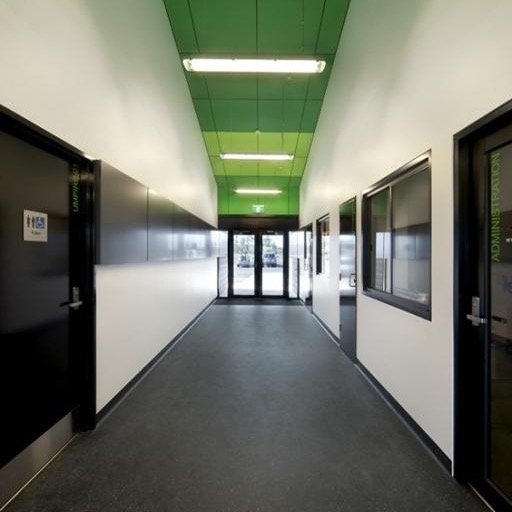}
\\

\includegraphics[width=\tmpwidth]{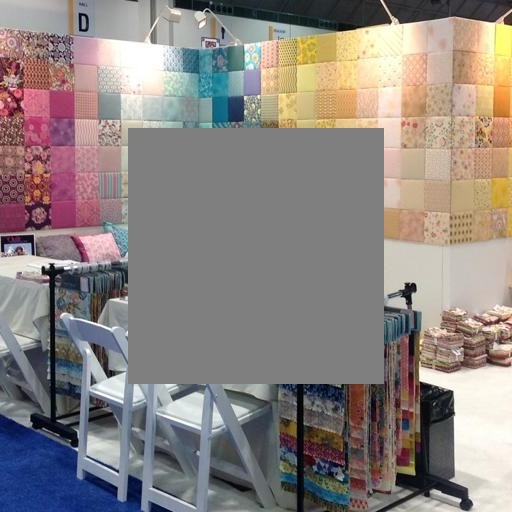}&
\includegraphics[width=\tmpwidth]{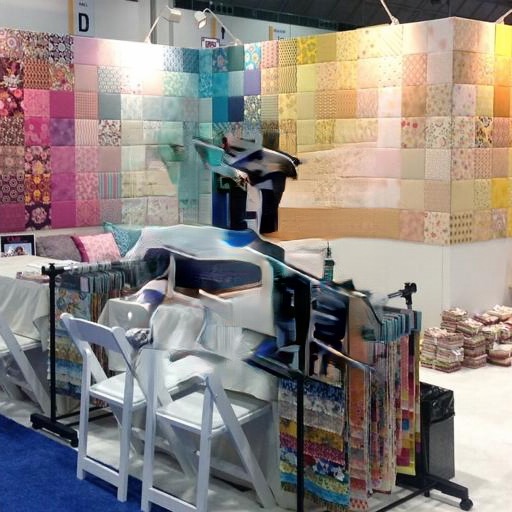}&
\includegraphics[width=\tmpwidth]{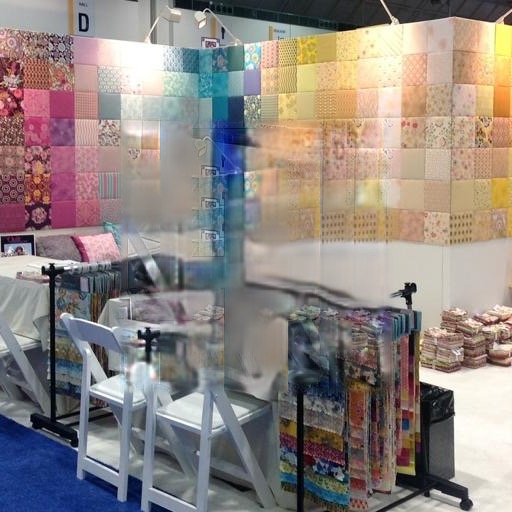}&
\includegraphics[width=\tmpwidth]{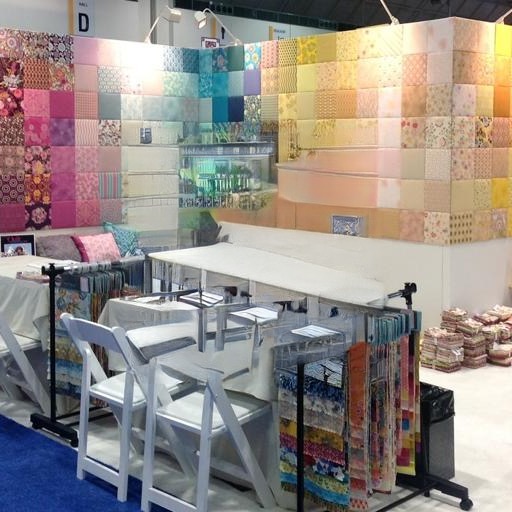}&
\includegraphics[width=\tmpwidth]{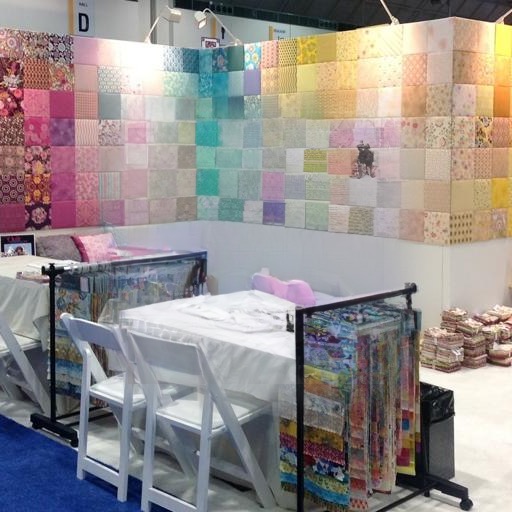}&
\includegraphics[width=\tmpwidth]{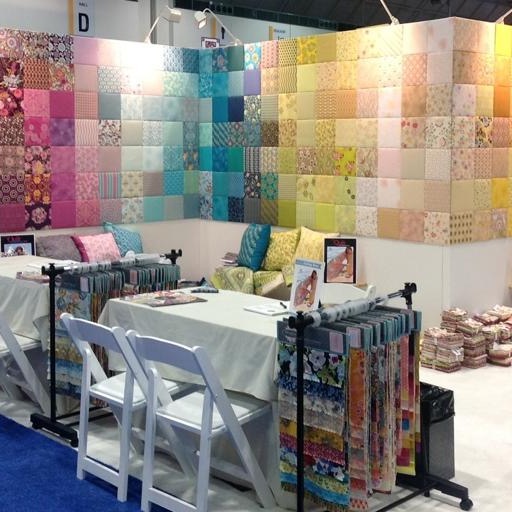}
\\

\includegraphics[width=\tmpwidth]{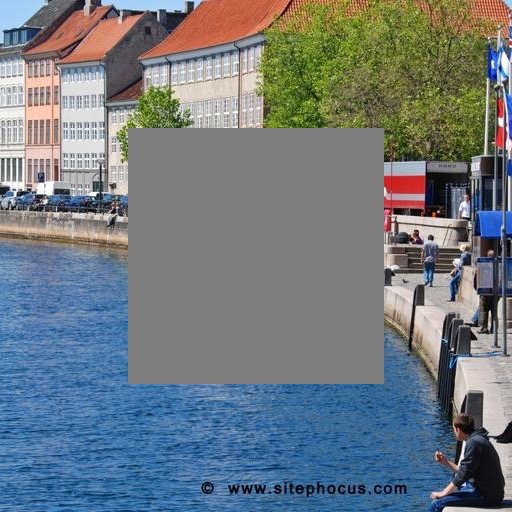}&
\includegraphics[width=\tmpwidth]{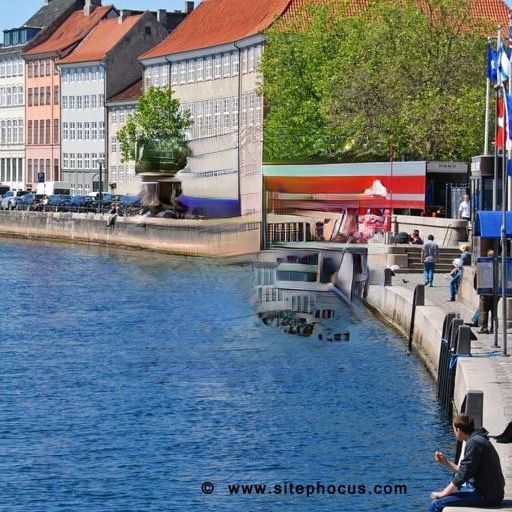}&
\includegraphics[width=\tmpwidth]{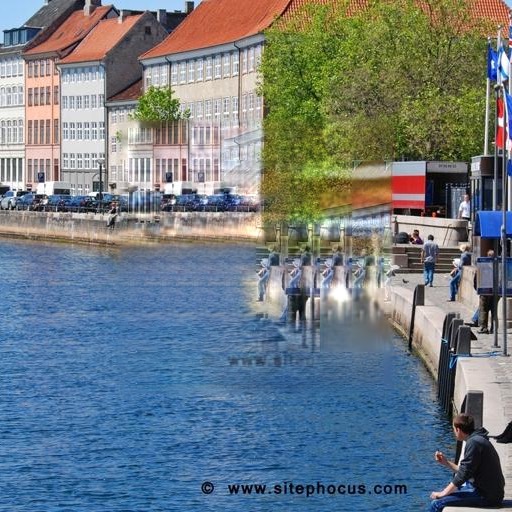}&
\includegraphics[width=\tmpwidth]{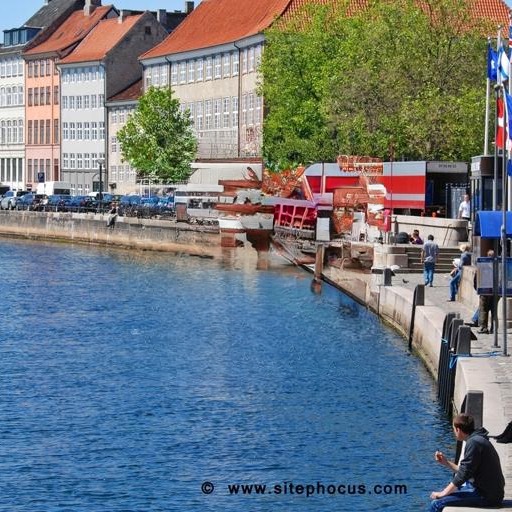}&
\includegraphics[width=\tmpwidth]{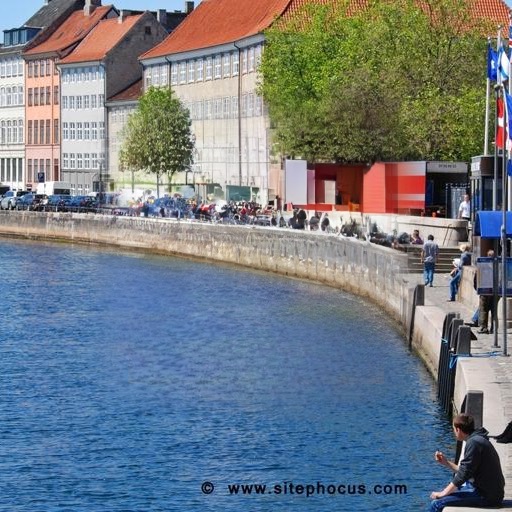}&
\includegraphics[width=\tmpwidth]{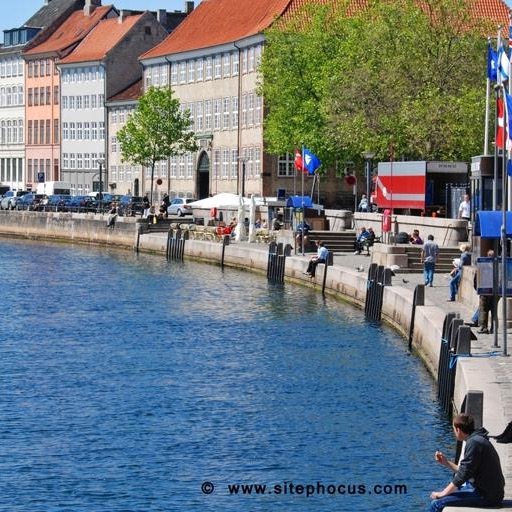}
\\

% \includegraphics[width=\tmpwidth]{figures/inpaint/000060_input.jpeg}&
% \includegraphics[width=\tmpwidth]{figures/inpaint/000060_deepfillv2.jpeg}&
% \includegraphics[width=\tmpwidth]{figures/inpaint/000060_hifill.jpeg}&
% &
% \includegraphics[width=\tmpwidth]{figures/inpaint/000060_ours.jpeg}&
% \includegraphics[width=\tmpwidth]{figures/inpaint/000060_gt.jpeg}
% \\

\includegraphics[width=\tmpwidth]{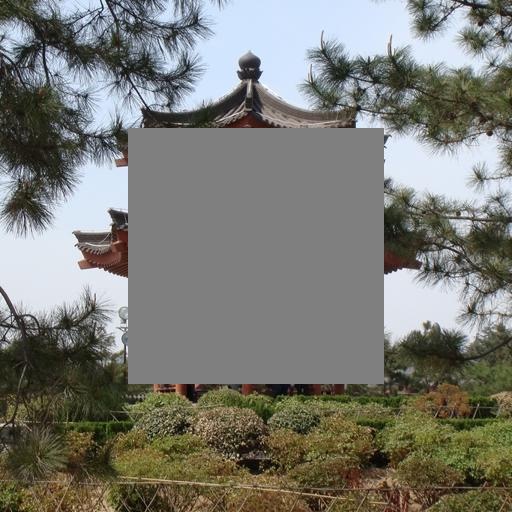}&
\includegraphics[width=\tmpwidth]{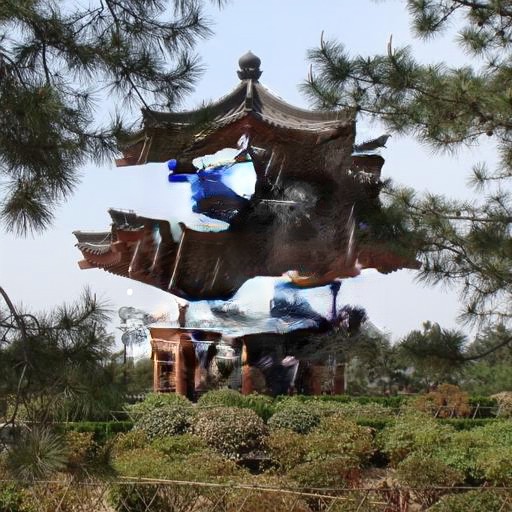}&
\includegraphics[width=\tmpwidth]{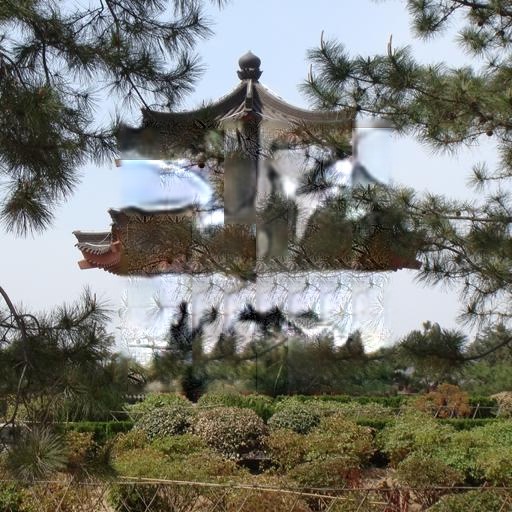}&
\includegraphics[width=\tmpwidth]{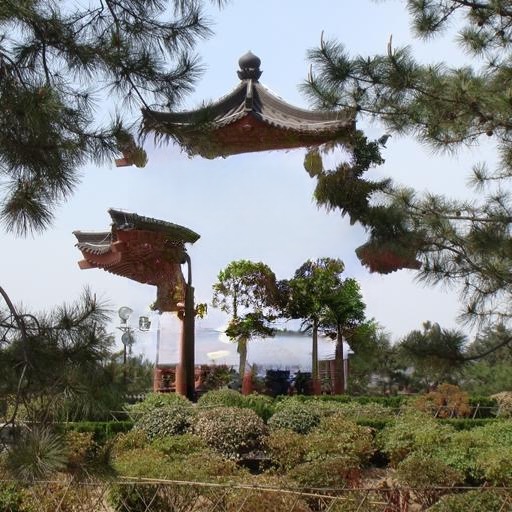}&
\includegraphics[width=\tmpwidth]{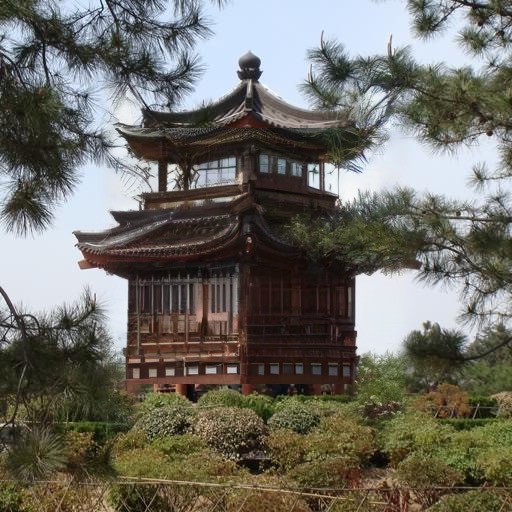}&
\includegraphics[width=\tmpwidth]{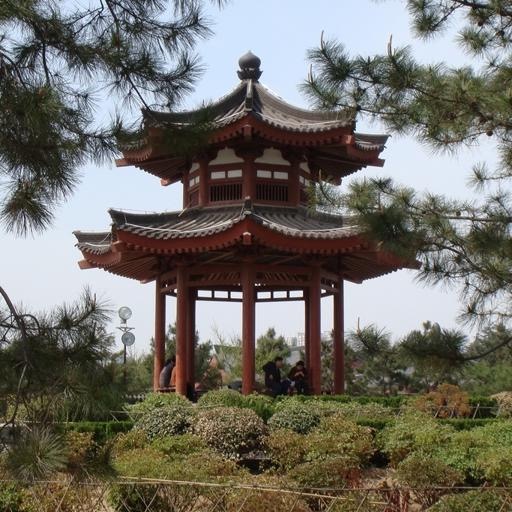}
\\

\includegraphics[width=\tmpwidth]{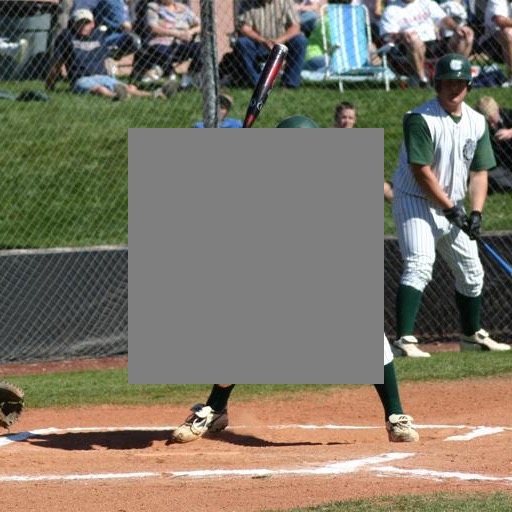}&
\includegraphics[width=\tmpwidth]{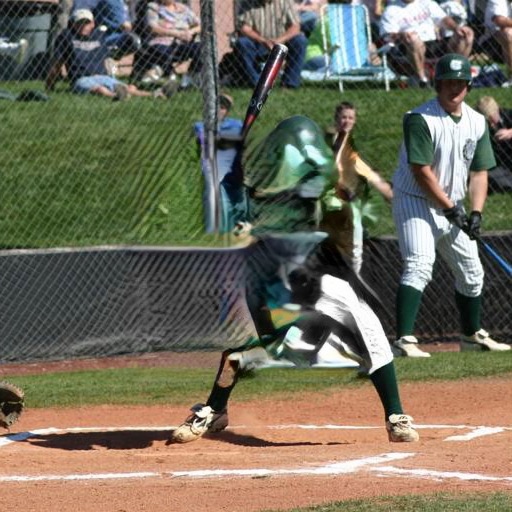}&
\includegraphics[width=\tmpwidth]{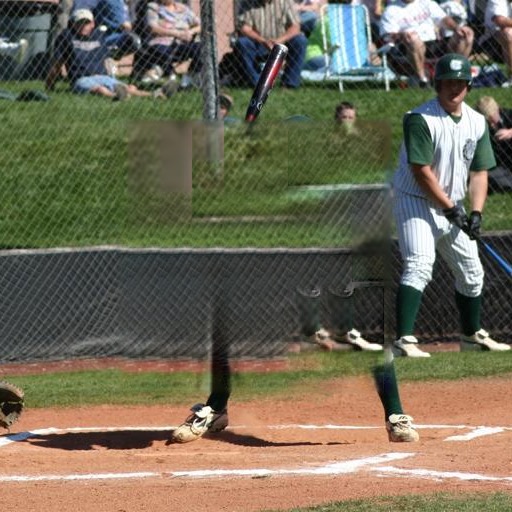}&
\includegraphics[width=\tmpwidth]{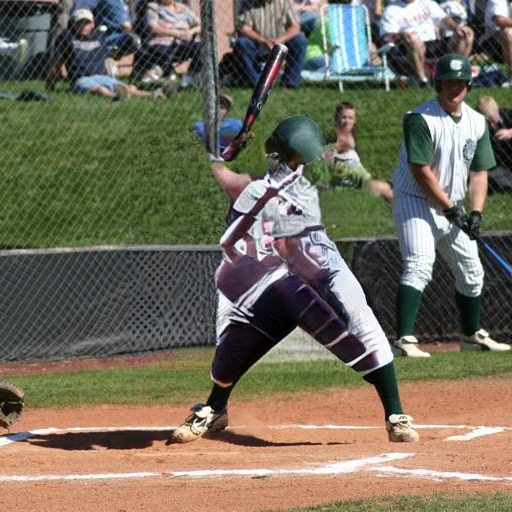}&
\includegraphics[width=\tmpwidth]{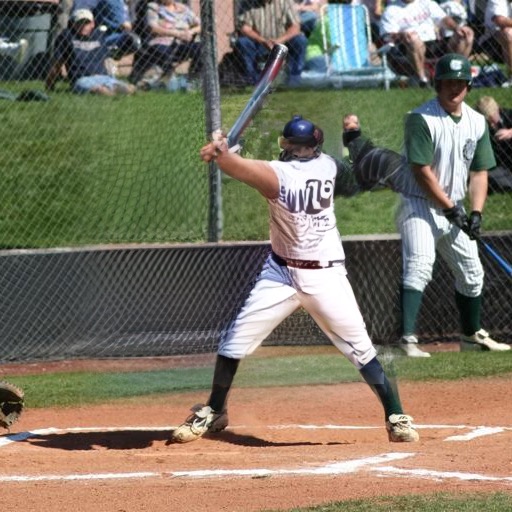}&
\includegraphics[width=\tmpwidth]{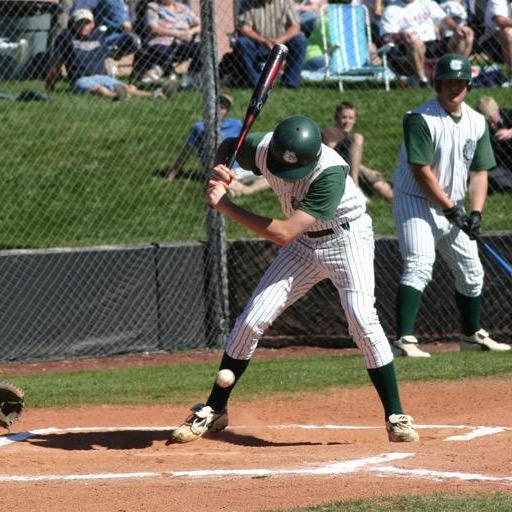}
\\
    \end{tabular}
    \caption{\textbf{More visual comparisons on image inpainting} on Places2\cite{zhou2017places} with state-of-the-art GAN methods.}

    \label{fig:supp_inpainting}
\end{figure*}

\begin{figure*}[!ht]
\centering
\begin{tabular}{cccccc}
Input & CoModGAN & \model (Ours) & Input & CoModGAN & \model (Ours) \\
\includegraphics[width=\tmpwidth]{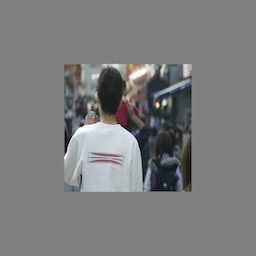}&
\includegraphics[width=\tmpwidth]{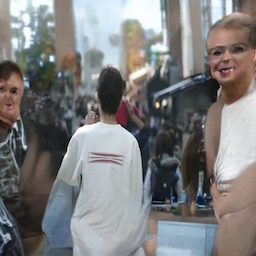}&
\includegraphics[width=\tmpwidth]{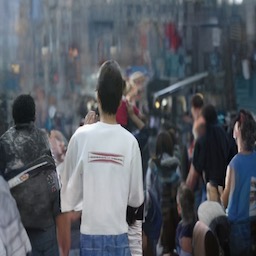}&
\includegraphics[width=\tmpwidth]{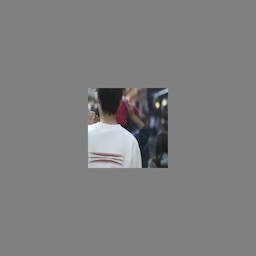}&

\includegraphics[width=\tmpwidth]{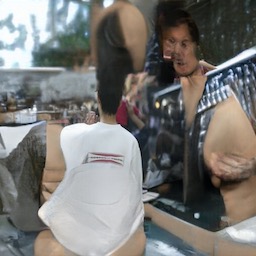}&

\includegraphics[width=\tmpwidth]{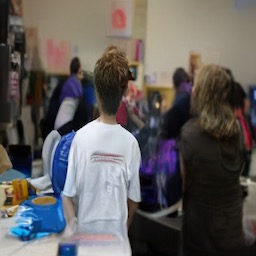} 
\\

\includegraphics[width=\tmpwidth]{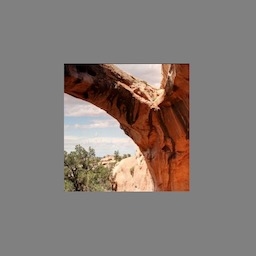}&
\includegraphics[width=\tmpwidth]{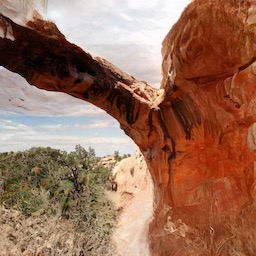}&
\includegraphics[width=\tmpwidth]{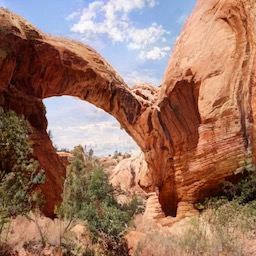}&
\includegraphics[width=\tmpwidth]{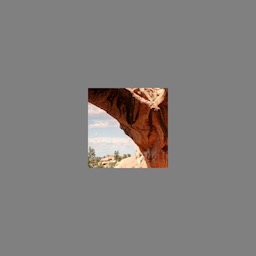} &
\includegraphics[width=\tmpwidth]{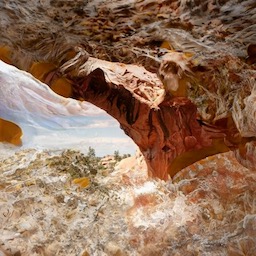}&
\includegraphics[width=\tmpwidth]{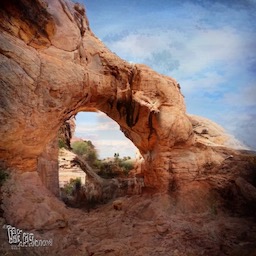}\\

\includegraphics[width=\tmpwidth]{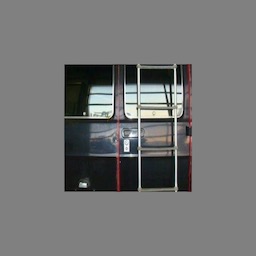} &
\includegraphics[width=\tmpwidth]{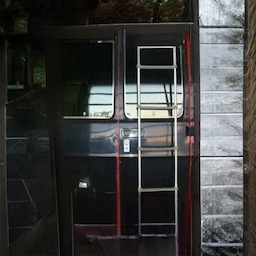}&
\includegraphics[width=\tmpwidth]{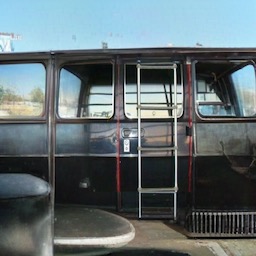}&
\includegraphics[width=\tmpwidth]{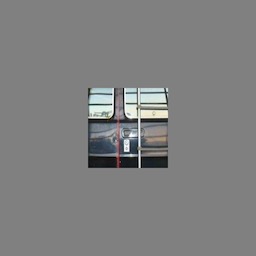}&
\includegraphics[width=\tmpwidth]{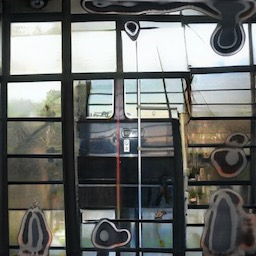}&
\includegraphics[width=\tmpwidth]{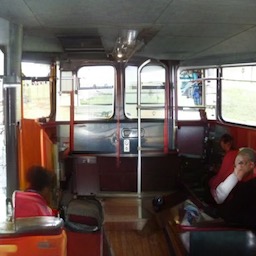} \\

\end{tabular}
\vspace{-3mm}
\caption{Visual comparisons of outpainting with CoModGAN\cite{zhao2021comodgan} on large outpainting mask.}
% \vspace{-4mm}
\label{fig:supp_uncrop_comparison}
\end{figure*}

In addition, we compare with CoModGAN on image completion tasks with large masking ratios, \ie conditioning on the center 50$\%\times$50\% and the center 31.25\%$\times$31.25\% respectively, which are challenging cases for traditional GANs. Examples are shown in Figure~\ref{fig:supp_uncrop_comparison}.

\setlength{\tabcolsep}{2pt}
\begin{figure*}[!ht]
    \centering
    \begin{tabular}{cccccc}
        Input & Boundless\cite{teterwak2019boundless} & InfinityGAN\cite{lin2021infinitygan}\footnotesize{\ding{83}} & CoModGAN & \textbf{\model (Ours)} & Groundtruth \\

    \includegraphics[width=\tmpwidth]{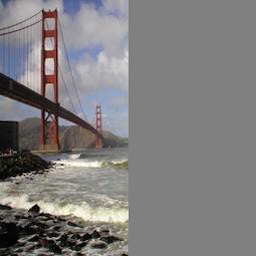}&
    \includegraphics[width=\tmpwidth]{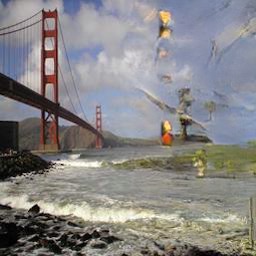}&
    \includegraphics[width=\tmpwidth]{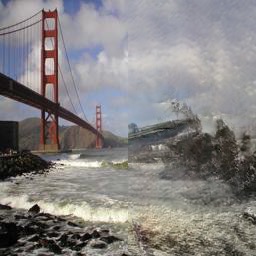}  &
    \includegraphics[width=\tmpwidth]{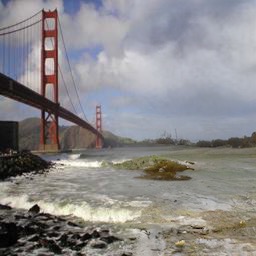}  &
    \includegraphics[width=\tmpwidth]{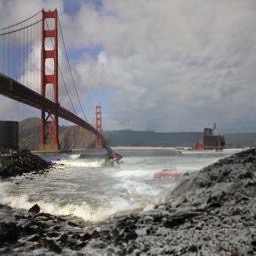}&
     \includegraphics[width=\tmpwidth]{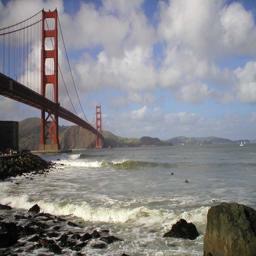}

    \\
    \includegraphics[width=\tmpwidth]{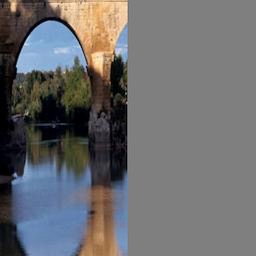}&
    \includegraphics[width=\tmpwidth]{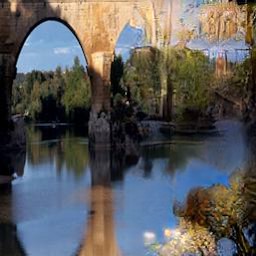}&
    \includegraphics[width=\tmpwidth]{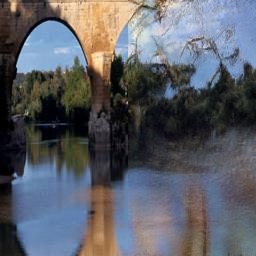}&
    \includegraphics[width=\tmpwidth]{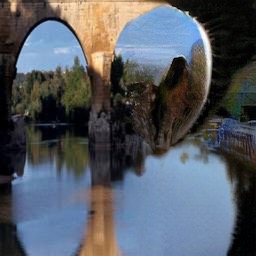}&  
    \includegraphics[width=\tmpwidth]{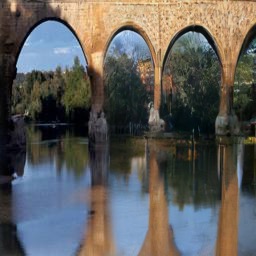} &
    \includegraphics[width=\tmpwidth]{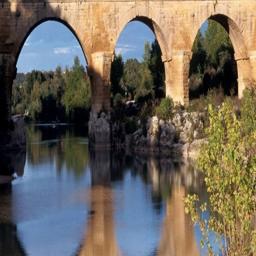} 
    \\
    \includegraphics[width=\tmpwidth]{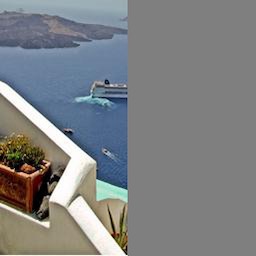}&
    \includegraphics[width=\tmpwidth]{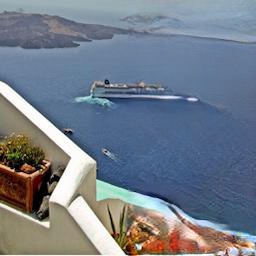}&
    \includegraphics[width=\tmpwidth]{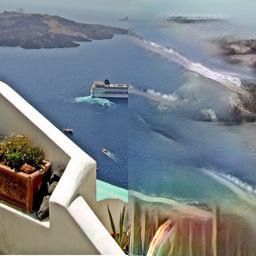} &
    \includegraphics[width=\tmpwidth]{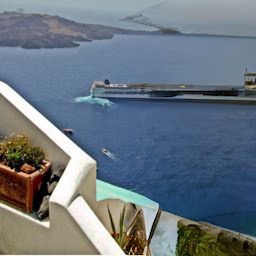} 
    & 
     \includegraphics[width=\tmpwidth]{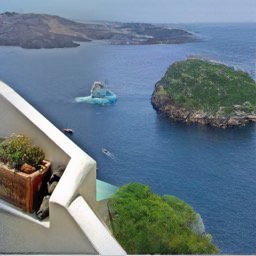}&
     \includegraphics[width=\tmpwidth]{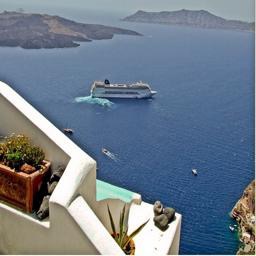}

    \\

    %  \includegraphics[width=\tmpwidth]{figures/uncrop/3145_input.jpeg}&
    % \includegraphics[width=\tmpwidth]{figures/uncrop/3145_boundless.jpeg}&
    % \includegraphics[width=\tmpwidth]{figures/uncrop/3145_inf.jpeg}&
    % &
    % \includegraphics[width=\tmpwidth]{figures/uncrop/3145_output7.jpeg}&
    % % \includegraphics[width=\tmpwidth]{figures/uncrop/3145_output1.jpeg}&
    % %\includegraphics[width=\tmpwidth]{figures/uncrop/3145_output2.jpeg}&
    % % \includegraphics[width=\tmpwidth]{figures/uncrop/3145_output6.jpeg}&
    % \includegraphics[width=\tmpwidth]{figures/uncrop/3145_gt.jpeg}
    % \\

    \end{tabular}
%    \vspace{-3mm}
   % \vspace{-4mm}
    \vspace{-2mm}   
    \caption{\textbf{More visual comparisons on image outpainting.} with state-of-the-art GAN methods. \ding{83} samples are graciously provided by the authors.
    }
    \label{fig:supp_uncrop_comparison_right}
    \vspace{-8mm}
\end{figure*}

\suppsection{Limitations and Failure Cases}
\label{sec:supp_limitations}

In Figure~\ref{fig:supp_failure}, we show several limitations and failure cases of our approach. (A) and (B) are examples of semantic and color shifts in \model's outpainting results. Due to its limited attention size, \model may "forget" the synthesized semantics or color from one end when it's outpainting the other end. (C) and (D) show cases where our approach may sometimes ignore or modify objects on the boundary when applied to outpainting and inpainting. (E) showcases \model's failure mode in which it causes oversmoothing or creates undesired artifacts on complex structures such as human faces, text and symmetric objects. The improvement for these circumstances remains future work. 

\renewcommand{\tmpwidth}{15mm}
\newcommand{\tmpheight}{30mm}
\setlength{\tabcolsep}{4pt}
\begin{figure*}[!ht]
    \centering
    \begin{tabular}{m{10mm}c c}

% (A) & \parbox[c]{\tmpwidth}{\includegraphics[ height=\tmpheight]{figures/supp_failure/chelsea-gates-0653_wY0nRc-unsplash.jpg_input.jpeg}} &
% \parbox[c]{9\tmpwidth}{\includegraphics[ height=\tmpheight]{figures/supp_failure/chelsea-gates-0653_wY0nRc-unsplash.jpg_output0006_1.jpeg}}
% \\ 

& Input & \parbox[c]{8\tmpwidth}{Our Outpainting Samples}  \\

~~~~~(A) & \parbox[c]{\tmpwidth}{\includegraphics[ height=\tmpheight]{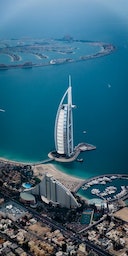}}  & 
\parbox[c]{9\tmpwidth}{\includegraphics[ height=\tmpheight]{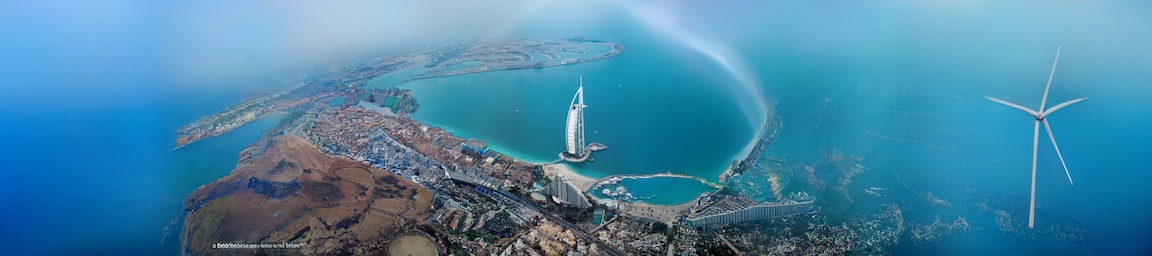}}
 \\

%  & Input & \parbox[c]{8\tmpwidth}{Our outpainting samples} \\ 
 
~~~~~(B) & \parbox[c]{\tmpwidth}{\includegraphics[ height=\tmpheight]{figures/panorama/janosch-diggelmann-qXzNdOfGnbw-unsplash.jpg}}  & 
\parbox[c]{9\tmpwidth}{\includegraphics[ height=\tmpheight]{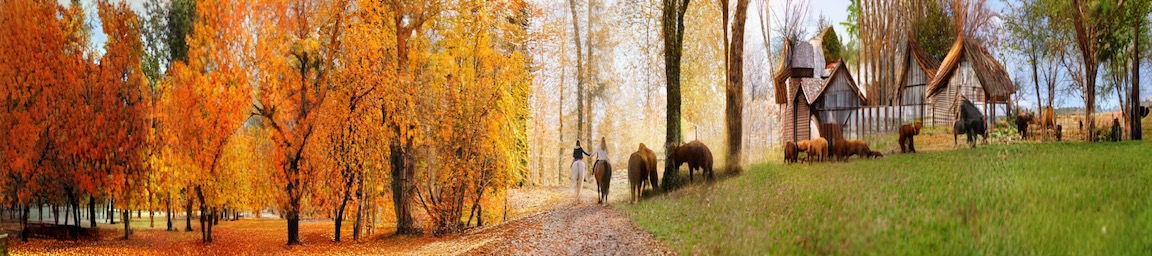}}
 \\ 
%  (C) & \parbox[c]{\tmpwidth}{\includegraphics[ height=\tmpheight]{figures/panorama/damiano-baschiera-hFXZ5cNfkOk-unsplash.jpg}}  & 
% \parbox[c]{9\tmpwidth}{\includegraphics[ height=\tmpheight]{figures/supp_failure/damiano-baschiera-hFXZ5cNfkOk-unsplash.jpg_output0001_1.jpeg}}
%  \\ 
 \end{tabular}
 \setlength{\tabcolsep}{1pt}
 \begin{tabular}{m{7mm}c c c c c }
    & Input & \multicolumn{3}{c}{------ Our Outpainting Samples ------} & Groundtruth
    \\
 (C) & 
     \parbox[c]{\tmpheight}{\includegraphics[height=\tmpheight]{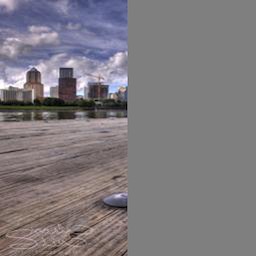}}&
    % \parbox[c]{\tmpheight}{\includegraphics[height=\tmpheight]{figures/uncrop/2982_boundless.jpeg}}&
    % \parbox[c]{\tmpheight}{\includegraphics[height=\tmpheight]{figures/uncrop/2982_inf.jpeg}}&
    \parbox[c]{\tmpheight}{\includegraphics[height=\tmpheight]{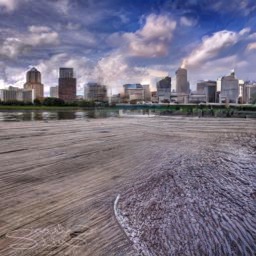}} &
    % \parbox[c]{\tmpheight}{\includegraphics[height=\tmpheight]{figures/uncrop/2982_output0.jpeg}} &
    \parbox[c]{\tmpheight}{\includegraphics[height=\tmpheight]{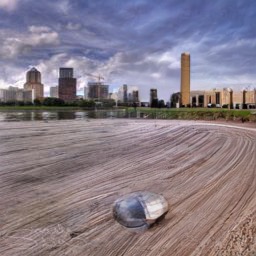}} &
    \parbox[c]{\tmpheight}{\includegraphics[height=\tmpheight]{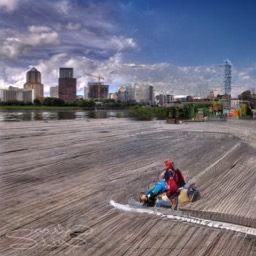}} &
    % \parbox[c]{\tmpheight}{\includegraphics[height=\tmpheight]{figures/uncrop/2982_output10.jpeg}} &
   \parbox[c]{\tmpheight}{ \includegraphics[height=\tmpheight]{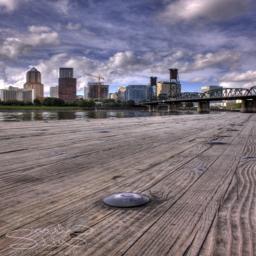}}
  \\
   & Input & \multicolumn{3}{c}{------Our Inpainting Samples ------} & Groundtruth 
   \\
  (D) & 
     \parbox[c]{\tmpheight}{\includegraphics[height=\tmpheight]{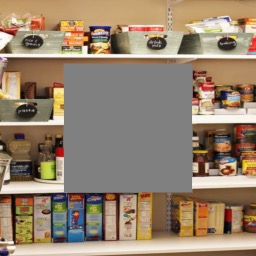}}&
    % \parbox[c]{\tmpheight}{\includegraphics[height=\tmpheight]{figures/uncrop/2982_boundless.jpeg}}&
    % \parbox[c]{\tmpheight}{\includegraphics[height=\tmpheight]{figures/uncrop/2982_inf.jpeg}}&
    \parbox[c]{\tmpheight}{\includegraphics[height=\tmpheight]{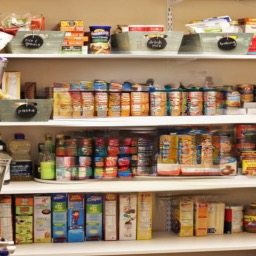}} &
    % \parbox[c]{\tmpheight}{\includegraphics[height=\tmpheight]{figures/uncrop/2982_output0.jpeg}} &
    \parbox[c]{\tmpheight}{\includegraphics[height=\tmpheight]{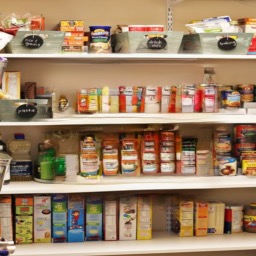}} &
    \parbox[c]{\tmpheight}{\includegraphics[height=\tmpheight]{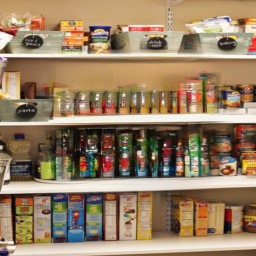}} &
    % \parbox[c]{\tmpheight}{\includegraphics[height=\tmpheight]{figures/uncrop/2982_output10.jpeg}} &
   \parbox[c]{\tmpheight}{ \includegraphics[height=\tmpheight]{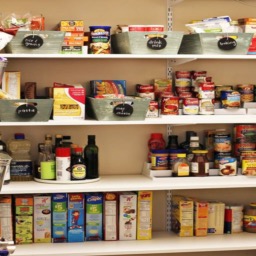}}
  \\
   & \multicolumn{5}{c}{------Our Class-conditional Samples ------} \\
(E) & \parbox[c]{\tmpheight}{\includegraphics[height=\tmpheight]{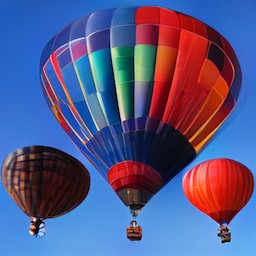}} & 
%  \parbox[c]{\tmpheight}{\includegraphics[height=\tmpheight]{figures/supp_failure/000967_0009_output.jpeg}} & 
\parbox[c]{\tmpheight}{\includegraphics[height=\tmpheight]{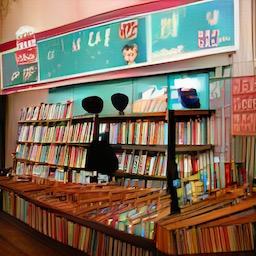}} & 
\parbox[c]{\tmpheight}{\includegraphics[height=\tmpheight]{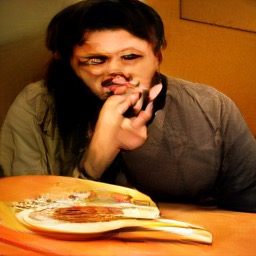}} &
\parbox[c]{\tmpheight}{\includegraphics[height=\tmpheight]{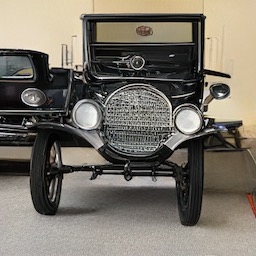}} &
\parbox[c]{\tmpheight}{\includegraphics[height=\tmpheight]{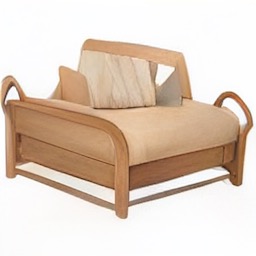}}
  
 \\
   \end{tabular}
    \caption{Limitations and Failure Cases. }
    \vspace{-6mm}
    \label{fig:supp_failure}
\end{figure*}

\end{document}